\documentclass[11pt]{article}
\pdfoutput=1

\def\pb{}

\usepackage[dvipsnames]{xcolor}

\def\beq{\begin{equation} }\def\eeq{\end{equation} }\def\1{\mathbf{1}}

\usepackage[framemethod=default]{mdframed}
\usepackage{caption}
\usepackage{indentfirst}
\usepackage{bm, mathrsfs, graphics,float,amssymb,amsmath,subeqnarray,setspace,graphicx,amsthm,epstopdf,subfigure, enumerate, color}
\usepackage[utf8]{inputenc}
\usepackage[colorlinks,
      linkcolor=red,
      anchorcolor=blue,
      citecolor=blue
      ]{hyperref}
\usepackage{natbib}

\usepackage{fullpage}
\numberwithin{equation}{section}

\newtheorem{lemma}{Lemma}
\newtheorem{theorem}{Theorem}

\newtheorem{remark}{Remark}

\ifx\assumption\undefined
\newtheorem{assumption}{Assumption}
\fi

\newcommand{\EE}{\mathbb{E}}
\newcommand{\RR}{\mathbb{R}}

\newcommand{\argmin}{\mathop{\mathrm{argmin}}}

\usepackage{bbm}

\usepackage{multirow}
\usepackage{tablefootnote}
\usepackage{colortbl}% http://ctan.org/pkg/xcolor
\usepackage{hhline}% http://ctan.org/pkg/hhline

\usepackage{algorithm}
\usepackage{algorithmic}

\begin{document}
\title{
Hessian-Aware Zeroth-Order Optimization for Black-Box Adversarial Attack
}

\author{
	Haishan Ye
	\thanks{
		Hong Kong University of Science and Technology;
		email: yhs12354123@gmail.com; zhuangbx@connect.ust.hk and tongzhang@tongzhang-ml.org
	}
	\and
	Zhichao Huang
	\footnotemark[1]
	\and
	Cong Fang
	\thanks{
		Peking University;
		email: fangcong@pku.edu.cn
	}
	\and
	Chris Junchi Li
	\thanks{
		Tencent AI Lab;
		email: junchi.li.duke@gmail.com
	}
	\and
	Tong Zhang
	\footnotemark[1]
}
\date{
	\today}

\maketitle

\def\RB{\RR}
\def\TH{\tilde{H}}
\newcommand{\ti}[1]{\tilde{#1}}
\def\diag{\mathrm{diag}}
\newcommand{\norm}[1]{\left\|#1\right\|}
\newcommand{\dotprod}[1]{\left\langle #1\right\rangle}
\def\EB{\EE}
\def\tr{\mathrm{tr}}

\begin{abstract}
Zeroth-order optimization is an important research topic in machine learning.
In recent years, it has become a key tool in black-box adversarial attack to neural network based image classifiers.
However, existing zeroth-order optimization algorithms rarely extract second-order information of the model function.
In this paper, we utilize the second-order information of the objective function and propose a novel \textit{Hessian-aware zeroth-order algorithm} called \texttt{ZO-HessAware}.
Our theoretical result shows that \texttt{ZO-HessAware} has an
improved zeroth-order convergence rate and query complexity under
structured Hessian approximation, where we propose a few approximation
methods for estimating Hessian.
Our empirical studies on the black-box adversarial attack problem validate that our algorithm can achieve improved success rates with a lower query complexity.
\end{abstract}

\tableofcontents

\pb\section{Introduction}
In this paper, we consider the following convex optimization problem:
\begin{align}
\min_{x\in\RB^d} f(x) \label{eq:prob}
\end{align}
where $f$ is differentiable and strongly convex. 
Optimization method that solves the above problem with function value access only is known as \emph{zeroth-order optimization} or \emph{black-box optimization} \citep{Nesterov2017,ghadimi2013stochastic}.

Zeroth-order optimization has attracted attention from the machine learning community \citep{bergstra2011algorithms,ilyas18a} and it is especially useful for solving problem~\eqref{eq:prob} where the evaluations of gradients $\nabla f(x)$ are difficult or even infeasible. 
One prominent example of the zeroth-order optimization is the \emph{black-box adversarial attack} on deep neural networks \citep{chen2017zoo,hu2017generating,papernot2017practical,ilyas18a}. 
In the black-box adversarial attack, only the inputs and outputs of the neural network are available to the system and backpropagation on the target neural network is prohibited.
Another application example is the hyper-parameter tuning which searches for the optimal parameters of deep neural networks or other learning models \citep{snoek2012practical,bergstra2011algorithms}. 

In the past years, theoretical works on zeroth-order optimization arise as alternatives of the corresponding first-order methods, and they estimate gradients using function value difference \citep{Nesterov2017,ghadimi2013stochastic,duchi2015optimal}.
However, these works on zeroth-order optimization have been concentrating on extracting gradient information of the objective function and \emph{failed} to utilize the second-order Hessian information and, to some extent, have \emph{not} fully exploited the information of models and enjoyed less competitive rates.
In this paper, we aim to take advantages of the model's second-order information and propose a novel method called \textit{Hessian-aware} zeroth-order optimization. 
Aligning with earlier works \citet{Nesterov2017,ghadimi2013stochastic} we present in this paper our gradient estimation as follows:
\begin{align}
g_\mu(x) = \frac{1}{b}\sum_{i=1}^{b}\frac{f(x+\mu\TH^{-1/2}u_i)-f(x)}{\mu}   \cdot     \TH^{1/2}u_i
, \quad\mbox{with} \;\mu >0 \label{eq:g-mu}
\end{align}
where $b$ is the batch size of points for gradient estimation and $\TH$ is an approximate Hessian at the evaluation point. 
With Eqn.~\eqref{eq:g-mu} at hands, the core update rule of our Hessian-aware zero-order algorithm, namely \texttt{ZO-HessAware}, is
\begin{align*}
x_{t+1} 
= x_t - \eta \TH^{-1}g_\mu(x_t)
=x_t - \eta \TH^{-1} \cdot \frac{1}{b}\sum_{i=1}^{b}\frac{f(x+\mu\TH^{-1/2}u_i)-f(x)}{\mu} \cdot \TH^{1/2}u_i
,
\end{align*}  
where $\eta$ is the step size. 
If one lets $\ti{g}_\mu(x)$ be defined as 
\begin{align}\label{eq:t_gmu}
\ti{g}_\mu(x) \triangleq \frac{1}{b} \sum_{i=1}^{b}\frac{f(x+\mu \ti{u}_i)-f(x)}{\mu} \cdot \ti{u}_i
,
\quad\mbox{with}\; \ti{u}_i \sim N(0, \TH^{-1})
,
\end{align} 
then using the linear transformation property of the multivariate Gaussian distribution, the update rule further reduces to 
\begin{equation}\label{eq:update}
	x_{t+1} = x_t - \eta \ti{g}_\mu(x_t)
	.
\end{equation}
In comparison, early zeroth-order literatures (for instance \cite{Nesterov2017}) conduct gradient estimation via
\begin{align}\label{eq:t_gmu_hat}
\hat{g}_\mu(x) \triangleq \frac{1}{b} \sum_{i=1}^{b}\frac{f(x+\mu u_i)-f(x)}{\mu}u_i
\quad\mbox{with}\;  u_i\sim N(0,I_d).
\end{align}
We refer \eqref{eq:t_gmu_hat} as \textit{vanilla zeroth-order optimization method}.
Comparing \eqref{eq:t_gmu} and \eqref{eq:t_gmu_hat}, one observes that the directions of updates $\ti{g}_\mu$ and $\hat{g}_\mu$ share the same form but admit different covariances.
Because $\ti{g}_\mu(x)$ contains Hessian information and shares the same form with estimated gradient $\hat{g}_\mu(x)$, $\ti{g}_\mu(x)$ can be regarded as a \emph{natural gradient} and \texttt{ZO-HessAware} can be regarded as \emph{natural gradient descent} method.
Perhaps surprising, in the context of zeroth-order optimization the difference between our zeroth-order update and earlier works boil down to the difference of search direction covariances.
Our special choice of covariance matrix as the approximate inversed Hessian allows us to incorporate Hessian information into the update rule in Eqn.~\eqref{eq:update}, which further achieves an improved theoretical convergence rate in zeroth-order optimization.

Let the approximate Hessian $\TH$ satisfy
$\rho \TH \preceq \nabla^2f(x) \preceq (2-\rho)\TH$, and  $\zeta\cdot I_d \preceq \TH$, 
with $0<\rho\leq 1$ and $\zeta>0$, and the algorithm be initialized at a point sufficiently close to the optimal solution.
Our main result concludes that \texttt{ZO-HessAware} with a proper step size achieves an iteration complexity of $N(\epsilon) = O\left(\frac{d}{b\rho}\log\left(\frac{1}{\epsilon}\right)\right)$, in order to obtain an $\epsilon$-accuracy solution.
If $f(x)$ has the strong convexity parameter $\tau$ and $\rho$ is chosen as $\tau/\lambda_{k+1}$ where $\lambda_{k+1}$ upper-bounds the Hessian's $(k+1)$-th largest eigenvalue, then \texttt{ZO-HessAware} enjoys an iteration complexity (or convergence rate) of
\begin{align*}
	N(\epsilon) = O\left(\frac{d\lambda_{k+1}}{b\tau}\log\left(\frac{1}{\epsilon}\right)\right).
\end{align*}
Furthermore, let \texttt{ZO-HessAware} be implemented with power-method based Hessian approximation (namely \texttt{ZOHA-PW}) \citep{balcan2016improved}, it achieves the following query complexity
\begin{equation*}
Q(\epsilon) = \tilde{O}\left(\frac{dk\lambda_{k+1}}{\tau}\log\left(\frac{1}{\epsilon}\right)\right)
.
\end{equation*}
Let $L$ be the smoothness parameter of the objective function, vanilla zeroth-order optimization method \citep{Nesterov2017} shows an iteration complexity of $O\left(\frac{dL}{b\tau}\log\left(\frac{1}{\epsilon}\right)\right)$ and a query complexity of $O\left(\frac{dL}{\tau}\log\left(\frac{1}{\epsilon}\right)\right)$).
Under a low-rank Hessian structure that $k\lambda_{k+1} \leq L$, our proposed \texttt{ZO-HessAware} algorithm enjoys a sharper theoretical convergence rate and query complexity than vanilla zeroth-order method \eqref{eq:t_gmu_hat}.

Though \texttt{ZOHA-PW} obtains a nice theoretical result of query complexity, it takes at least $O(d)$ queries due to the power method procedure. This is very expensive especially when the dimension $d$ is very large. Hence, we also propose two heuristic but practical methods to construct approximate Hessians with much lower query complexity.
\begin{enumerate}[(i)]
\item
First, we use Gaussian sampling method to approximate Hessian. This method only samples a small batch of points from the Gaussian distribution to estimate the Hessian with batch size  being much smaller than the data dimension $d$.

\item
Second, we propose diagonal Hessian approximation which is a popular method in training deep neural networks. This approximation approach does not need extra query to function value and can keep principal information of the Hessian which has been proved in training deep neural networks \citep{kingma2015adam,duchi2011adaptive,zeiler2012adadelta}. 
\end{enumerate}

To numerically justify the effectiveness of our zeroth-order algorithm \texttt{ZO-HessAware}, we apply it to the task of black-box adversarial attack in the neural network based image classifier \citep{ilyas18a,chen2017zoo}. 
The adversarial attack aims to find an example $x$ deviated microscopically from the given image $x_0$ but misclassified by the neural network classifier. 
We compare our algorithms with two state-of-the-arts, namely \texttt{PGD-NES} \citep{ilyas18a} and \texttt{ZOO} \citep{chen2017zoo}.
The comparison shows that our two \texttt{ZO-HessAware} algorithm variants take reduced function value queries while obtaining an improved success rate of attack and exhibits outstanding performance especially when the attack task is hard.
Furthermore to promote the attack success rate and reduce the query complexity, we propose a novel strategy call \texttt{Descent-Checking} which empirically brings an even higher success rate and lower query complexity.

\subsection{Main Contribution}

We summarize our main contribution as follows.
\begin{enumerate}[(i)]
	\item We exploit the Hessian information of the model function and propose a novel Hessian-aware zeroth-order algorithm called \texttt{ZO-HessAware}. We integrate Hessian information into gradient estimation while keeping the algorithmic form similar to vanilla zeroth-order method. Theoretically we show \texttt{ZO-HessAware} has a faster convergence rate and lower query complexity with power-method based Hessian approximation than existing work without Hessian information. 
	\item Several novel structured Hessian approximation methods are proposed including Gauss sampling method as well as the diagonalization method. The Hessian estimation via Gauss sampling only takes a few extra queries to the function value. In the construction of diagonal approximate Hessian, we use natural gradient which contains Hessian information other than a ordinary gradient which is used in training deep neural networks. 
	\item We propose a descent-checking trick for the black-box adversarial attack. This trick can significantly improve the success rate and reduce the number of queries.
	\item We empirically prove the power of Hessian information in zeroth-order optimization especially in the black-box adversarial attack. Experiment results show that our \texttt{ZO-HessAware} type algorithm can achieve better success rates and need fewer queries than state-of-the-art algorithms especially when the problem is hard. 
\end{enumerate}

\pb\subsection{Related Work}

Zeroth-order optimization minimizes functions only through the function value oracles. It is an important research topic in optimization \citep{Nesterov2017,matyas1965random,ghadimi2013stochastic}. \citet{Nesterov2017} utilized random Gaussian vectors as the search directions and gave the convergence properties of the zeroth-order algorithms when the objective function is convex. \citet{ghadimi2013stochastic} proposed new zeroth-order algorithms which has better convergence rate when the problem is non-smooth. Zeroth-order method with variance reduction was proposed to solve non-convex problem recently \citep{fang2018spider,Liu2018}. Zeroth-order algorithm is also a crucial research topic in the on-line learning. Lots of results were obtained in the recent years \citep{shamir2017optimal,bach2016highly,duchi2015optimal}. In these works, one can only access to the function value and uses this feed-back to approximate the gradient or sub-gradient. 

Recently, zeroth-order algorithm is becoming the main tool for the black-box adversarial attack \citep{chen2017zoo,ilyas18a}. \citet{chen2017zoo} extended the CW \citep{carlini2017towards} attack which is a powerful white-box method to the black-box attack and proposed \texttt{ZOO}. Algorithm \texttt{ZOO} can be viewed as a kind of zeroth-order stochastic coordinate descent \citep{chen2017zoo}. It chooses a coordinate randomly, then uses zeroth-order oracles to estimate the gradient of current coordinate. However, \texttt{ZOO} suffers from a poor query complexity because it needs $O(d)$ queries to estimate the gradients of all the coordinates theoretically. To reduce the query complexity, \citet{ilyas18a} resorted to the natural evolutionary strategies \citep{wierstra2014natural} to estimate gradients. And \citet{ilyas18a} used the so-called `antithetic sampling' technique \citep{salimans2017evolution} to get better performance. 

Furthermore, covariance matrix adaptation evolution strategy (\texttt{CMA-ES}) is another important zeroth-order method which is closely related to our algorithm \citep{hansen2001completely}. 
\texttt{CMA-ES} uses a learned covariance to generate search direction and this covariance matrix is much like the inversion of our approximate Hessian. The main difference between these two algorithms is the way to use zeroth-order oracles. Eqn.~\eqref{eq:t_gmu} shows that \texttt{ZO-HessAWare} queries to function value to approximate a natural gradient. In contrast, \texttt{CMA-ES} generates $u_i$'s and picks up such $\ti{u}_i$'s as the search directions that $f(x+\ti{u}_i)$ is mall.

\vspace{.1in}
\noindent\textbf{Organization.}
The rest of this paper is organized as follows. In Section~\ref{sec:NP}, we present notation and preliminaries. In Section~\ref{sec:HessAware}, we depict Algorithm \texttt{ZO-HessAware} in detail and analyze its local and global convergence rate, and query complexity with power-method based Hessian approximation, respectively. In Section~\ref{sec:H_app}, we propose two different strategies to construct a good approximate Hessian. In Section~\ref{sec:experiments}, we compare our \texttt{ZO-HessAware} type algorithms with two state-of-the-art algorithms in the adversarial attack problem. Finally, we conclude our work in Section~\ref{sec:conclusion}. All the detailed proofs are deferred to the appendix in their order of
appearance.

\pb\section{Notation and Preliminaries}\label{sec:NP}

We first introduce notation that will be used in this paper. Then, we give some assumptions about the objective function that will be used. 
\pb\subsection{Notation}
Given a positive semi-definite matrix $A\in \RB^{d \times d}$ of rank-$\ell$ and a positive integer $k\leq \ell$, its eigenvalue decomposition is given as
\begin{align}
A=U\Lambda U^{T}=U_{k} \Lambda_{k} U_{k}^{T}+U_{\setminus k} \Lambda_{{\setminus} k} U_{{\setminus}k}^{T}, \label{eq:svd}
\end{align}
where $U_{k}$ and $U_{{\setminus}k}$ contain the eigenvectors of $A$, and $\Lambda=\diag(\lambda_1, \ldots, \lambda_d)$ with $\lambda_1\geq \lambda_2 \geq \cdots \geq \lambda_\ell>0$ are
the nonzero eigenvalues of $A$. We also use $\lambda_{\max}$ and $\lambda_{\min}$ to denote the largest and smallest eigenvalue of a positive semi-definite matrix, respectively. 

Using matrix $A$, we can define $A$-norm as $\|x\|_A = \sqrt{x^TAx}$. Furthermore, if $B$ is a positive semi-definite matrix, we say $B\preceq A$ when $A-B$ is positive semi-definite.

\pb\subsection{Properties of Smoothness and Convexity}
In this paper, we consider functions with $L$-smoothness and $\tau$-strongly convexity. It indicates the following properties.

\paragraph{$L$-smoothness}
Let function $f(x)$ be $L$-smooth, i.e.~for all $x,y\in\RR^d$, we have
\begin{equation}\label{eq:L_1}
\norm{\nabla f(x) - \nabla f(y)} \leq L\norm{x - y},
\end{equation}
or equivalently
\begin{equation}
\left|f(y) - f(x) - \dotprod{\nabla f(x), y-x}\right|\leq \frac{L}{2}\norm{x-y}^2. \label{eq:L_2}
\end{equation}

\paragraph{$\tau$-strong convexity}
Let function $f(x)$ be $\tau$-strongly convex, i.e.~for all $x,y \in \RR^d$
\begin{align*}
f(y)\geq f(x) + \dotprod{\nabla f(x), y-x} + \frac{\tau}{2}\norm{x-y}^2
.
\end{align*}

\paragraph{$\gamma$-Lipschitz Hessian}
Let the Hessian of $f(x)$ be $\gamma$-Lipschitz continuous, that is, for all $x,y \in \RR^d$
\begin{align}
\norm{\nabla^2 f(y) - \nabla^2f(x)} \leq \gamma \norm{y-x}
.
\label{eq:gamma_1}
\end{align}
or equivalently 
\begin{align}
	\left|f(y) - f(x) - \dotprod{\nabla f(x), y-x} -\frac{1}{2}\dotprod{\nabla^2f(x)(y-x),y-x}\right|\leq\frac{\gamma}{6}\norm{x-y}^3,\; x,y\in\RB^d \label{eq:gamma_2}
\end{align}

\pb\subsection{Gaussian Smoothing}
Let $f(x)$ be a function which is differentiable along any direction in $\RB^d$. The \emph{Gaussian smoothing} of $f(x)$ is defined as
\begin{align}
	f_\mu(x) \triangleq \frac{1}{M}\int_{\RB^d}f(x+\mu u)\;\exp{\left(-\frac{\norm{u}^2}{2}\right)}du, \;\mbox{where}\; u\sim N(0, I_d), \label{eq:f_mu}
\end{align}
and 
\begin{align*}
	M = \int_{\RB^d}\;\exp{\left(-\frac{\norm{u}^2}{2}\right)}du = (2\pi)^{d/2}.
\end{align*}
And $\mu$ is the parameter to control the smoothness. $f_\mu(x)$ preserves several important properties of $f(x)$. For example, if $f(x)$ is convex, then $f_\mu(x)$ is also convex. If $f(x)$ is $L$-smooth, then $f_\mu(x)$ is also $L$-smooth.

\pb\section{Hessian-Aware Zeroth-Order Method}\label{sec:HessAware}

In this section, we will exploit the Hessian information of the model function which was commonly ignored in the past works on zeroth-order optimization and propose \texttt{ZO-HessAware} algorithm.

Our algorithm first constructs an approximate Hessian $\TH$ for the current point $x$ satisfies 
\begin{align}
\rho\TH \preceq \nabla^2f(x) \preceq (2-\rho)\cdot \TH, \;\mbox{and},\;\; \zeta\cdot I_d \preceq \TH, \label{eq:prec_cond}
\end{align}
with $0<\rho\leq 1$. Parameter $\rho$ measures how well $\TH$ approximates $\nabla^2f(x)$. If $\rho=1$, then $\TH$ is the exact Hessian.  
One can use different methods to construct such $\TH$. In Section~\ref{sec:H_app}, we will provide several approaches to compute a good approximate Hessian with a small number of queries to the function value. Note that, we do not need to construct an approximate Hessian for each iteration. Empirically, we can only update it every $p$ iterations where $p$ is a parameter controls the frequency of the Hessian approximation.

Then we begin to estimate the gradient by derivative-free oracles. Different from the existing zeroth-order works \citep{Nesterov2017,ghadimi2013stochastic,duchi2015optimal}, the Hessian information is used in our gradient estimation represented as follows: 
\begin{align}
	g_\mu(x) = \frac{1}{b}\sum_{i=1}^{b}\frac{f(x+\mu Ku_i)-f(x)}{\mu}K^{-1}u_i,\;\mbox{with}\; K = \TH^{-1/2}, \; u_i \sim N(0, I)	\label{eq:g_mu}
\end{align}
where $b$ is the batch size. 
On the point $x$, we sample $b+1$ points  to obtain a good gradient estimation. This strategy is widely used in real applications such as adversarial attack.

Finally, analogue to Newton-style algorithms, we update $x_{t+1}$ using the approximate Hessian and estimated gradient as $x_{t+1} = x_t - \eta\TH^{-1}g_\mu(x)$.
Combining with Eqn.~\eqref{eq:g_mu}, we represent the algorithmic procedure of \texttt{ZO-HessAware} as follows 
\begin{equation*}
	\left\{
	\begin{split}
	\ti{g}_\mu(x_t) =& \frac{1}{b} \sum_{i=1}^{b}\frac{f(x_t+\mu \TH_t^{-1/2} u_i)-f(x_t)}{\mu}\TH_t^{-1/2} u_i,\;\mbox{with}\; u_i \sim N(0, I)\\
	x_{t+1}=& x_t - \eta \ti{g}_\mu(x_t).
	\end{split}
	\right.
\end{equation*}
We depict the detailed algorithmic procedure of \texttt{ZO-HessAware}in Algorithm~\ref{alg:zero_order}.

In the rest of this section, we will first give some important properties of the estimated gradient computed as Eqn.~\eqref{eq:g_mu}. Then we analyze the local and global convergence property of Algorithm~\ref{alg:zero_order}, respectively. Finally, the query complexity will be analyzed with power-method based Hessian approximation.
 
\begin{algorithm}[tb]
	\caption{Algorithm \texttt{ZO-HessAware}}
	\label{alg:zero_order}
	\begin{small}
		\begin{algorithmic}[1]
			\STATE {\bf Input:} $x^{(0)}$ is an initial point sufficient close to $x^{*}$. And $b$ is the batch size and $p$ is an integer. Parameter $\eta$ is the step size.
			\FOR {$t=0,\dots,T$ }
			\IF {$t\bmod p == 0$}
			\STATE Compute an approximate Hessian $\TH_t$ satisfies Eqn.~\eqref{eq:prec_cond}.
			\ENDIF
			\STATE Generate $b$ samples with $u_i\sim N(0, I_d)$ and construct $\ti{g}_\mu(x_t) = \frac{1}{b}\sum_{i=1}^b \frac{f(x+ \TH_t^{-1/2} u_i) -f(x)}{\mu} \TH_t^{-1/2} u_i$;
			\STATE Update $x_{t+1} = x_t - \eta \ti{g}_\mu(x_t)$.\label{step:update} 
			\ENDFOR
		\end{algorithmic}
	\end{small}
\end{algorithm}	
 
\pb\subsection{Properties of Estimated Gradient} \label{subsec:grad_mu}

Now, we list some important properties of $g_\mu(x)$ defined in Eqn.~\eqref{eq:g_mu} that will be used in our analysis of convergence rate of \texttt{ZO-HessAware} in the following lemmas. These lemmas are also of independent interest in zeroth-order algorithm.
\begin{lemma}\label{lem:T_1}
	Let $f(x)$ be $L$-smooth, then $g_\mu(x)$ defined in Eqn.~\eqref{eq:g_mu} satisfies that
	\begin{align*}
	\norm{\EB_u[g_\mu(x)] - \nabla f(x)}_{K^2}^2 \leq \frac{\mu^2}{4}L^2\norm{K}^4(d+3)^{3}.
	\end{align*}
\end{lemma}

\begin{lemma}\label{lem:T_2}
	Let $f(x)$ be $L$-smooth, then $\norm{g_\mu(x)}_{K^2}^2$ can be bounded as
	\begin{align*}
	\EB_{u} \norm{g_\mu(x)}^2_{K^2} \leq\frac{\mu^2}{2b}L^2\norm{K}^4(d+6)^3 + \left(\frac{2(d+2)}{b} + 2\right) \cdot\norm{\nabla f(x)}_{K^2}^2 + \frac{\mu^2L^2(d+3)^3}{2}.
	\end{align*}
\end{lemma}

Now we give the bound of $\EB_u\norm{K^2 g_\mu(x)}_{K^{-2}}^3$ in the following lemma.
\begin{lemma}\label{lem:T_3}
	Let $f(x)$ be $L$-smooth, then $g_\mu(x)$ has such a property that
	\begin{align*}
	\EB_u\norm{K^2 g_\mu(x)}_{K^{-2}}^3 \leq 2\mu^3L^3\norm{K}^6\cdot(d+9)^{9/2}+12(d+5)^{3/2}\norm{\nabla f(x)}_{K^2}^3.
	\end{align*}
\end{lemma}

\pb\subsection{Local Convergence}\label{subsec:local}

Now we begin to analyze the local convergence property of Algorithm~\ref{alg:zero_order}. To achieve a fast convergence rate, the initial point should be close enough to the optimal point. At the same time, the Hessian should be well-approximated. 
\begin{theorem}\label{thm:local}
	Let $f(x)$ be $\tau$-strongly convex and $L$-smooth, and $\nabla^2f(x)$ is $\gamma$-Lipschitz continuous. Let the approximate Hessian $\TH_t$ satisfy Eqn.~\eqref{eq:prec_cond}. 
	Setting the step size $\eta = \frac{b}{16(d+2)}$ and $b\leq d+2$, then Algorithm~\ref{alg:zero_order} has the following convergence properties:
	\begin{align*}
	\EE\left[f(x_{t+1})-f(x^\star)\right] 
	\leq \left(1 - \frac{b\rho}{64(d+2)}\right)\left(f(x_t) -f(x^\star) \right) +\Delta_\mu,
	\end{align*}
	if $x_t$ satisfies that 
	\begin{align}
	\norm{x_t - x^\star} \leq \frac{\rho}{\gamma}\cdot\min\left(\frac{3\tau\zeta^{1/2}}{ 64L^{1/2}}, \frac{d^{3/2}\zeta^{2}}{33 L(d+2)b^2}\right)
	, \label{eq:x_local}
	\end{align}	
	and $\Delta_\mu$ is defined as 
	\begin{align*}
	\Delta_\mu=b\cdot\bigg(\frac{\mu^2L^2}{64\zeta^2}(d+5)^2+\frac{\mu^2L^2}{128\zeta}(d+38)+\frac{\gamma b^2\mu^3L^3}{6144\zeta^{9/2}}\cdot(d+110)^{3/2}\bigg).
	\end{align*}
\end{theorem}

\begin{remark}\label{rk:check}
	Note that, the local convergence properties rely on the condition~\eqref{eq:x_local}. However, this condition may be violated for next iteration if the descent direction is not good. This problem can be remedied by checking the value of $f(x_{t+1})$. We will discard the current $x_{t+1}$ if $f(x_{t+1})$ is larger than $f(x_t)$.
\end{remark}
Let $\rho$ be chosen $O(\tau/ \lambda_{k+1})$, then
to achieve an $\epsilon$-accuracy solution, by Theorem~\ref{thm:local}, our \texttt{ZO-HessAware} needs 
\begin{align*}
N(\epsilon) = O\left(\frac{d}{b\rho}\log\left(\frac{1}{\epsilon}\right)\right) = O\left(\frac{d\lambda_{k+1}}{b\tau}\log\left(\frac{1}{\epsilon}\right)\right)
\end{align*}
iterations. 
In contrast, without the second-order information, vanilla zeroth-order method needs $O\left(db^{-1}\kappa\log\left(\frac{1}{\epsilon}\right)\right)$ iterations \citep{Nesterov2017}, where $\kappa = L/\tau$ is the condition number . Since it holds that $\lambda_{k+1} \leq L$, \texttt{ZO-HessAware} has a faster convergence rate. Especially when the Hessian can be well approximated by a rank-$k$ matrix, that is $\lambda_{k+1} \ll L$, our algorithm will show great advantages.

\pb\subsection{Global Convergence}\label{subsec:glb}

We will analyze the global convergence property of Algorithm~\ref{alg:zero_order} in this section. To guarantee a global convergence, we have to set a smaller step size compared with the one set in Theorem~\ref{thm:local}. Then, we have the following theorem.
\begin{theorem}\label{thm:glb}
	Let function $f(x)$ satisfy the properties described in Theorem~\ref{thm:local}. For each iteration, the approximate Hessian $\TH_t$ satisfies Eqn.~\eqref{eq:prec_cond}. By choosing the step size $	\eta = \frac{\zeta}{4(d+2)L}$,
	Algorithm~\ref{alg:zero_order} has the following convergence property
	\begin{align*}
	\EB_u\left[f(x_{t+1})-f(x^\star)\right] \leq\left(1-\frac{b\zeta}{16(d+2)\kappa L}\right)\cdot(f(x_t) -f(x^\star))+\Delta_\mu,
	\end{align*}
	where $\Delta_\mu$ is defined as
	\begin{align*}
	\Delta_\mu =\frac{b\mu^2L}{64\zeta(d+2)}\left(2(d+3)^3+\frac{(d+6)^3}{d+2}\right).
	\end{align*}
\end{theorem}

In our analysis of global convergence, we set a fixed step size. We can also use the line search method to get a better convergence property at the cost of extra query to the function value.
\subsection{Query Complexity Analysis}\label{subsec:qca}

In this section, we will analyze the query complexity of \texttt{ZOHA-PW} which implements \texttt{ZO-HessAware} with power-method based Hessian approximation. The power method only needs to access Hessian-Vector product which can be approximated by 
{\small
	\begin{align}
	[\nabla^2f(x) v]_i \approx [H_\mu v]_i \triangleq \frac{f(x+\mu_1 \cdot (v + e_i)) -f(x+\mu_1 \cdot (v- e_i)) + f(x-\mu_1\cdot e_i)-f(x+\mu_1\cdot e_i)}{2\mu_1^2}, \label{eq:H_v_mu}
	\end{align}
}
where $[H_\mu v]_i$ means the $i$-th entry of vector $H_\mu v$. Note that, $H_\mu$ does not need to be explicitly represented. And it can be regarded as the Hessian $\nabla^2f(x)$ with some small perturbations.

Given the above results to approximating Hessian-vector product, we conduct power method to obtain the $k$-largest eigenvalue and their corresponding eigenvectors. The detailed algorithmic procedure is depicted in Algorithm~\ref{alg:lw_app}. Then the approximate Hessian $\TH$ computed based on power method has the following properties.

\begin{algorithm}[tb]
	\caption{Power-method Based Hessian Approximation.}
	\label{alg:lw_app}
	\begin{small}
		\begin{algorithmic}[1]
			\STATE {\bf Input:} Orthonormal matrix $V_0\in\RB^{d\times k}$ where $k$ is the target rank;
			\FOR {$t=0,\dots,T-1$}
			\STATE Approximate the $ \nabla^2f(x)V_t$ by $Y_t = H_\mu V_t$ implemented as Eqn.~\eqref{eq:H_v_mu};
			\STATE QR factorization: $Y_ t= V_{t+1}R_{t+1}$, where $V_{t+1}$ consists of orthonormal columns.
			\ENDFOR
			\STATE Compute $Y = H_\mu V_{T}$ and compute the SVD decomposition $Y = \hat{U}\Lambda \hat{V}^\top$. \label{step:6}
			\STATE {\bf Return:}  $\TH = V\Lambda V^\top + 5\lambda_{k+1} I_d$ with  $V = V_{T}\hat{V}$ 
		\end{algorithmic}
	\end{small}
\end{algorithm}

\begin{theorem}\label{thm:lw_app}
	Let the objective function satisfy Eqn.~\eqref{eq:gamma_1}. Let $\mu_1$ and the iteration number $T$ satisfy that 
	\begin{align*}
	\mu_1 \leq \min\left\{\frac{C_1}{4k\gamma}\lambda_{k+1}, \frac{C_2}{4}\cos(U_k, V_0)\right\} \;\mbox{and} \;\; T = 2C_3\log\left(2\tan(U_k, V_0)\right) 
	\end{align*}
	where $U_k$ is the matrix consists of eigenvectors corresponding to the first $k$ largest eigenvalues of $\nabla^2f(x)$.
	$C_1$, $C_2$, and $C_3$ are absolute constants.
	Then $\TH$ returned from Algorithm~\ref{alg:lw_app} has the following property
	\[
	\frac{\lambda_{\min}}{\lambda_{\min} + 10\lambda_{k+1}}\TH\preceq \nabla^2f(x) \preceq \TH, \;\mbox{and},\;\; 5\lambda_{k+1} I_d \preceq \TH.
	\]
\end{theorem}

The definition of the angle between two spaces $U_k$ and $V_0$ can be found in the textbook \citep{golub2012matrix}. 
Theorem~\ref{thm:lw_app} provides a specific choice of  $\rho = \frac{\lambda_{\min}}{\lambda_{\min} + 10\lambda_{k+1}}$ in Eqn.~\eqref{eq:prec_cond}. Combining the convergence rate in Theorem~\ref{thm:local}, we give the query complexity analysis of \texttt{ZOHA-PW} which implements \texttt{ZO-HessAware} with power-method based Hessian approximation \citep{balcan2016improved}. 
\begin{theorem}\label{thm:query}
	Set the $b = O(d)$ and $p = 1$ in Algorithm~\ref{alg:zero_order}. Then the query complexity of \texttt{ZOHA-PW} is
	\[
	Q(\epsilon) = \tilde{O}\left(\frac{dk\lambda_{k+1}}{\tau }\cdot\log\left(\frac{1}{\epsilon}\right)\right).
	\]
\end{theorem}

%\red{CHECK!!! MISTAKE}

The approximate Hessian $\TH$ constructed using power method can capture the dominant rank-$k$ information of $\nabla^2f(x)$. 
We have the empirical fact that the Hessian of model function can be represented as a rank-$k$ matrix containing the dominant information, plus a perturbation matrix of small spectral norm \citep{yuan2007dimension,bakker2018understanding,sainath2013low}. 
In the case that $k\cdot\lambda_{k+1} < L$, the query complexity of \texttt{ZOHA-PW} is smaller than vanilla zeroth-order methods without Hessian information which takes $O\left(\frac{dL}{\tau }\cdot\log\left(\frac{1}{\epsilon}\right)\right)$ queries shown in the work of \citet{Nesterov2017}.
If the Hessian update frequency $p>1$, then the query complexity of \texttt{ZOHA-PW} can further be reduced.

Furthermore, we can use $V_T$ of the last iteration of Algorithm~\ref{alg:zero_order} as the input $V_0$ of Algorithm~\ref{alg:lw_app}. Because $x_t$ is close to the optimal point $x^\star$, the value of $\tan(U_k, V_0)$ can be regarded as a constant, that is, we can obtain an approximate Hessian in $O(dk)$ query complexity. Hence, the query complexity of \texttt{ZOHA-PW} can be further improved to $O\left(\frac{dk\lambda_{k+1}}{\tau }\cdot\log\left(\frac{1}{\epsilon}\right)\right)$.

\pb\section{Structured Hessian Approximation}\label{sec:H_app}

In this section, we will provide two kinds of \emph{heuristic} methods to construct approximate Hessian. These methods takes much fewer queries to function value compared with power-method based Hessian approximation which takes at least $O(dk)$ queries. The first method is based on Gauss sampling and the second one is based on diagonalization.

\pb\subsection{Gaussian-Sampling Based Hessian Approximation} \label{subsec:Hess_gauss}

We propose a novel method to approximate the Hessian of $f(x)$ with a much lower query complexity than power method. It is based on Gaussian Sampling, and we name \texttt{ZO-HessAware} implemented with such Hessian approximation as \texttt{ZOHA-Gauss}. 

Our new method is going to estimate the Hessian of $f_\mu(x)$ defined in Eqn.~\eqref{eq:f_mu}. Since $\nabla^2f_\mu(x)$ is close to $\nabla^2f(x)$ if $\mu$ is small, a good approximation of $\nabla^2f_\mu(x)$ will also approximate $\nabla^2f(x)$ well. In fact, we can bound the error between $\nabla^2f(x)$ and $\nabla^2f_\mu(x)$ as follows.
\begin{lemma}\label{lem:diff_H}
	Let $f_\mu(x)$ be defined in Eqn.~\eqref{eq:f_mu}. The objective function $f(x)$ satisfies Eqn.~\eqref{eq:gamma_1}. Then, we have
	\[
	\norm{\nabla^2f_\mu(x) - \nabla^2f(x)} \leq \gamma\mu(d+1)^{1/2}.
	\] 
\end{lemma}

Using Gaussian sampling, we can approximate the Hessian of $f_\mu(x)$ as follows:
\begin{align}\label{eq:H_gauss}
\TH = b^{-1} \sum_{i=1}^{b}\frac{f(x+\mu u_i) + f(x-\mu u_i) -2f(x)}{2\mu^2}u_iu_i^\top + \lambda I_d, \;\mbox{with}\; u_i\sim N(0, I_d)
\end{align}
where $\lambda$ is a properly chosen regularizer to keep $\TH$ invertible. In the construction of $\TH$ of Eqn.~\eqref{eq:H_gauss}, we only take a small batch of points, that is $b$ is small even much smaller the dimension $d$. Hence, the construction of such $\TH$ has a low query complexity.

The approximate Hessian $\TH$ constructed as Eqn.~\eqref{eq:H_gauss} has the following property. 
\begin{lemma}\label{lem:H_mu}
	Let $\TH$ be an approximate Hessian defined in Eqn.~\eqref{eq:H_gauss}. Function $f_\mu(x)$ is the smoothed function defined in Eqn.~\eqref{eq:f_mu}. Then $\TH$ has the following property
	\begin{align*}
		\nabla^2f_\mu(x)\preceq \EB_u[\TH] = \nabla^2f_\mu(x) + \left(\lambda - \frac{f(x) - f_\mu(x) }{\mu^2}\right) \cdot I_d
	\end{align*}
\end{lemma}

Combining Lemma~\ref{lem:diff_H} and~\ref{lem:H_mu}, we can obtain the result that if $\mu$ is small, and the batch size $b$ in Eqn.~\ref{eq:H_gauss} is large, then $\TH$ constructed as Eqn.~\eqref{eq:H_gauss} can approximate $\nabla^2f(x)$  well. In practice, we prefer to choosing  a small batch size $b$ to achieve query efficiency.

\pb\subsection{Diagonalization Based Hessian Approximation} \label{subsec:diag}

We propose to use a diagonal matrix to approximate the Hessian. This method has been used in the optimization of deep neural networks \citep{kingma2015adam,zeiler2012adadelta,tieleman2012lecture} and online learning \citep{duchi2011adaptive}.

First, we compute an  approximate Hessian in the manner of \texttt{ADAM} \citep{kingma2015adam} as follows:
\begin{equation}\label{eq:adam}
	\begin{split}
	\ti{g}_\mu(x_{t-1}) =& \frac{1}{b} \sum_{i=1}^{b}\frac{f(x_{t-1}+\mu \ti{u}_i)-f(x_{t-1})}{\mu}\ti{u}_i,\;\mbox{with}\; \ti{u}_i \sim N(0, \TH_{t-1}^{-1})\\
	D_{t} =& \nu D_{t-1} + (1-\nu) \ti{g}_\mu^2(x_{t-1})\\
	\TH_t =& \diag\left(\frac{D_t}{1 - \nu^t}\right)
	\end{split}
\end{equation}
with $0\leq\nu\leq 1$. And $\ti{g}_\mu^2(x)$ means the entry-wise square of $\ti{g}_\mu(x)$.

Second, we can also use the method of \texttt{ADAGRAD} \citep{duchi2011adaptive} to construct the approximate Hessian as
\begin{align*}
\ti{g}_\mu(x_{t-1}) =& \frac{1}{b} \sum_{i=1}^{b}\frac{f(x_{t-1}+\mu \ti{u}_i)-f(x_{t-1})}{\mu}\ti{u}_i,\;\mbox{with}\; \ti{u}_i \sim N(0, \TH_{t-1}^{-1})\\
	D_t =& D_{t-1} + \ti{g}_\mu^2(x_{t-1})\\
	\TH_t =&\diag\left(\frac{D_t}{n}\right).
\end{align*}
Other methods of constructing diagonal Hessian approximation such as \texttt{ADADELTA}~\citep{zeiler2012adadelta} used in training deep neural networks can also be use to in our diagonal Hessian approximation.

These kinds of Hessian approximation are heuristic. We can not give an exact convergence rate of \texttt{ZO-HessAware} with diagonal Hessian approximation by Theorem~\ref{thm:local}. However, diagonal Hessian approximations have shown their power in training deep neural networks. Furthermore, diagonal approximate Hessian has an important advantage that it does not need extra queries to the function value and need less computational and storage cost.

Though the construction procedure of the diagonal Hessian approximation is the same with the one of \texttt{ADAM} and \texttt{ADAGRAD}, there some difference between these diagonal Hessians. First, \texttt{ADAM} and \texttt{ADAGRAD} use $\TH^{1/2}$ as the approximate Hessian in training neural network which is different from our approximate Hessian. Second, in the construction of our diagonal Hessian, we use the `natural gradient' defined in Eqn.~\eqref{eq:t_gmu} which contains the Hessian information other than the ordinary gradient. And the information of the current diagonal Hessian will be used in the estimation of next `natural gradient'. In contrast, the diagonal Hessian will not affect the computation of gradients. 

\pb\section{Experiments}
\label{sec:experiments}
In this section, we apply our Hessian-aware zeroth-order algorithm to the black-box adversarial attacks.  This is an important research topic in security of deep learning because neural networks are widely used in image classification. However, current neural network-based classifiers are susceptible to adversarial examples.

Our adversarial attack experiments include both targeted attack and un-targeted attack. 
The targeted attack aims to find an adversarial example $x$ of a given image $x_0$ with a targeted class label $\ell$ toward misclassification. 
In this case, we are going to minimize  the following problem proposed in the work of \citet{carlini2017towards}:
\begin{equation}\label{eq:tar}
\small
\begin{split}
&f ( x , \ell ) = \max \left\{ \max _ { i \neq \ell } \log [ Z ( x ) ] _ { i } - \log [ Z ( x ) ] _ { \ell } , - \omega \right\},
\end{split}
\end{equation}
with constraint that
\begin{equation}\label{eq:constrain}
\norm{x - x_0}_\infty \leq \varepsilon.
\end{equation}
$Z(x)$ is the logit layer representation after softmax in the DNN for x such that $[Z(x)]_i$ represents the predicted probability that x belongs to class $i$. $\omega$ is a tuning parameter for attack transferability and we set  $\omega = 1$ in our experiments. The constraint means that the adversarial image should be close to the given image.

The un-targeted adversarial attack aims to find an example $x$ of the given image $x_0$ with label $\ell$ but misclassified by the neural network. In this case, we will minimize the following function \citep{carlini2017towards} with constraint~\eqref{eq:constrain}:
\begin{equation}
\small
f(x) = \max \left\{ \log [ Z ( x ) ] _ { \ell } - \max _ { i \neq \ell } \log [ Z ( x ) ] _ { i } , - \omega \right\}. \label{eq:untar}
\end{equation}

\pb\subsection{Algorithm Implementation}

In the experiments, we will implement \texttt{ZO-HessAware} (Algorithm~\ref{alg:zero_order}) with two different kinds of Hessian approximation. The first one is based on the Gaussian sampling described in Section~\ref{subsec:Hess_gauss}, and we call it \texttt{ ZOHA-Gauss}. 
The second implementation is using the diagonal Hessian approximation described in Section~\ref{subsec:diag} with the update procedure as \texttt{ADAM} defined in Eqn.~\eqref{eq:adam}. And we name it as \texttt{ZOHA-Diag}. We do not implement \texttt{ZO-HessAware} with other kinds of diagonal Hessian approximation because these methods have the similar performance.

Furthermore, because the adversarial problem is of constraint, we will modify the update step~\eqref{step:update} of Algorithm~\ref{alg:zero_order} as follows:
\begin{align}
	x_{t+1} = \Pi[x_t - \eta \ti{g}_\mu(x_t)],\label{eq:project}
\end{align}
where $\Pi[\cdot]$ is a projection operator to make $x_{t+1}$ satisfy $\norm{x_{t+1} - x_0}_\infty \leq \varepsilon$. Note that, this projection is exact for \texttt{ZOHA-Diag}. But as to \texttt{ZOHA-Gauss}, we should compute $x_{t+1}$ by optimizing the following sub-problem:
\begin{align} \label{eq:sub-prob}
x_{t+1} = \argmin_{y\in[x_0 - \epsilon, x_0 +\epsilon]}\norm{y - \left(x_t - \eta \TH_t^{-1}g_\mu(x_t)\right)}_{\TH}^2.
\end{align}
However, the projection as Eqn.~\eqref{eq:project} performs well and is of simple implementation even it is just an approximation to the true one computed by Eqn.~\eqref{eq:sub-prob}.

Because the objective function of the neural network model may be non-convex, we implement the approximate Hessian in \texttt{ZOHA-Gauss} as follows:
\begin{align*}
\TH = b^{-1} \sum_{i=1}^{b}\frac{\left|f(x+\mu u_i) + f(x-\mu u_i) -2f(x)\right|}{2\mu^2}u_iu_i^\top + \lambda I_d, \;\mbox{with}\; u_i\sim N(0, I_d).
\end{align*}
Such modification ensures that such $\TH$ is positive definite. Furthermore, we observe that $\TH$ can be written as $\TH = CC^\top + \lambda I$. Then we can compute $\TH^{-1/2}$ as follows. First, we compute the SVD decomposition of $C$ as $C = U_C \Lambda_C U_C^\top$ with $U_C\in\RB^{d\times b}$ and $\Lambda_C\in\RB^{b\times b}$. 
And we get $\TH^{-1/2} = U_C\left((\Lambda_C^2 + \lambda I)^{-1/2} - \lambda^{-1/2} I\right) U_C^\top + \lambda^{-1/2} I$. In practice, the value of $\lambda$ can be set as a fraction of $\norm{CC^\top}$ or tuned by several tries.

\begin{algorithm}[tb]
	\caption{Algorithm \texttt{ZO-HessAware} with descent checking (\texttt{ZOHA-DC})}
	\label{alg:zo-dc}
	\begin{small}
		\begin{algorithmic}[1]
			\STATE {\bf Input:} $x^{(0)}$ is an initial point sufficient close to $x^{*}$. And $b$ is the batch size and $p$ is an integer. Parameter $\eta$ is the step size. $\beta$ is the threshold of sample size in descent checking. $\delta_b$ is the parameter of the sample size increment.
			\FOR {$t=0,\dots,T$ }
			\IF {$t\bmod p == 0$}
			\STATE Compute an approximate Hessian $\TH_t$ satisfies Eqn.~\eqref{eq:prec_cond}.
			\ENDIF
			\STATE Generate $b$ samples with $u_i\sim N(0, I_d)$ and construct $\ti{g}_\mu(x_t) = \frac{1}{b}\sum_{i=1}^b \frac{f(x+ \TH_t^{-1/2} u_i) -f(x)}{\mu} \TH_t^{-1/2} u_i$;
			\STATE Compute $y_{t+1} = x_t- \eta \ti{g}_\mu(x_t)$ and set $N_b = b$.
			\WHILE {$f(y_{t+1}) > f(x_t)$ and $N_b< \beta$}
			\STATE Generate another $\delta_b$ samples with $u_i\sim N(0, I_d)$ and set $N_b = N_b+\delta_b$;
			\STATE Construct $\ti{g}_\mu(x_t) = \frac{1}{N_b}\sum_{i=1}^{N_b} \frac{f(x+ \TH_t^{-1/2} u_i) -f(x)}{\mu} \TH_t^{-1/2} u_i$ ;
			\STATE Compute $y_{t+1} = x_t- \eta \ti{g}_\mu(x_t)$.
			\ENDWHILE
			\STATE Update $x_{t+1} = x_t- \eta \ti{g}_\mu(x_t)$.
			\ENDFOR
		\end{algorithmic}
	\end{small}
\end{algorithm}	

\subsubsection{Descent Checking}
To improve the success rate and the query efficiency, we introduce an important technique called \texttt{Descent-Checking} which has been discussed in Remark~\ref{rk:check} but with a slight different implementation. \texttt{Descent-Checking} has the following algorithmic procedure. After obtaining the $x_{t+1}$, we will query to the value $f(x_{t+1})$ and check if $f(x_{t+1}) \leq f(x_t)$. If it holds, we will go to the next iteration. Otherwise, we will discard current $x_{t+1}$ and take extra $\delta_b$ samples combining with existing samples to estimate a new gradient until a new $x_{t+1}$ satisfies $f(x_{t+1}) \leq f(x_t)$ or the total sample size exceeds a threshold. If the total sample size exceeds the threshold, we will accept this `bad' $x_{t+1}$ and go to the next iteration. Because we will often set $b$ be of several tens, \texttt{Descent-Checking} strategy will not bring many extra queries. As a result of \texttt{Descent-Checking}, we can filter some bad search direction effectively which will lead to a higher attack success rate. We depict the detailed algorithmic procedure of \texttt{ZO-HessAware} with \texttt{Descent-Checking} in Algorithm~\ref{alg:zo-dc}. Accordingly, we name \texttt{ZOHA-Gauss} and \texttt{ZOHA-Diag} with \texttt{Descent-Checking} strategy as \texttt{ZOHA-Gauss-DC} and \texttt{ZOHA-Diag-DC}, respectively.

\begin{figure}[]
	\subfigtopskip = 0pt
	\begin{center}
		\centering
		\subfigure[\textsf{\texttt{ZOO}}]{\includegraphics[width=56mm]{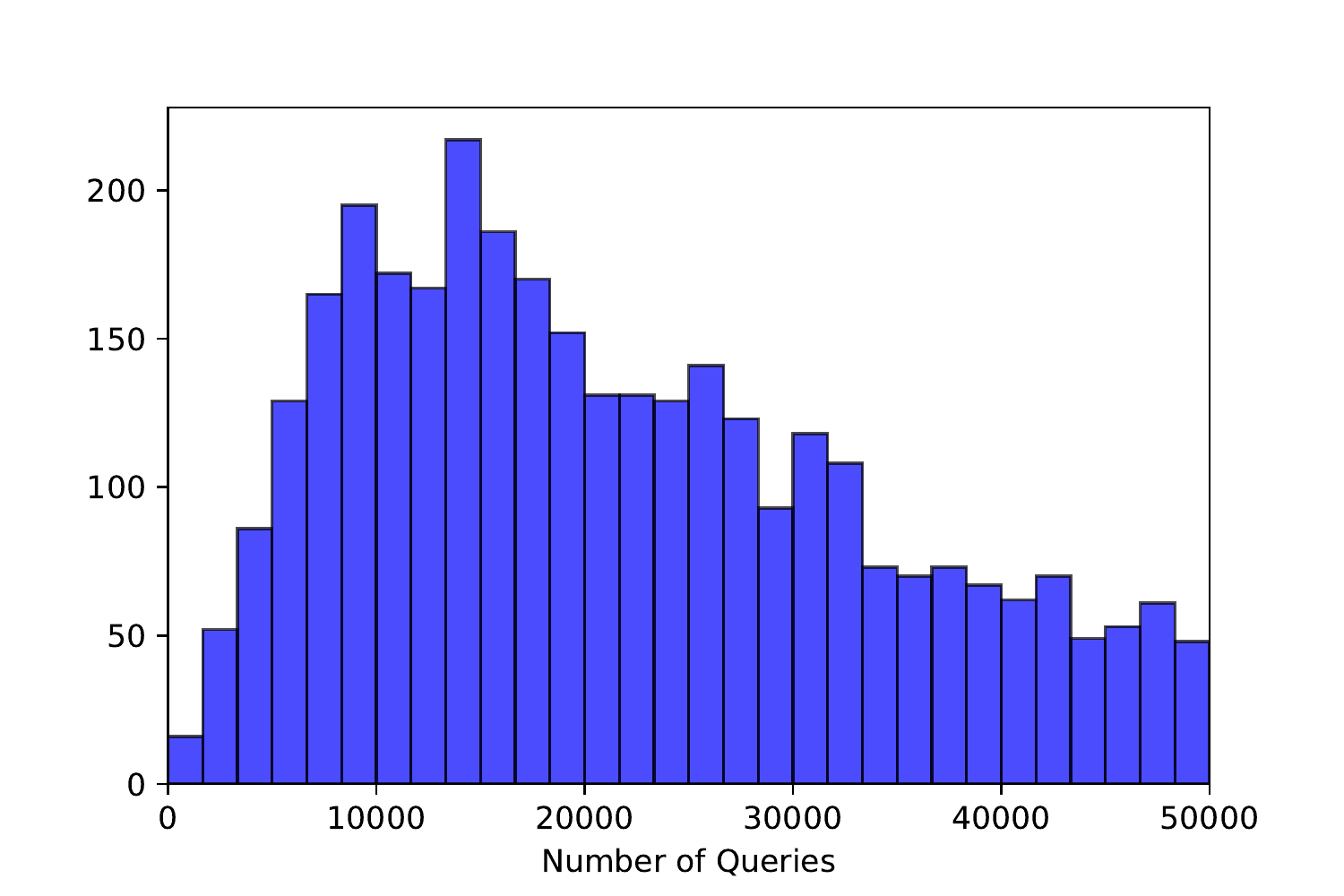}}~
		\subfigure[\textsf{ \texttt{ZOHA-Gauss}}]{\includegraphics[width=56mm]{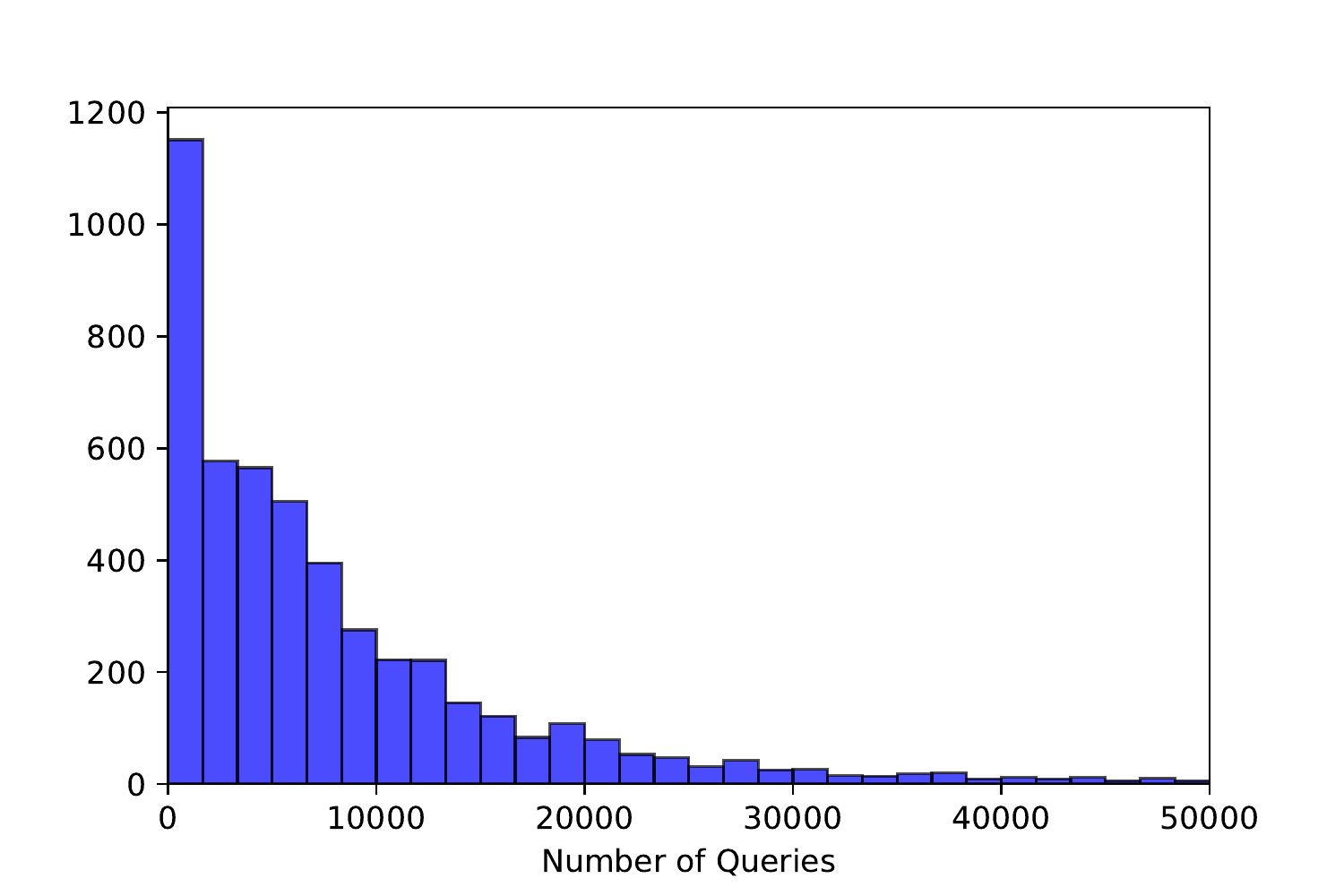}}~
		\subfigure[\textsf{\texttt{ZOHA-Diag}}]{\includegraphics[width=56mm]{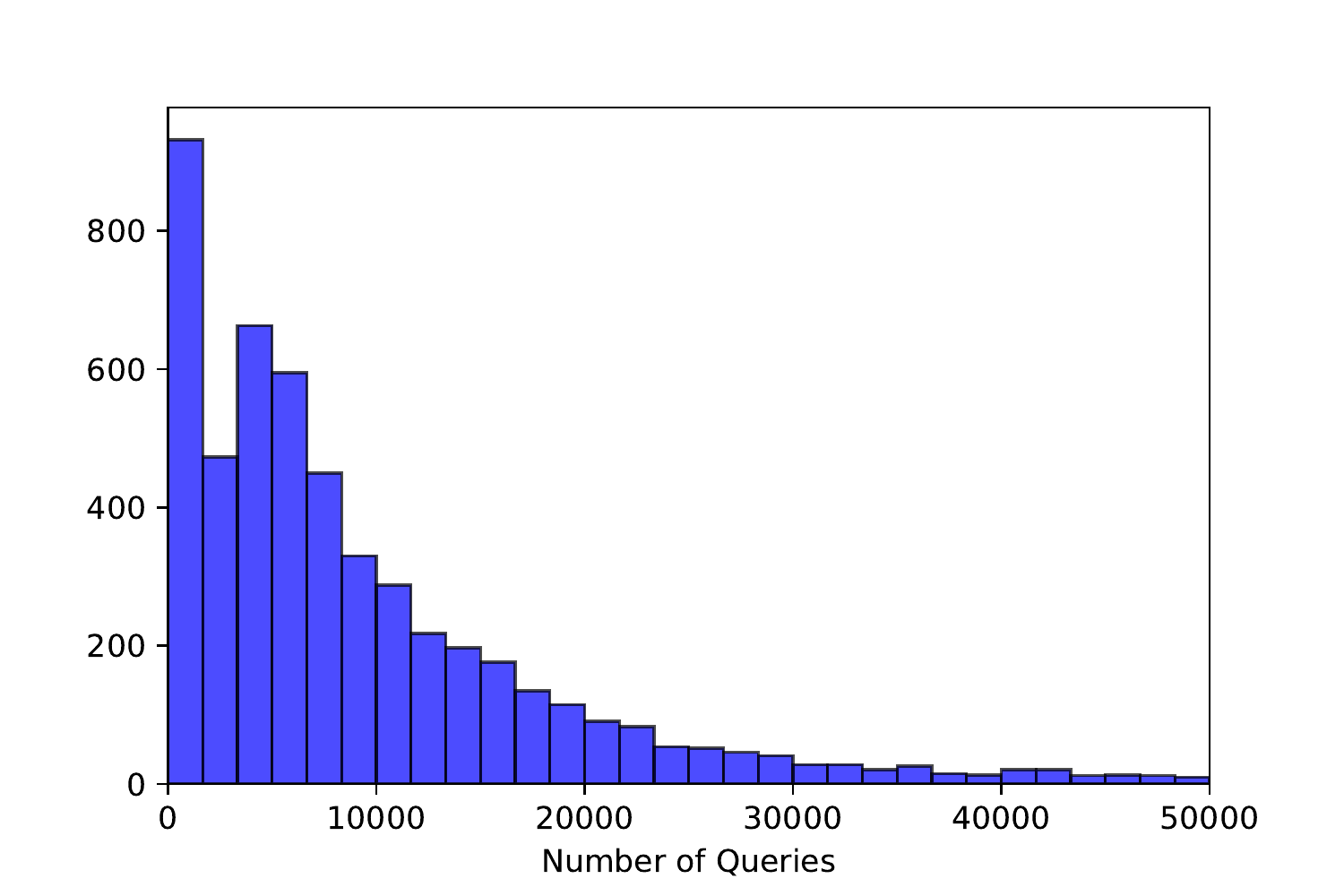}}\\
		\subfigure[\textsf{\texttt{PGD-NES}}]{\includegraphics[width=56mm]{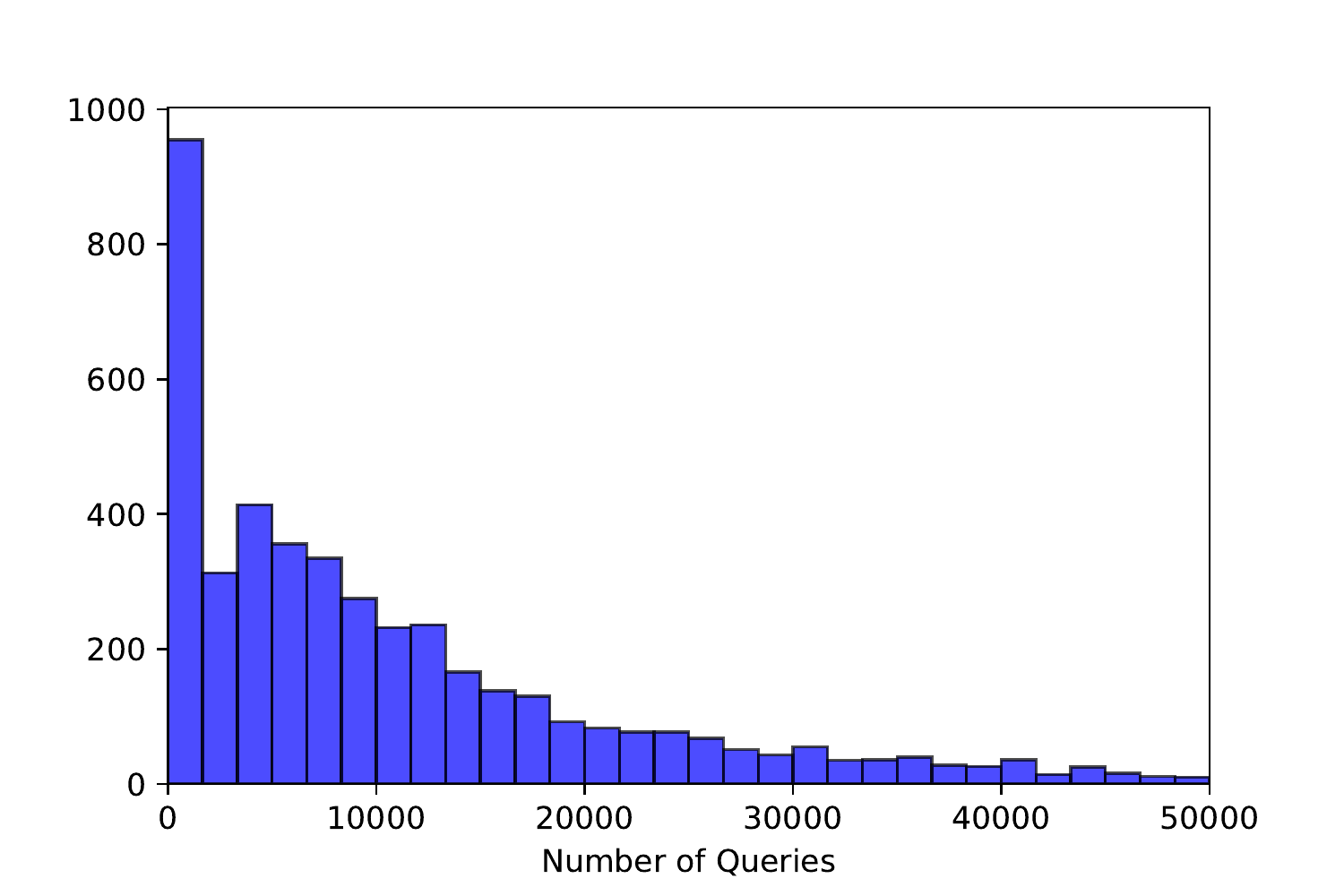}}~
		\subfigure[\textsf{ \texttt{ZOHA-Gauss-DC}}]{\includegraphics[width=56mm]{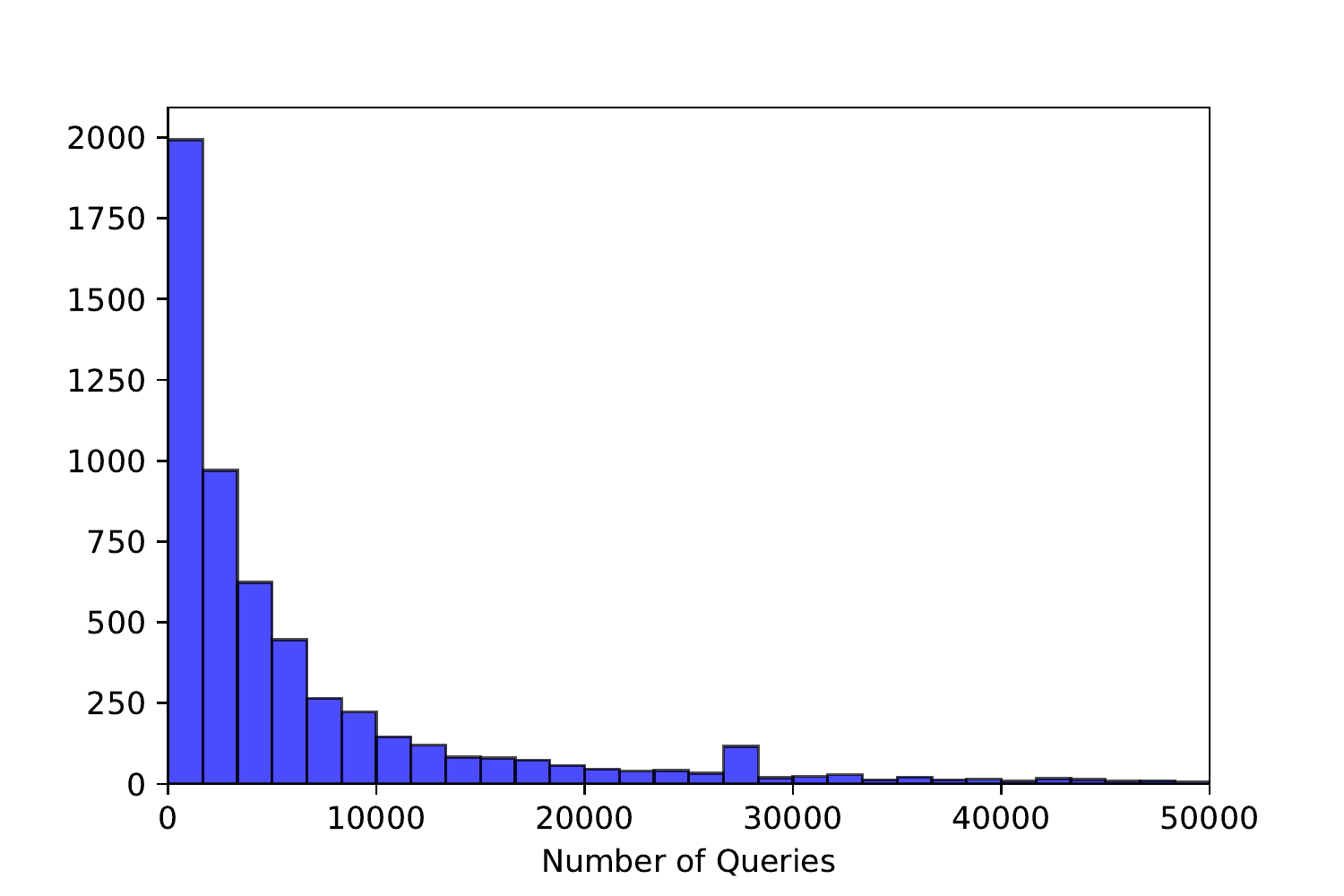}}~
		\subfigure[\textsf{\texttt{ZOHA-Diag-DC}}]{\includegraphics[width=56mm]{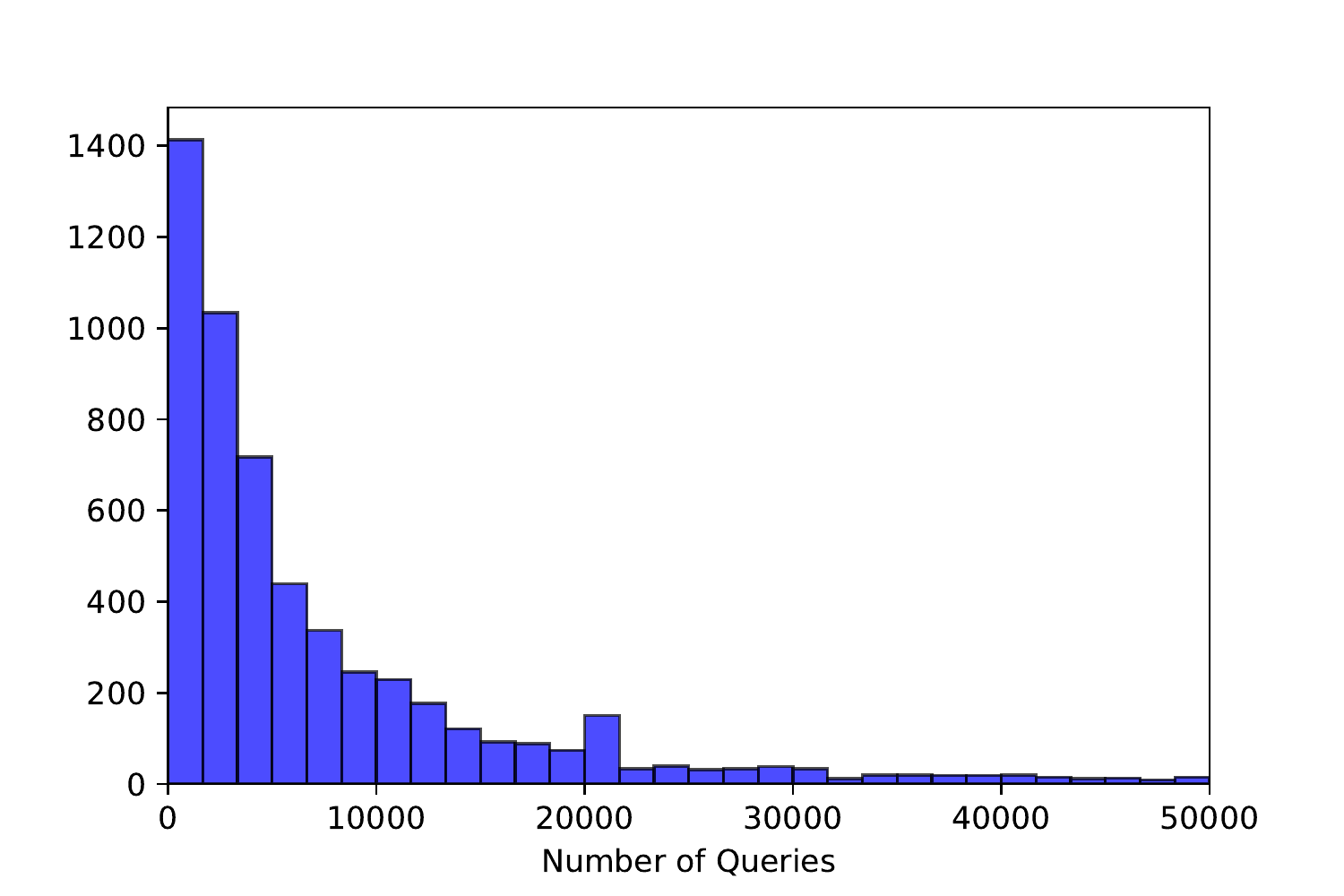}}
	\end{center}
	\caption{The distribution of the number of queries on \emph{targeted} black-box attacks on CNN model and MNIST}
	\label{fig:mnist_tar}
\end{figure}

\begin{figure}[!ht]
	\subfigtopskip = 0pt
	\begin{center}
		\centering
		\subfigure[\textsf{\texttt{ZOO}}]{\includegraphics[width=56mm]{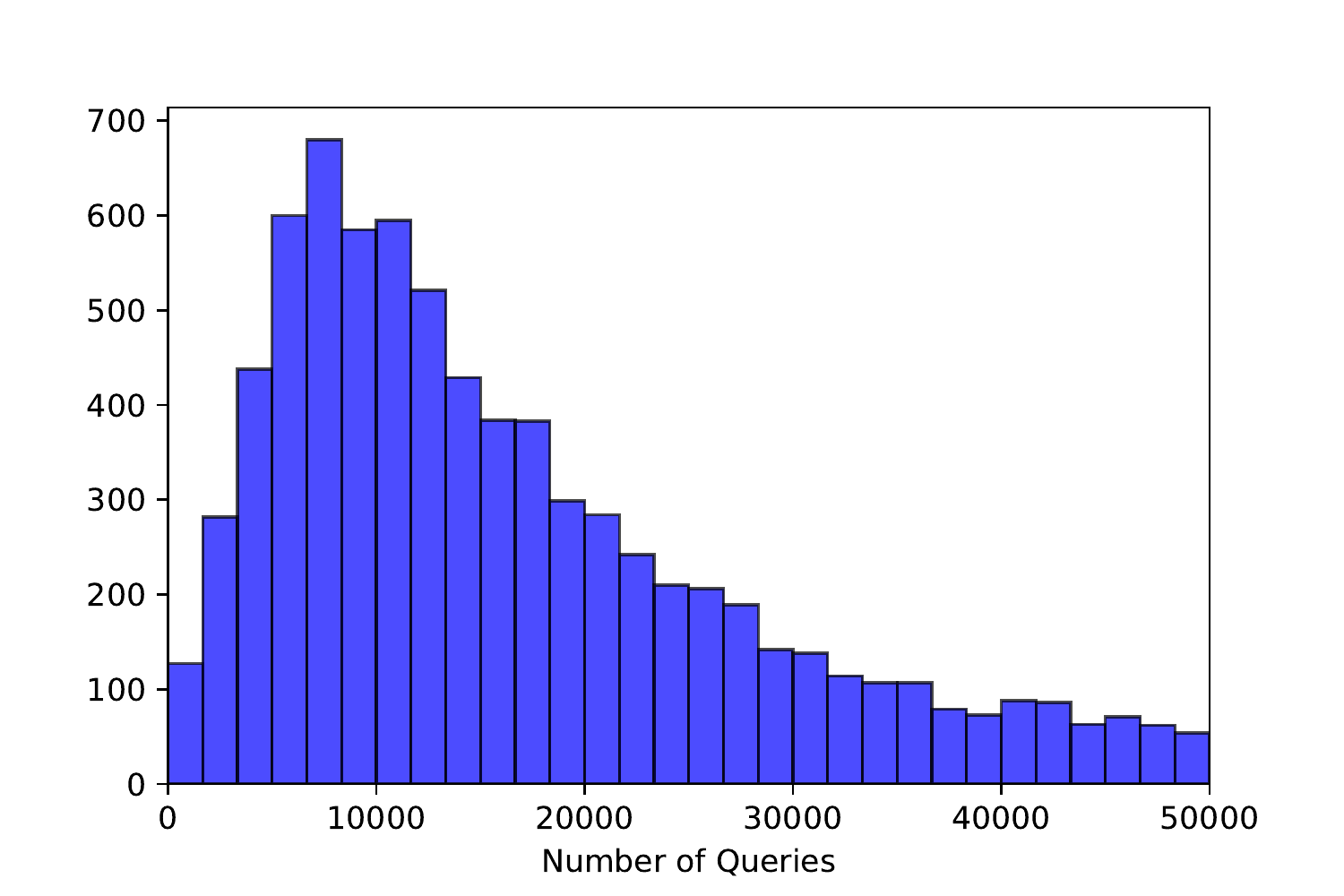}}~
		\subfigure[\textsf{ \texttt{ZOHA-Gauss}}]{\includegraphics[width=56mm]{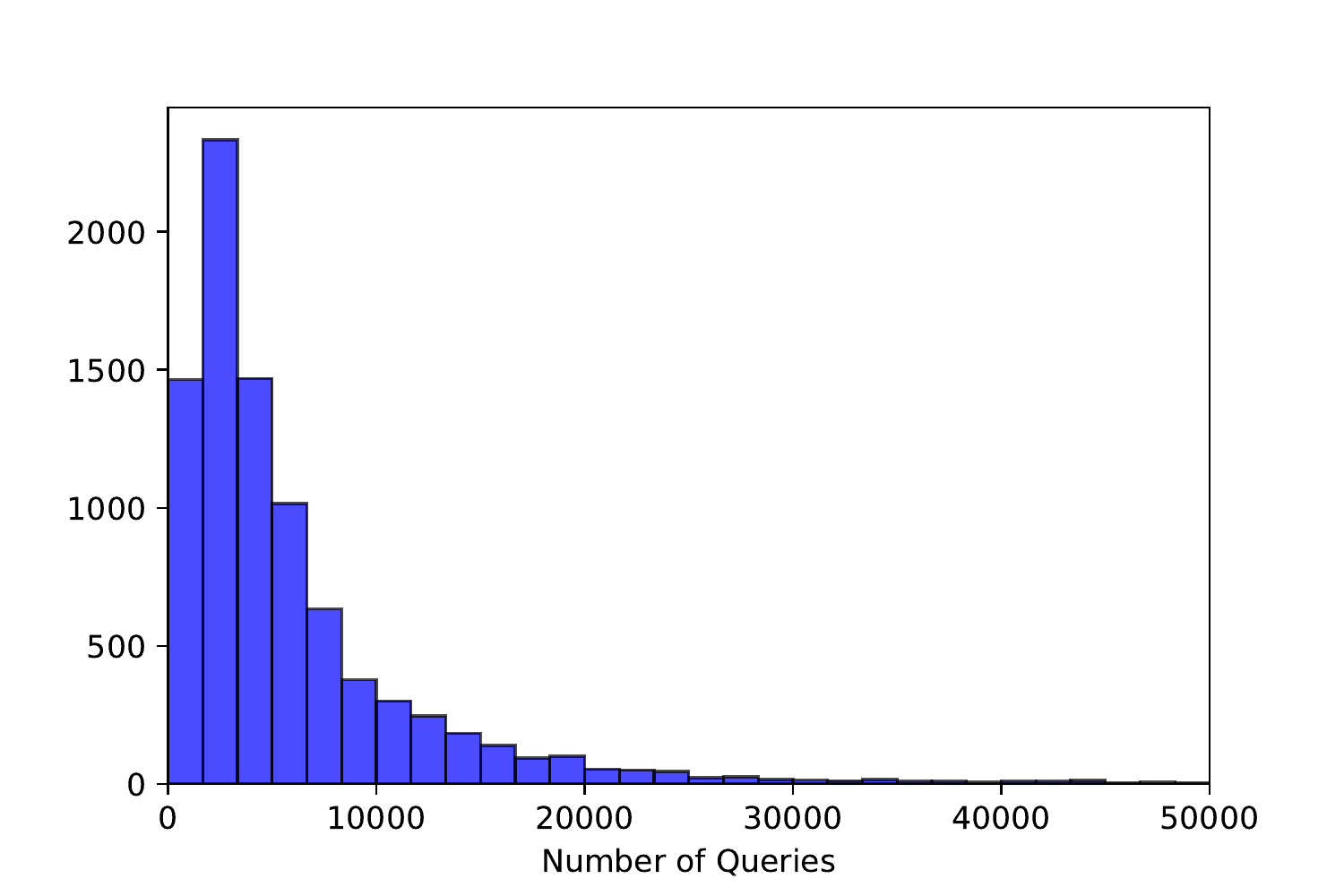}}~
		\subfigure[\textsf{\texttt{ZOHA-Diag}}]{\includegraphics[width=56mm]{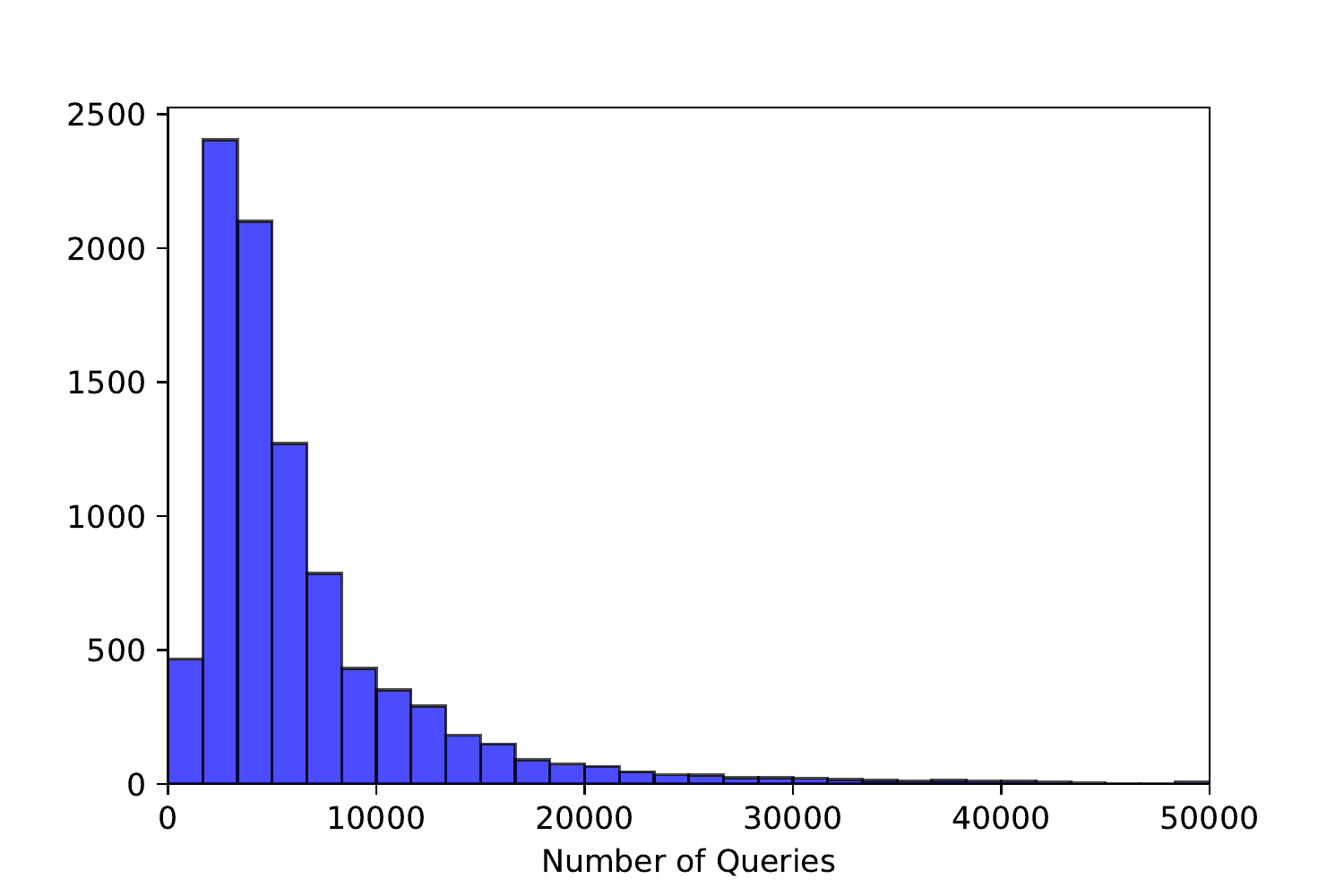}}\\
		\subfigure[\textsf{\texttt{PGD-NES}}]{\includegraphics[width=56mm]{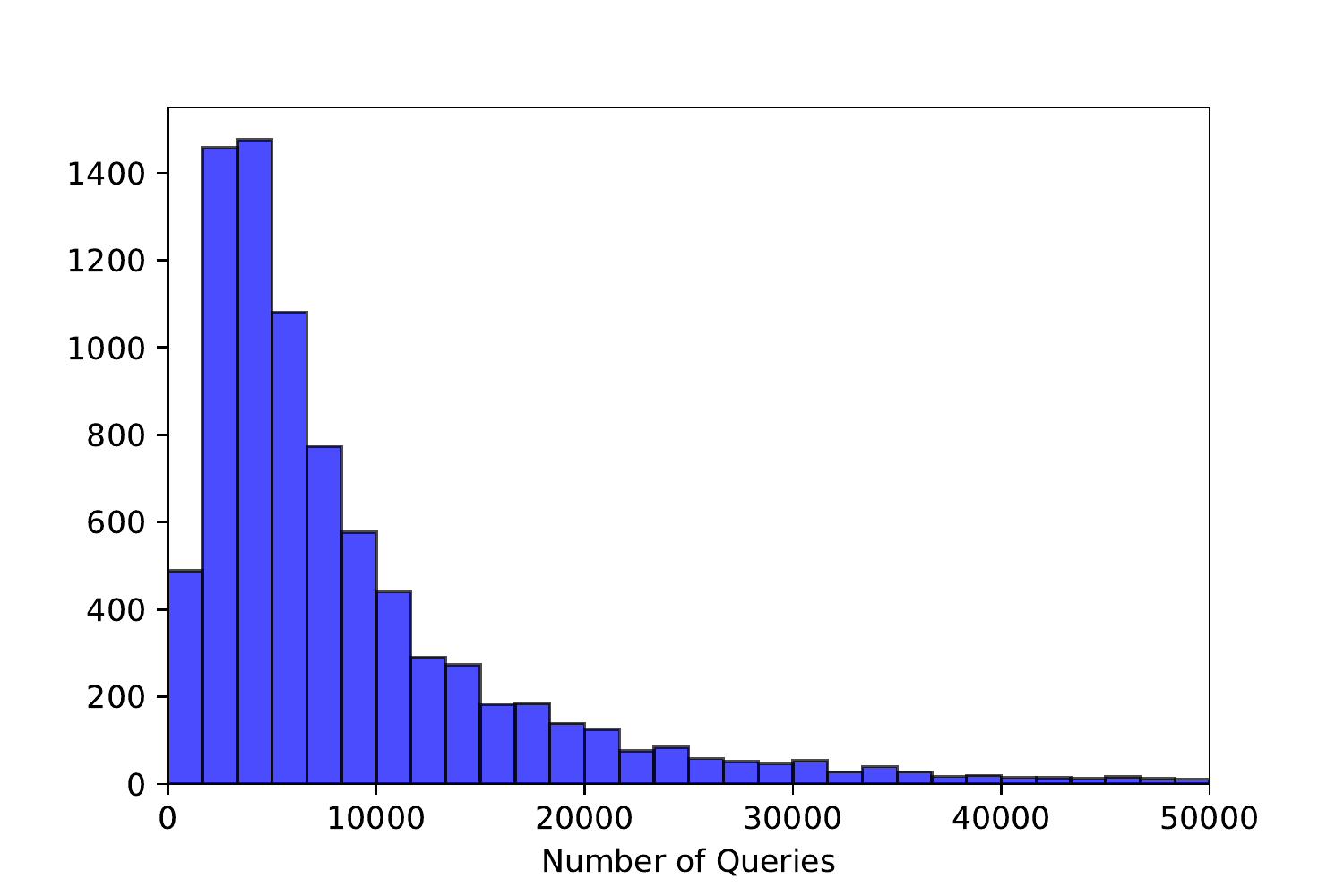}}~
		\subfigure[\textsf{ \texttt{ZOHA-Gauss-DC}}]{\includegraphics[width=56mm]{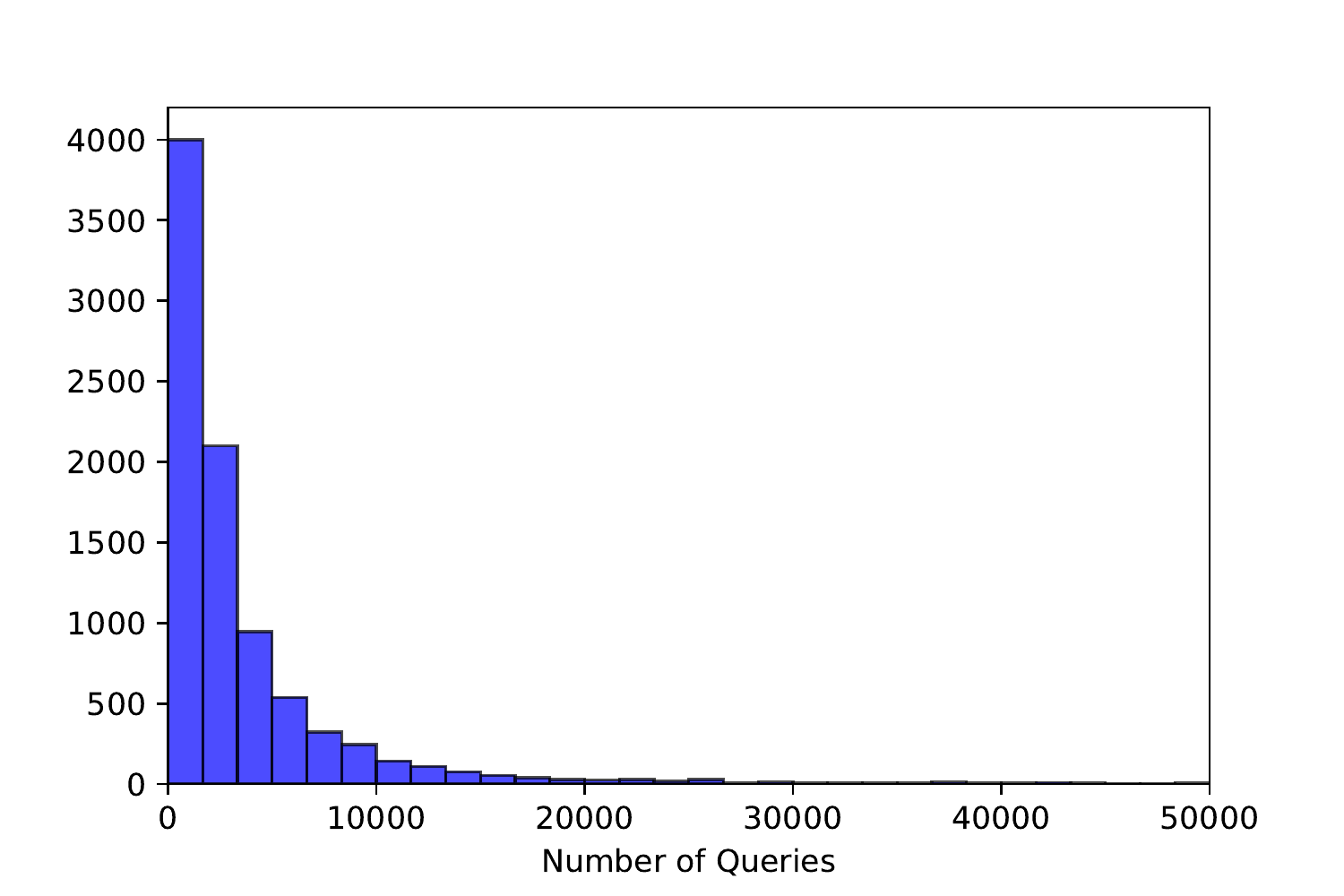}}~
		\subfigure[\textsf{\texttt{ZOHA-Diag-DC}}]{\includegraphics[width=56mm]{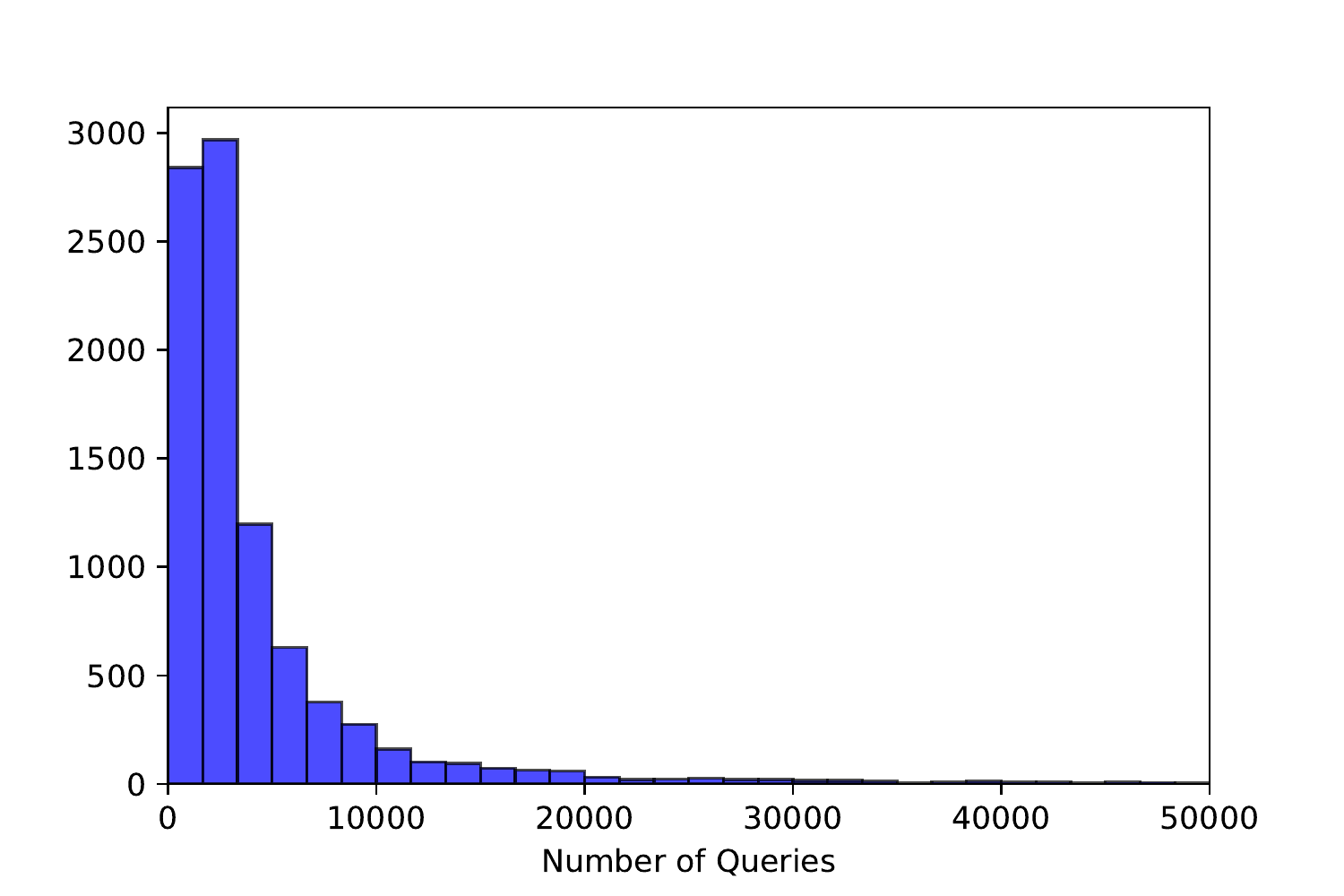}}
	\end{center}
	\caption{The distribution of the number of queries on \emph{un-targeted} black-box attacks on CNN model and MNIST}
	\label{fig:mnist_untar}
\end{figure}

\pb\subsection{Evalution on MNIST}

We evaluate the effectiveness of our attacks against an convolution neural network (CNN) on the MNIST dataset. The network for MNIST is composed of two 5$\times$5 convolutional layers with output $16$ and $64$ channels following two fully connected layers with $128$ and $10$ units. We use 2$\times$2 max-pooling after each convolutional layer and use ReLU after every layer expect the layer. The network is trained for 100 epochs with learning rate starting at 0.1 and decay 0.5 every 20 epochs. The accuracy of the model is $98.95\%$.

We test the attack algorithms on $10000$ images from the test set. 
The limit of $\ell_\infty$ perturbation is $\varepsilon = 0.2$. 
We run all the attack until getting the an adversarial examples unless the number of queries is more than $50,000$. The parameter settings of algorithms is listed in Table~\ref{tb:mnist_para} in Appendix. Furthermore, in \texttt{ZOHA-Gauss} and \texttt{ZOHA-Gauss-DC}, we update approximate Hessian once every $p = 20$ iterations and $\mu = 0.5$ in Eqn.~\eqref{eq:H_gauss}. To make a fair comparison, we do not add `momentum' which is an effective technique to reduce queries to all algorithms including \texttt{ZOHA}-type algorithms, \texttt{PGD-NES}, and \texttt{ZOO}.  

We report the experiment results in Table~\ref{tb:mnist} and Figure~\ref{fig:mnist_tar} and \ref{fig:mnist_untar}. The visualization of adversarial attack is present in Appendix~\ref{app:vis_att}. We can observe that \texttt{ZO-HessAware} with different implementations obtain much better success rates than two state-of-the-art algorithms. This validates the effectiveness of second-order information of model function in zeroth-order optimization. Especially, our \texttt{ZOHA-DC} type algorithms obtain the best success rates both target and un-target adversarial attacks which are much higher than \texttt{ZOO} and \texttt{PGD-NES} while taking less queries to function values. Furthermore, on the un-target attack, \texttt{ZOHA-DC} type algorithms only take less than half of queries of \texttt{PGD-NES}. This greatly shows the query efficiency of our Hessian-aware zeroth-order algorithm.

The comparison of the distribution of the query number in Figure~\ref{fig:mnist_tar} and~\ref{fig:mnist_untar} shows that \texttt{Descent Checking} technique can reduce the query number effectively. For example, comparing \texttt{ZOHA-Gauss} with \texttt{ZOHA-Gauss-DC} in Figure~\ref{fig:mnist_untar}, we can observe that the percentage of the query number between $0$ and $2000$ of \texttt{ZOHA-Gauss-DC} is much higher than the one of \texttt{ZOHA-Gauss}.

\begin{table*}
	\centering
	\caption{Comparison of $\ell_\infty$ norm based black-box attacks on CNN model and MNIST with $\varepsilon = 0.2$}
	\label{tb:mnist}
	\begin{tabular}{c|lccc}
		\hline
		& Algorithm & success rate $\%$ & median queries & average queries\\ \hline
		\multirow{6}*{targeted} 
		&\texttt{ZOO} \citep{chen2017zoo} & 42.13 & 15,200  &17,091 \\
		&\texttt{PGD-NES} \citep{ilyas18a} & 44.19 & 7,300  &10,496 \\
		&\texttt{ZOHA-Gauss}         & 50.03 & 3,712 & 6,649 \\
		&\texttt{ZOHA-Gauss-DC}         &\bfseries 56.14 & \bfseries 2,941 & \bfseries 6,246 \\
		&\texttt{ZOHA-Diag}         & 52.13& 6,400  & 9,128 \\
		&\texttt{ZOHA-Diag-DC}      & 55.56   & 3,936 & 7,239 \\\hline
		\multirow{6}*{un-targeted} 
		&\texttt{ZOO} \citep{chen2017zoo} & 77.18 & 13,300&16,390 \\
		&\texttt{PGD-NES} \citep{ilyas18a} & 81.55 & 5,800 &8,567 \\ 
		&\texttt{ZOHA-Gauss}         & 85.06& 3,612 &5,000 \\
		&\texttt{ZOHA-Gauss-DC}         & 88.80& \bfseries2,152 &\bfseries3,629 \\
		&\texttt{ZOHA-Diag}         & 90.37      & 4,500 & 6,439 \\
		&\texttt{ZOHA-Diag-DC}       & \bfseries 91.90 & 2,460 & 4,352 \\\hline
	\end{tabular}
\end{table*}

\pb\subsection{Evaluation on ImageNet}

In this experiment, we use a pre-trained ResNet50 that has $78.15\%$ top-$1$ accuracy and $92.87\%$ top-$5$ accuracy for evaluation. 
The limit of $\ell_\infty$ perturbation is $\varepsilon = 0.05$. 
We will choose $1,000$ images randomly from ImageNet test-set for evaluation and run the attack method until getting an adversarial example or the number of queries being more than $1,000,000$. 
Furthermore, if the attack is targeted, the target label will be randomly chosen from $1,000$ classes. The parameter settings of algorithms is listed in Table~\ref{tb:imagenet_para} in Appendix. Furthermore, in \texttt{ZOHA-Gauss} and \texttt{ZOHA-Gauss-DC}, we update approximate Hessian once every $p = 20$ iterations and $\mu = 0.5$ in Eqn.~\eqref{eq:H_gauss}. To make a fair comparison, we do not add `momentum' which is an effective technique to reduce queries to all algorithms including \texttt{ZOHA}-type algorithms, \texttt{PGD-NES}, and \texttt{ZOO}.  

In the experiment on ImageNet, instead of Eqn.~\eqref{eq:g_mu},  we use the following method to estimate the gradient
\begin{align*}
g_\mu(x) = \frac{1}{b}\sum_{i=1}^{b}\frac{f(x+\mu \TH^{-1/2}u_i)-f(x-\mu \TH^{-1/2}u_i)}{2\mu}\TH^{1/2}u_i.
\end{align*}

We can see that such $g_\mu$ have the same expectation with the one defined in Eqn.~\eqref{eq:g_mu}. However, it has a better performance in this experiment. Accordingly, the natural gradient $\ti{g}_\mu(x)$ is modified similarly as 
\begin{align*}
\ti{g}_\mu(x) = \frac{1}{b} \sum_{i=1}^{b}\frac{f(x+\mu \ti{u}_i)-f(x-\mu \ti{u}_i)}{2\mu}\ti{u}_i,\;\mbox{with}\; \ti{u}_i \sim N(0, \TH^{-1}).
\end{align*}

We report the results in Table~\ref{tb:imagenet} and Figure~\ref{fig:iamgenet_tar},~\ref{fig:iamgenet_untar}. The visualization of adversarial attack is present in Figure~\ref{fig:imagenet_tar} and~\ref{fig:imagenet_untar} of Appendix~\ref{app:vis_att}. We can observe that our algorithms take much less queries than \texttt{ZOO} and \texttt{PGD-NES}. For the un-target attack, the median queries of \texttt{ZOHA-Diag-DC} is only about $5\%$ of \texttt{ZOO} and about $38\%$ of \texttt{PGD-NES} with the same success rate. For the target attack, compared with the un-target attack problem, all these algorithms take much more queries. But our algorithms still show great query efficiency. Especially, both \texttt{ZOHA-Diag} and \texttt{ZOHA-Diag-DC} achieve $100\%$ attack success rate which is higher than \texttt{PGD-NES} but only with about $50\%$ queries of \texttt{PGD-NES}. Though \texttt{ZOO} also obtain a $100\%$ success rate, it takes several times of queries as \texttt{ZOHA-Diag} and \texttt{ZOHA-Diag-DC}.

\begin{table*}
	\centering
	\caption{Comparison of $\ell_\infty$ norm based black-box attacks on ResNet50 model and ImageNet with $\varepsilon = 0.05$}
	\label{tb:imagenet}
	\begin{tabular}{c|lccc}
		\hline
		& Algorithm & success rate $\%$ & median queries & average queries\\ \hline
		\multirow{6}*{targeted} 
		&\texttt{ZOO} \citep{chen2017zoo}			& \bfseries 100		& 39,100&45,822 \\
		&\texttt{PGD-NES} \citep{ilyas18a} 			& 99.37			 	& 11,270&17,435 \\
		&\texttt{ZOHA-Gauss}        			& 99.62			 	& 8,748	&12,257 \\
		&\texttt{ZOHA-Gauss-DC}         	& 100				& 8,588 & 11,770 \\
		&\texttt{ZOHA-Diag}         		&\bfseries 100		& 7,400 	& 9,123 \\
		&\texttt{ZOHA-Diag-DC}         	&\bfseries 100& \bfseries 6,273 & \bfseries 8,574 \\\hline
		\multirow{4}*{un-targeted} 
		&\texttt{ZOO} \citep{chen2017zoo} & 100& 12,700&14,199 \\
		&\texttt{PGD-NES} \citep{ilyas18a} & 100& 1,500 &2,283 \\ 
		&\texttt{ZOHA-Gauss}         & 100& 1,212 &2,259 \\
		&\texttt{ZOHA-Gauss-DC}        & 100& 1,124 &1,959 \\
		&\texttt{ZOHA-Diag}          & 100& 800 & 1,149 \\
		&\texttt{ZOHA-Diag-DC}        & 100& \bfseries 561 & \bfseries 945\\\hline
	\end{tabular}
\end{table*}

\pb\subsection{Discussion}

From above two experiments, we can find some important insights. 
First, the comparison between attack success rates of two deep learning models indicates that a deeper or more complicate neural network potentially involves more vulnerability to the adversarial attack. On the ResNet50, all algorithms achieve success rates over $99\%$. In contrast, the attack success rate on the simple convolution network with several layers is much lower. 

Second, the great gap of success rates between our Hessian-aware zeroth-order methods and two state-of-the-art algorithms on the MNIST may reveal such a potential that the Hessian information will bring great advantages on the hard adversarial attack problem. 

\begin{figure}[]
	\subfigtopskip = 0pt
	\begin{center}
		\centering
		\subfigure[\textsf{\texttt{ZOO}}]{\includegraphics[width=56mm]{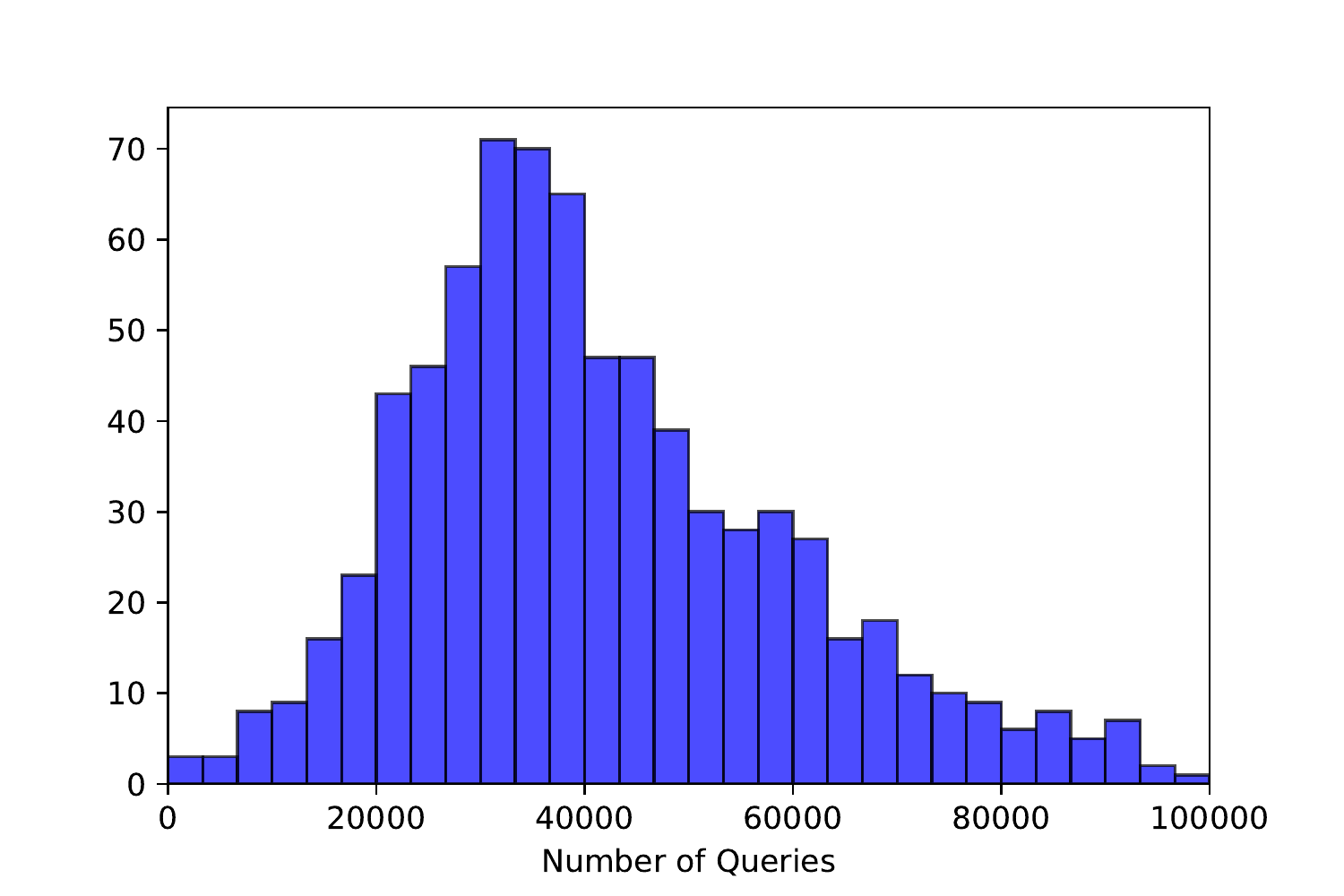}}~
		\subfigure[\textsf{ \texttt{ZOHA-Gauss}}]{\includegraphics[width=56mm]{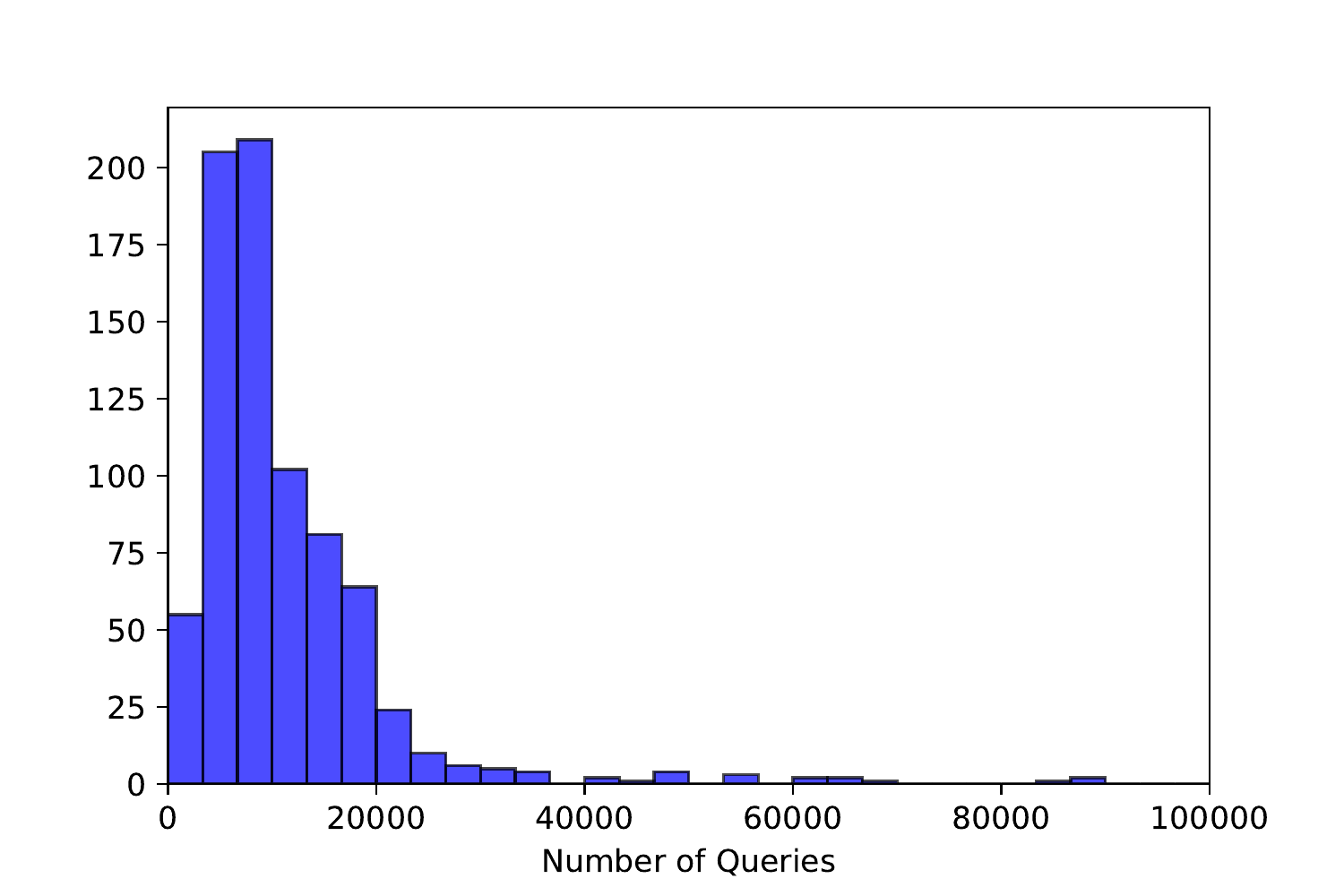}}~
		\subfigure[\textsf{\texttt{ZOHA-Diag}}]{\includegraphics[width=56mm]{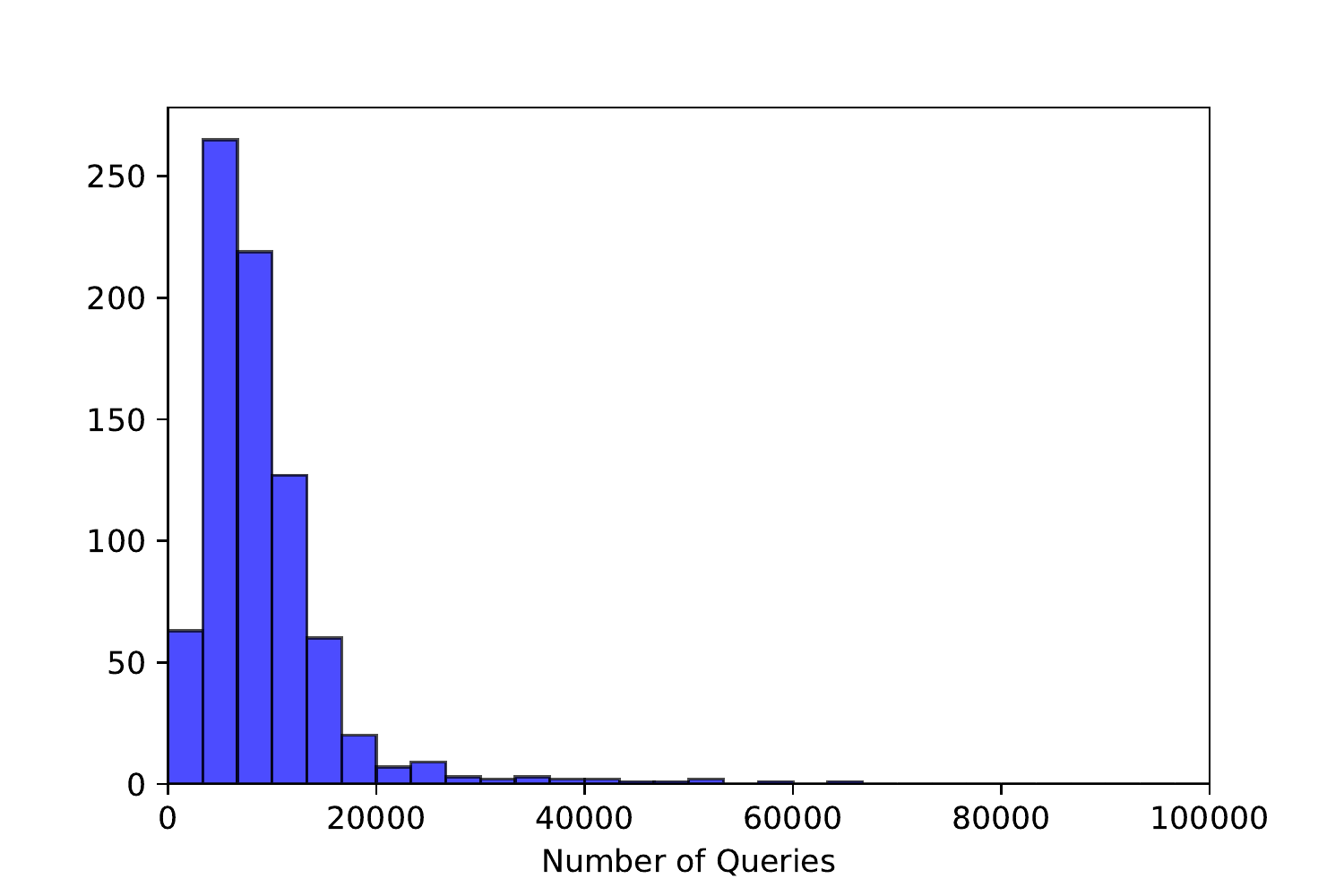}}\\
		\subfigure[\textsf{\texttt{PGD-NES}}]{\includegraphics[width=56mm]{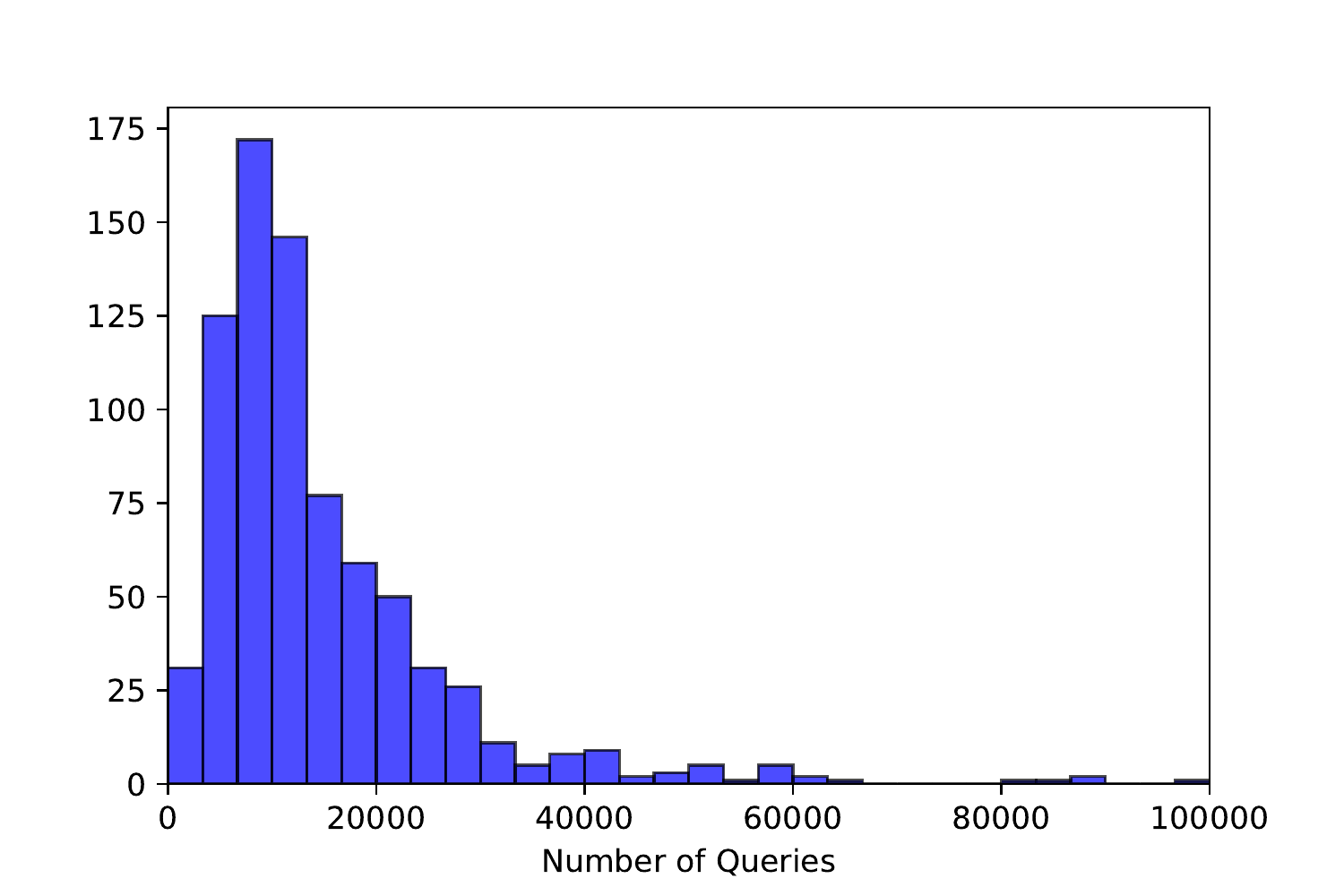}}~
		\subfigure[\textsf{ \texttt{ZOHA-Gauss-DC}}]{\includegraphics[width=56mm]{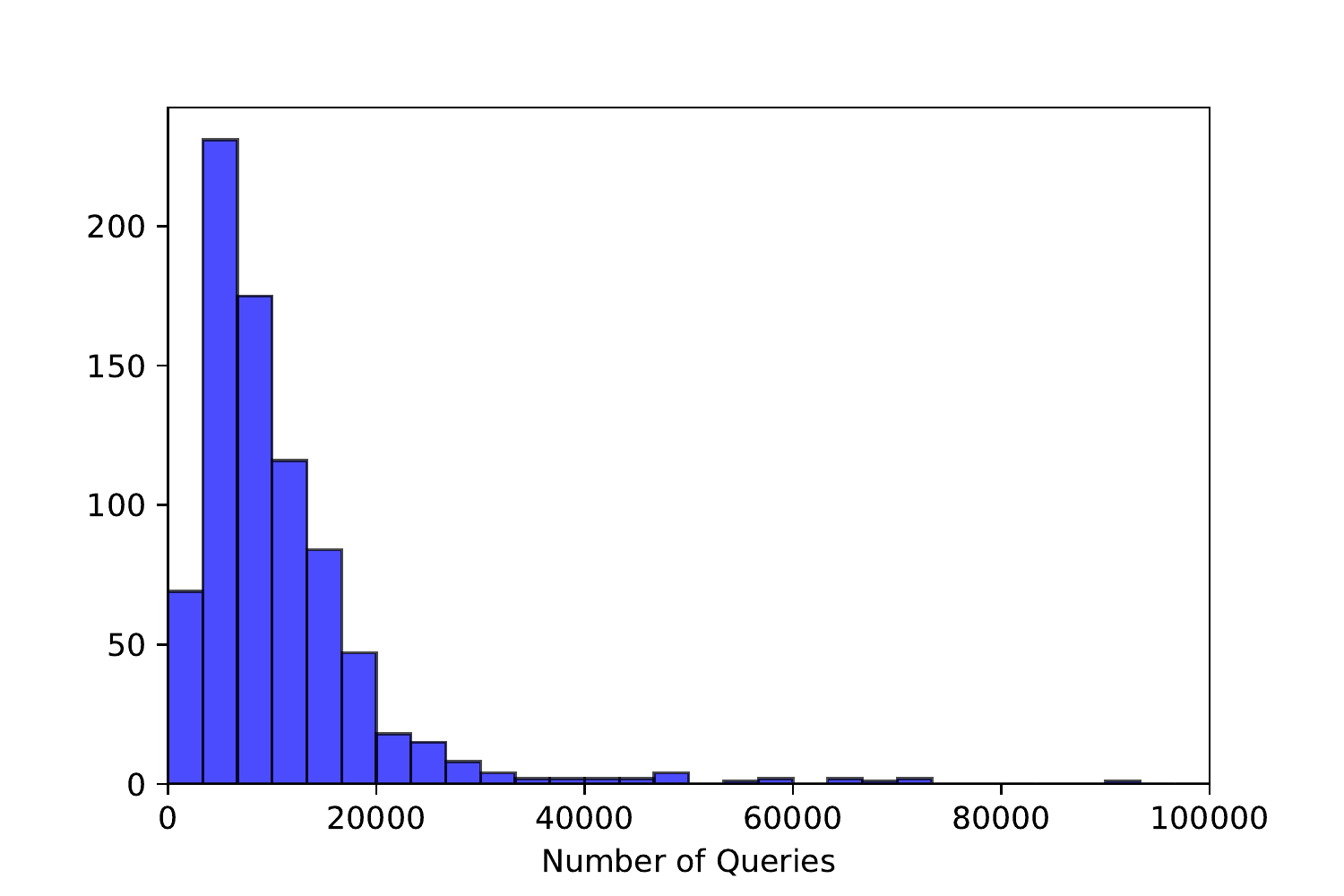}}~
		\subfigure[\textsf{\texttt{ZOHA-Diag-DC}}]{\includegraphics[width=56mm]{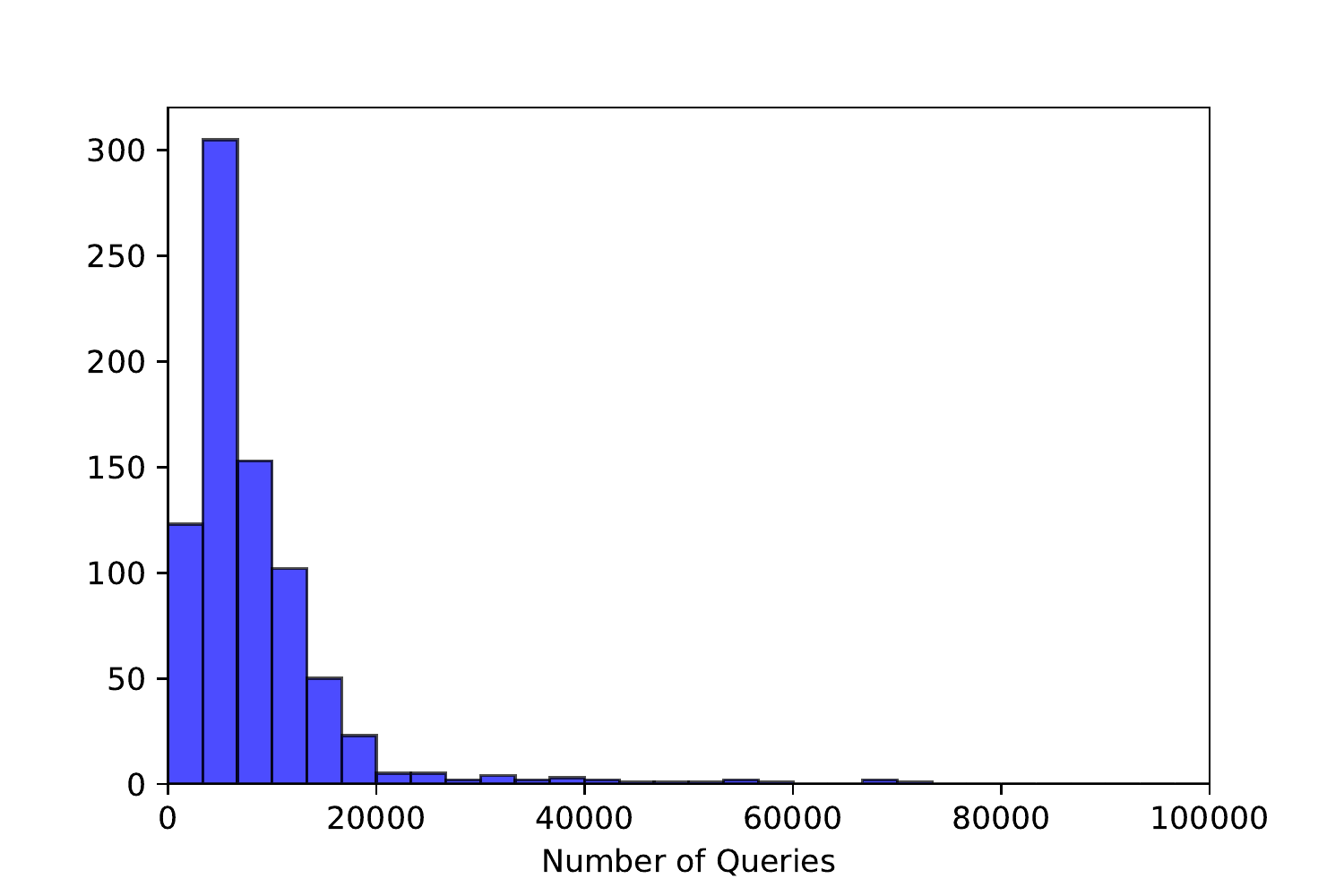}}
	\end{center}
	\caption{The distribution of the number of queries on \emph{targeted} black-box attacks on ResNet50 model and ImageNet}
	\label{fig:iamgenet_tar}
\end{figure}

\begin{figure}[!ht]
	\subfigtopskip = 0pt
	\begin{center}
		\centering
		\subfigure[\textsf{\texttt{ZOO}}]{\includegraphics[width=56mm]{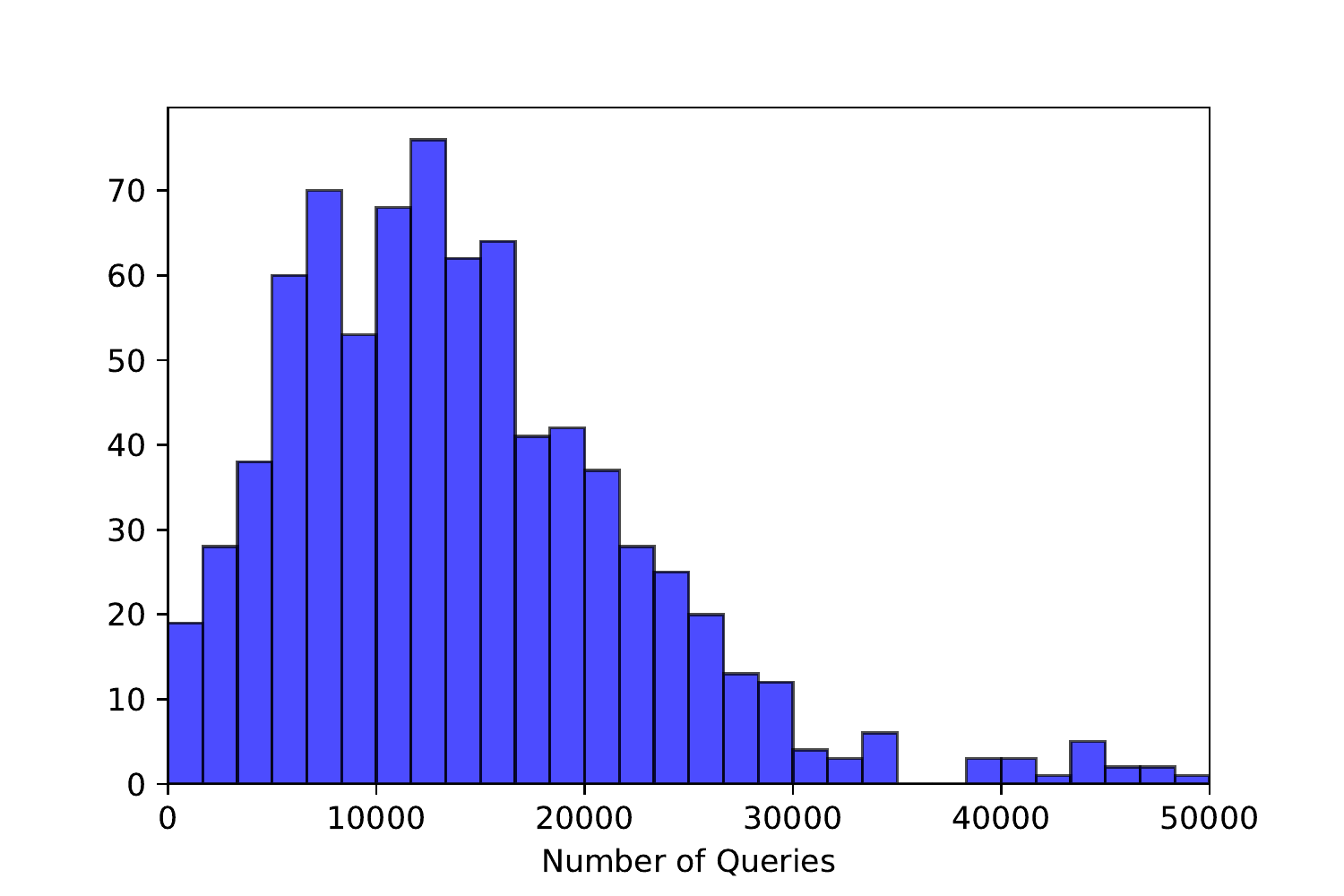}}~
		\subfigure[\textsf{ \texttt{ZOHA-Gauss}}]{\includegraphics[width=56mm]{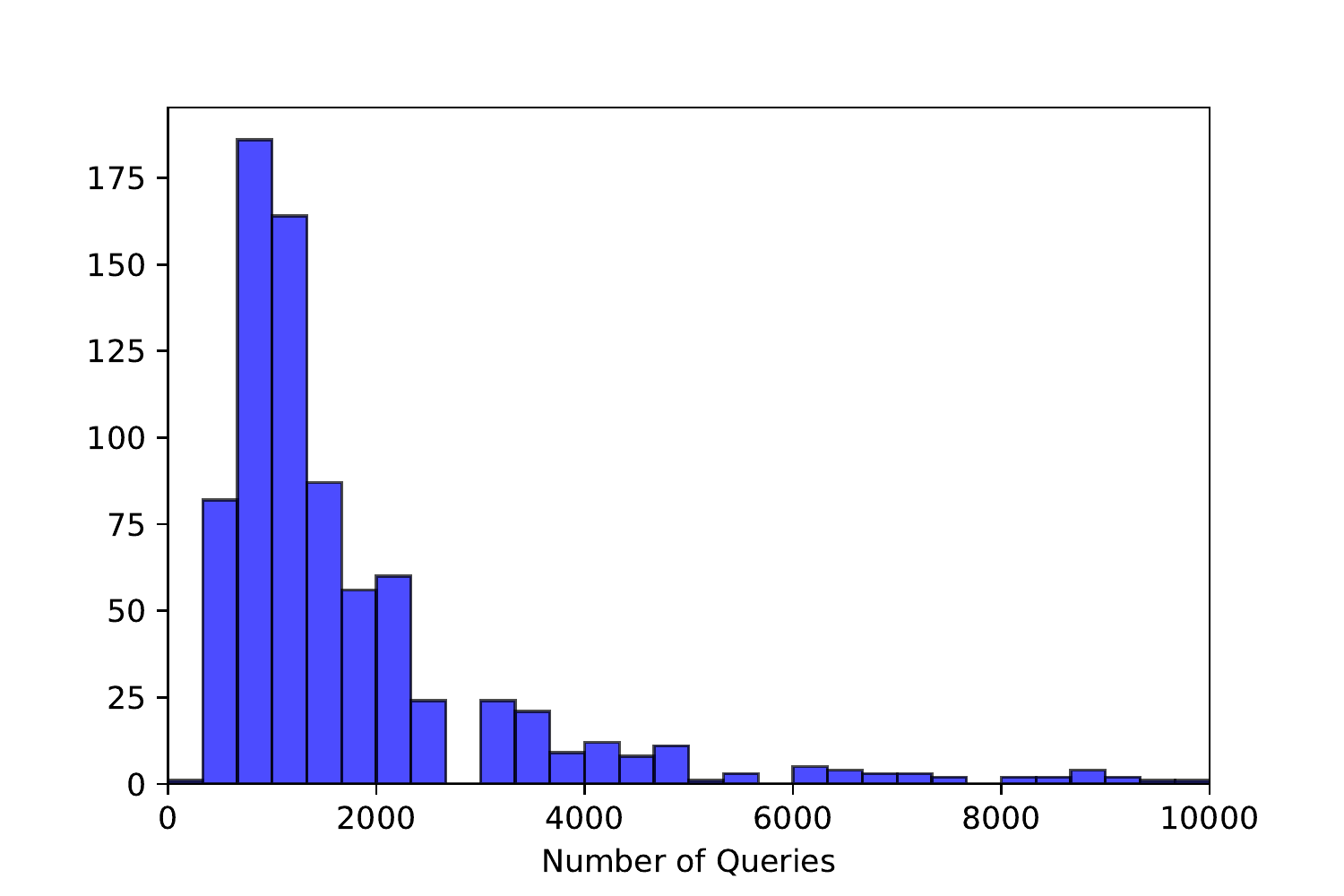}}~
		\subfigure[\textsf{\texttt{ZOHA-Diag}}]{\includegraphics[width=56mm]{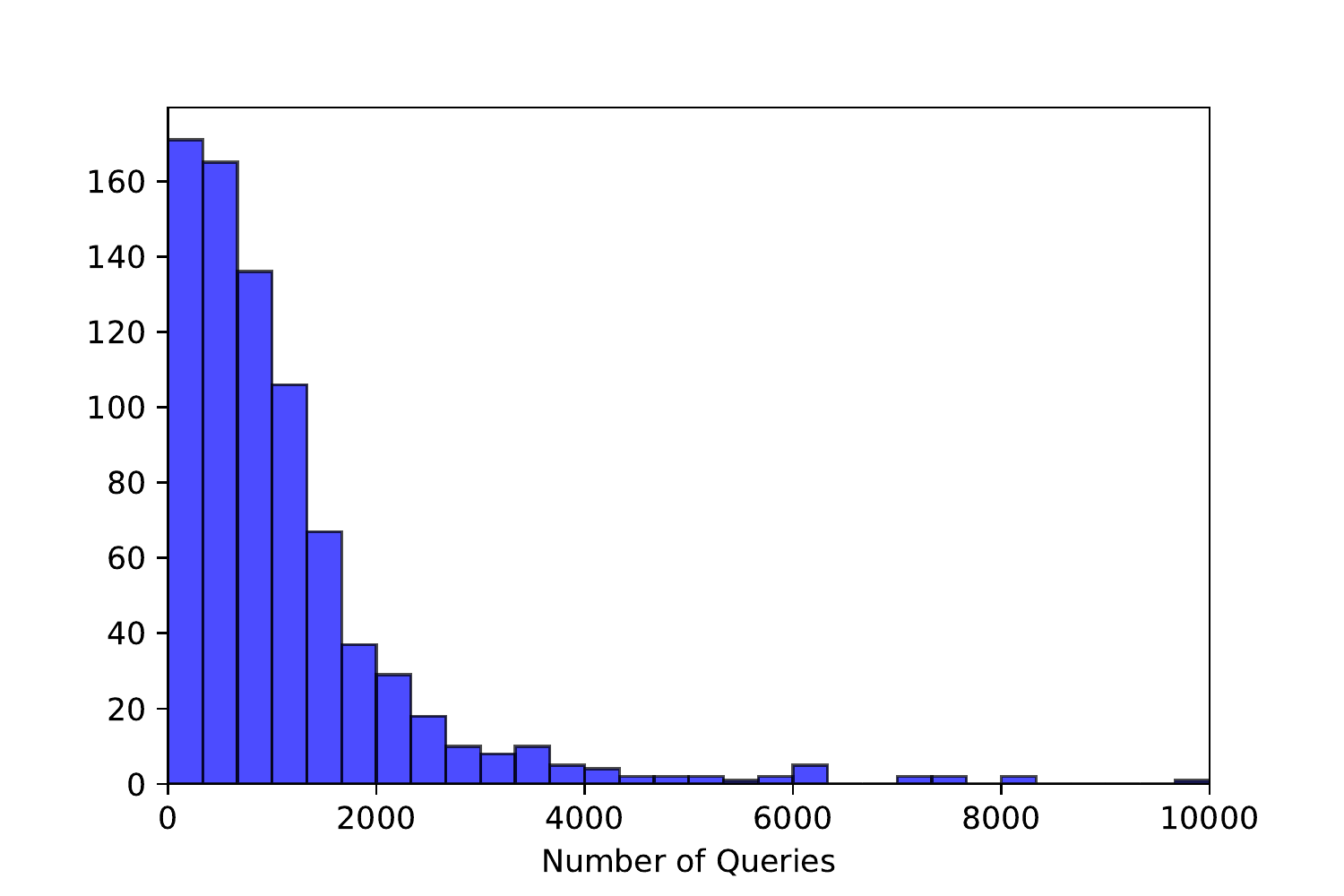}}\\
		\subfigure[\textsf{\texttt{PGD-NES}}]{\includegraphics[width=56mm]{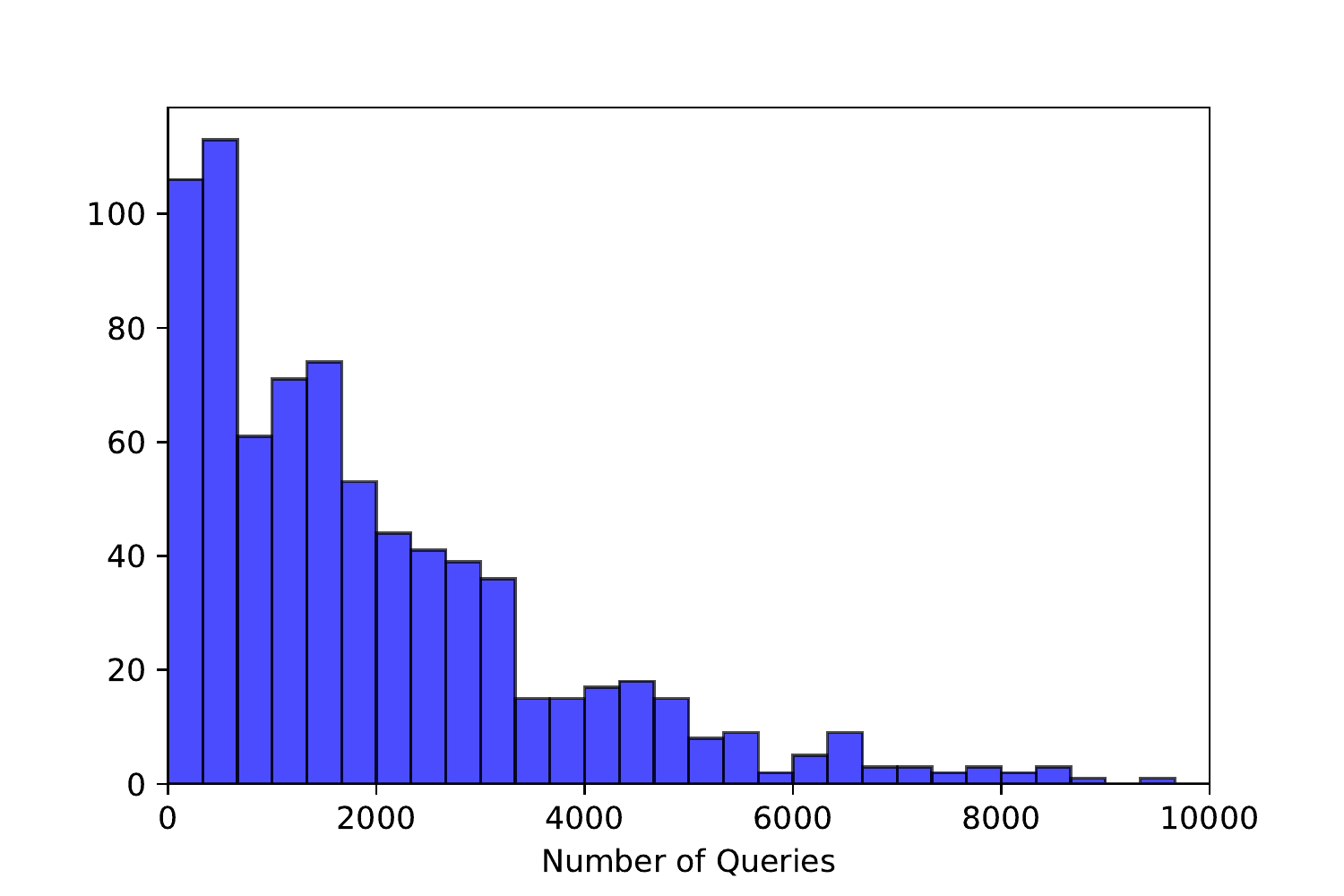}}~
		\subfigure[\textsf{ \texttt{ZOHA-Gauss-DC}}]{\includegraphics[width=56mm]{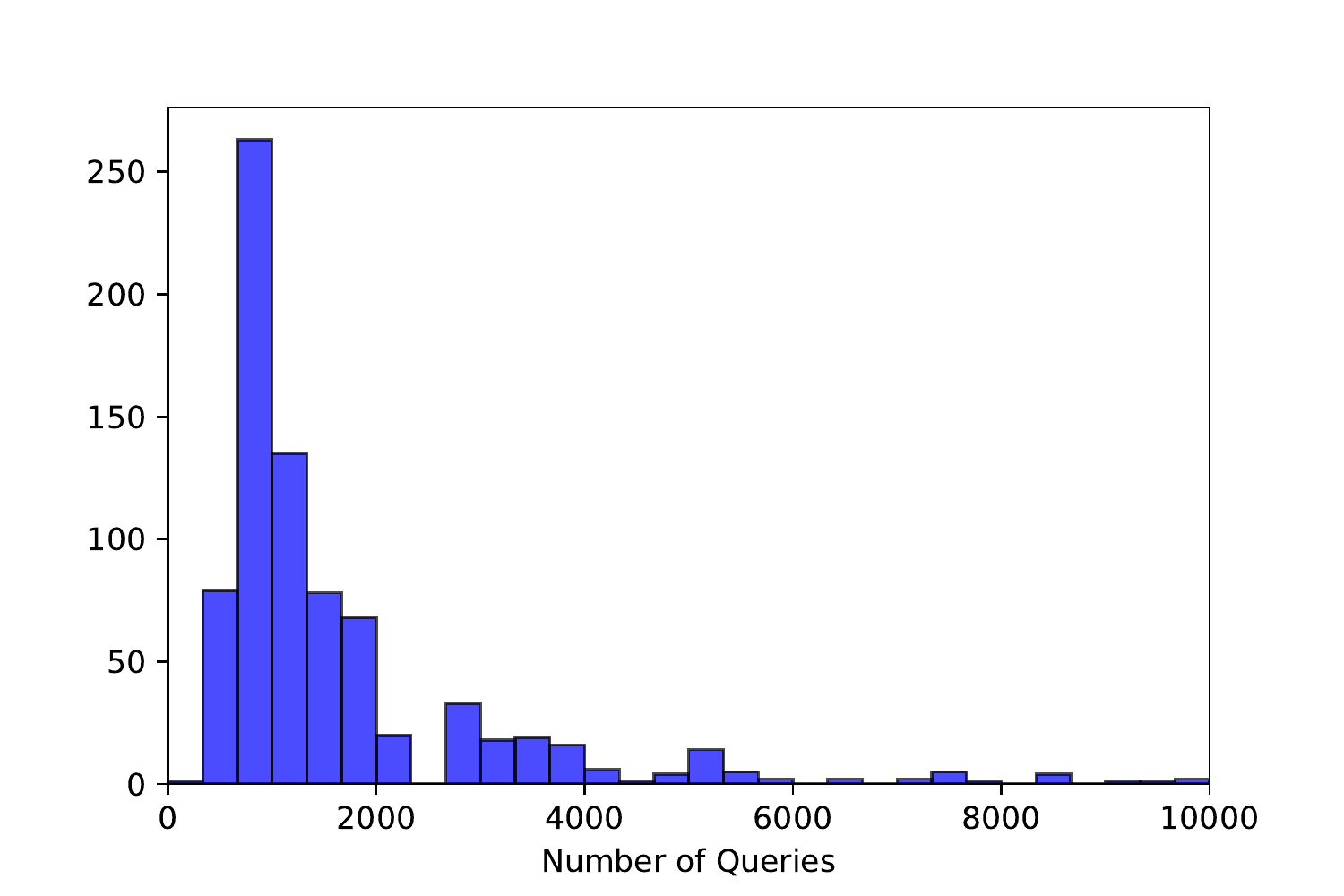}}~
		\subfigure[\textsf{\texttt{ZOHA-Diag-DC}}]{\includegraphics[width=56mm]{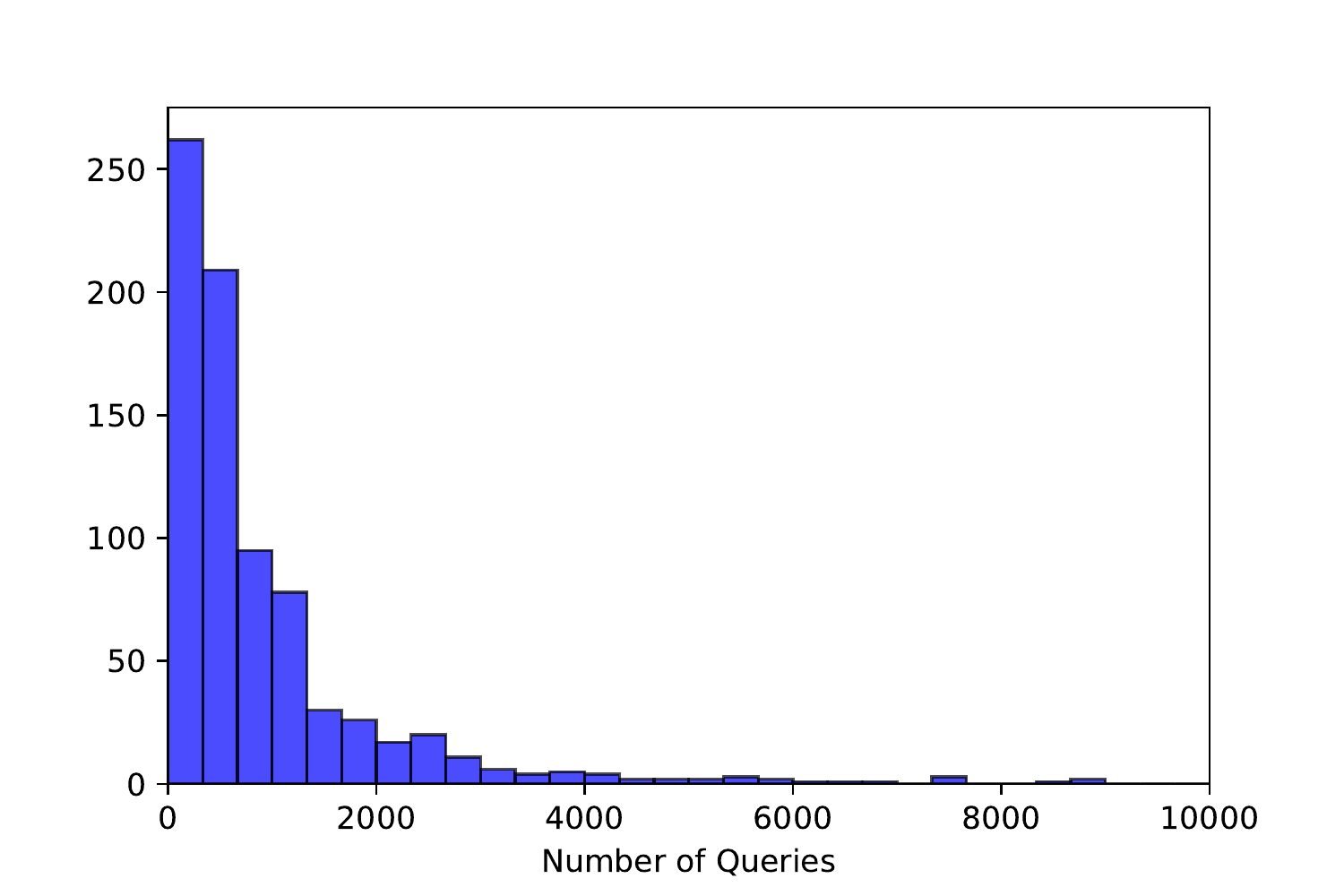}}
	\end{center}
	\caption{The distribution of the number of queries on \emph{un-targeted} black-box attacks on ResNet50 model and ImageNet}
	\label{fig:iamgenet_untar}
\end{figure}

Third, we can observe that our algorithms with \texttt{Descent-Checking} have better performance than the ones without \texttt{Descent-Checking}. The experiment results on the MNIST show that \texttt{Descent-Checking} strategy can promote attack success rate effectively both for targeted and un-targeted attack. At the same time, \texttt{Descent-Checking} is an effective way to reduce query complexity. This can be easily observed from the distribution of the number of queries in Figure~\ref{fig:mnist_tar}-\ref{fig:iamgenet_untar}.

\pb\section{Conclusion}\label{sec:conclusion}

In this paper, we propose a novel zeroth-order algorithmic framework called \texttt{ZO-HessAware} which exploits the second-order Hessian information of the model function. 
\texttt{ZO-HessAware} achieves a faster convergence rate and lower query complexity than vanilla zeroth-order algorithms that does \textit{not} use the second-order information. 
Simultaneously, we propose several novel query-efficient approaches to capture the dominant information of the Hessian. 
Experiments on the black-box adversarial attack show that our \texttt{ZO-HessAware} algorithms improve the attack success rate and reduce the query complexity effectively.
This validates the effectiveness of the Hessian information in zeroth-order optimization and our theoretical analysis empirically. 
We also propose a technique called \texttt{Descent-Checking} which can empirically further promote attack success rate and reduce query complexity.

\pb
\clearpage
\bibliography{ref.bib}
\bibliographystyle{apalike2}

\appendix

\pb\section{ Proof of Section~\ref{subsec:grad_mu}}

In this section, we will prove three properties of the estimated gradient defined in Eqn.~\eqref{eq:g_mu}. Before that, we first list some important lemmas related to the Gaussian distribution that will be used in our proof. 

\subsection{Important Lemmas}
\begin{lemma}[\citep{Nesterov2017}]\label{lem:L-smooth}
	If $f(x)$ is $L$-smooth, then we have
	\begin{align*}
	\norm{\nabla f_\mu(y) - \nabla f_\mu(x)} \leq L\norm{x-y}. 
	\end{align*}
\end{lemma}

\begin{lemma}[\citep{Nesterov2017}]\label{lem:grad_err}
	If $f(x)$ is $L$-smooth, then 
	\begin{align*}
	\norm{\nabla f_\mu(x) - \nabla f(x)}\leq \frac{\mu}{2}L(d+3)^{3/2}.
	\end{align*}
	If the Hessian is $\gamma$-Lipschitz continuous, then we can guarantee that
	\begin{align*}
	\norm{\nabla^2 f_\mu(x) - \nabla^2 f(x)}\leq \frac{\mu^2}{6}\gamma(d+4)^2.
	\end{align*}
\end{lemma}

\begin{lemma}[\citep{Nesterov2017}]\label{lem:gauss_power}
	Let $p\geq 2$, $u$ be from $N(0, I_d)$, then we have the following bound
	\begin{align*}
	d^{p/2} \leq \EB_u\left[\norm{u}^p\right]\leq (p+d)^{p/2}.
	\end{align*}
\end{lemma}

Then we give the results of moments of products quadratic forms in normal distribution.
\begin{lemma}[\citep{magnus1978moments}]\label{lem:prob_res}
	Let $A$ and $B$ be two symmetric matrices, and $u$ has the Gaussian distribution, that is, $u \sim N(0,I_d)$. Define $z = u^\top A u\cdot u^\top B u$. The expectation of $z$ and $z^2$ are:
	\begin{align*}
	\EB_u(z) = & (\tr\;A)(\tr\;B) + 2(\tr\; AB)\\
	\EB_u(z^2) =& (\tr\;A)^2(\tr\;B)^2+16\left[(\tr\;A)(\tr\;AB^2)+(\tr\; B)(\tr\; A^2B)\right]\\
	&+2\left[(\tr\;A)^2(\tr\;B)^2+4(\tr\;A)(\tr\;B)(\tr\;AB)+(\tr \;B)^2(\tr\;A^2)\right]\\
	&+4\left[(\tr\;A^2)(\tr\; B^2)+2(\tr\; AB)^2\right]+16\left[(\tr\;AB)^2+2(\tr\;A^2B^2)\right].
	\end{align*}
\end{lemma}

\subsection{Proof of Lemma~\ref{lem:T_1}}
\begin{proof}[Proof of Lemma~\ref{lem:T_1}]
	First, the gradient of $f_\mu(x)$ can be represented as \citep{Nesterov2017}:
	\begin{align}
	\nabla f_\mu(x) =& \frac{1}{M}\int_{\RB^d}\frac{f(x+\mu u)-f(x)}{\mu}u \;\exp{\left(-\frac{\norm{u}^2}{2}\right)}du\notag\\
	=&\EB_u\left[\frac{f(x+\mu u)-f(x)}{\mu}u \right] \label{eq:def_grad}
	\end{align}
	
	Let us denote 
	\[
	h(y) \triangleq f(x+Ky).
	\]
	Then we have
	\[
	f(x) = h(0).
	\]
	By Eqn.~\eqref{eq:def_grad}, we have
	\begin{align*}
	\EB_u \frac{f(x+\mu K u) - f(x)}{\mu}u=\EB_u \frac{h(0 +\mu u) - h(0)}{\mu}u=\nabla h_\mu(0).
	\end{align*}
	
	And we also have
	\begin{align*}
	\nabla h(0) = K \cdot \nabla f(x)
	\end{align*}
	By Lemma~\ref{lem:grad_err}, we can obtain that 
	\begin{align}
	\norm{\nabla h_\mu(0) - \nabla h(0)} \leq \frac{\mu}{2}L(h_\mu)(d+3)^{3/2}.
	%\norm{\nabla h_\mu(0) - \nabla h(0)} \leq \frac{\mu^2}{6}\gamma(h)(d+4)^{2}.
	\end{align}
	where $L(h_\mu)$ is the smoothness parameter of $h_\mu(y)$. By Lemma~\ref{lem:L-smooth}, we know that $L(h_\mu)$ is no larger than the one of $h(y)$. And $L(h)$ has the following upper bound:
	\begin{align*}
	L(h)\leq L(f)\norm{K}^2 = L\norm{K}^2.
	\end{align*}
	Therefore, we have 
	\begin{align*}
	\norm{\EB_u[g_\mu(x)] - \nabla f(x)}_{K^2}^2 =& \norm{K^{-1}(\nabla h_\mu(0) - \nabla h(0))}_{K^2}^2 \\
	=&\norm{\nabla h_\mu(0) - \nabla h(0)}^2\\
	\leq&\frac{\mu^2}{4}L^2\norm{K}^4(d+3)^{3}.
	\end{align*}

\end{proof}

\subsection{Proof of Lemma~\ref{lem:T_2}}
\begin{proof} [Proof of Lemma~\ref{lem:T_2}]
	By the definition of $g_\mu(x)$, we have
	\begin{align*}
	\EB_{u} \norm{g_\mu(x)}^2_{K^2}=&\frac{1}{\mu^2b^2}\EB\norm{\left(\sum_{i=1}^{b}\left[f(x+\mu K u_i) - f(x)\right]K^{-1}u_i\right)}_{K^2}^2\\
	=&\frac{1}{\mu^2b^2}\EE_{u_i, u_j}\sum_{i,j} \left(f(x+\mu K u_i) - f(x)\right) \left(f(x+\mu K u_i) - f(x)\right)u_iu_j\\
	=&\frac{1}{\mu^2b}\EB_{u}\left(\left[f(x+\mu K u) - f(x)\right]^2\norm{K^{-1}u}^2_{K^2}\right) \\&+ \frac{1}{\mu^2b^2}\sum_{i\neq j} \left(\EE_{u_i}\left(f(x+\mu K u_i) - f(x)\right) u_i\right)\left(\EE_{u_j}\left(f(x+\mu K u_j) - f(x)\right) u_j\right)\\
	=&\frac{1}{\mu^2b}\EB_{u}\left(\left[f(x+\mu K u) - f(x)\right]^2\norm{K^{-1}u}^2_{K^2}\right) + \frac{b^2-b}{b^2}\norm{\EB[g_\mu(x)]}_{K^2}^2\\
	\leq&\frac{1}{\mu^2b}\EB_{u}\left(\left[f(x+\mu K u) - f(x)\right]^2\norm{K^{-1}u}^2_{K^2}\right) +  2\norm{\nabla f(x)}_{K^2}^2 + \frac{\mu^2L^2(d+3)^3}{2}
	\end{align*}
	The first equality is because $u_i$'s are independent, and we have
	\begin{align*}
	\left[f(x+\mu K u) - f(x)\right]^2 =& [f(x+\mu K u) - f(x) - \mu\dotprod{\nabla f(x), Ku} +\mu\dotprod{\nabla f(x), Ku} ]\\
	\leq& 2\left[\frac{\mu^2}{2}L\norm{K}^2\norm{u}^2\right]^2 + 2\mu^2\langle \nabla f(x), Ku\rangle^2
	\end{align*}
	where the last inequality is from Lemma~\ref{lem:T_1}. We can also obtain that
	\begin{align*}
	\EB_u\left[\langle K \nabla f(x), u\rangle^2\norm{K^{-1}u}^2_{K^2}\right]=& \EB_u\left[u^\top K\nabla f(x) \nabla f(x)^\top K^\top u \cdot u^\top K^{-1} K^2 K^{-1} u\right]\\
	=& (\tr \,K\nabla f(x) \nabla f(x)^\top K^\top)(\tr \, I) + 2\tr(K\nabla f(x) \nabla f(x)^\top K^\top)\\
	=&d\norm{\nabla f(x)}_{K^2}^2 + 2\norm{\nabla f(x)}_{K^2}^2\\
	=&(d+2)\norm{\nabla f(x)}_{K^2}^2,
	\end{align*}
	where the second equation is because of Lemma~\ref{lem:prob_res} with $A = K\nabla f(x) \nabla f(x)^\top K^\top$ and $B = K^{-1} K^2 K^{-1} = I$.
	Therefore, we have
	\begin{align}
	&\EB_{u} \norm{g_\mu(x)}^2_{K^2} \notag\\
	\leq& \frac{\mu^2}{2b}L^2\norm{K}^4\EB_u\left[\norm{u}^4\cdot\norm{K^{-1}u}^2_{K^2}\right]+\frac{1}{b}\cdot\EB_u\left[\langle K \nabla f(x), u\rangle^2\norm{K^{-1}u}^2_{K^2}\right]+ 2\norm{\nabla f(x)}_{K^2}^2 + \frac{\mu^2L^2(d+3)^3}{2}\notag\\
	\leq& \frac{\mu^2}{2b}L^2\norm{K}^4\EB_u\left[\norm{u}^6\right] + \left(\frac{2(d+2)}{b} + 2\right)\cdot \norm{\nabla f(x)}_{K^2}^2 + \frac{\mu^2L^2(d+3)^3}{2}\notag\\
	\leq&\frac{\mu^2}{2b}L^2\norm{K}^4(d+6)^3 + \left(\frac{2(d+2)}{b} + 2\right) \cdot\norm{\nabla f(x)}_{K^2}^2 + \frac{\mu^2L^2(d+3)^3}{2}, \label{eq:grad_variance}
	\end{align}
	where the last inequality is due to Lemma~\ref{lem:gauss_power}.
\end{proof}

\subsection{Proof of Lemma~\ref{lem:T_3}}
\begin{proof} [Proof of Lemma~\ref{lem:T_3}]
	First, we have 
	\begin{align*}
	\EB_u\norm{K^2 g_\mu(x)}_{K^{-2}}^3 \leq & \frac{1}{\mu^3b^3} \cdot b^2\sum_{i=1}^{b}\EB_{u_i}\left([f(x+\mu K u_i) - f(x)]^{3}\cdot\norm{K u_i}_{K^{-2}}^3\right)\\
	=&\frac{1}{\mu^3}\EB_u\left([f(x+\mu K u) - f(x)]^{3}\cdot\norm{u}^3\right).
	\end{align*}
	Furthermore, we can obtain that
	\begin{align*}
	[f(x+\mu K u) - f(x)]^{3} =& [f(x+\mu K u) - f(x) - \mu\dotprod{\nabla f(x), Ku} +\mu\dotprod{\nabla f(x), Ku} ]^3\\
	\leq& 4\left[\frac{\mu^2}{2}L\norm{K}^2\norm{u}^2\right]^3 + 4\mu^3\langle K\nabla f(x), u\rangle^3
	,
	\end{align*}
	and we also have
	\begin{align*}
	&\EB_u\left[\norm{u}^3\cdot\dotprod{K\nabla f(x), u}^3\right]\\
	=&\EB_u\left[\left((u^\top u)^2\cdot(u^\top K\nabla f(x) \nabla^\top f(x) Ku)^2\right)^{3/4}\right]\\
	\leq&\left(\EB_u\left[(u^\top u)^2\cdot(u^\top K\nabla f(x) \nabla^\top f(x) Ku)^2\right]\right)^{3/4},
	\end{align*}
	where the last inequality is because Jensen's inequality. 
	
	Let us denote $A = K\nabla f(x) \nabla^\top f(x) K$. It is easy to check that $A$ is a rank one positive semi-definite matrix, and its trace satisfies that $\tr(A) = \norm{\nabla f(x)}_{K^2}^2$. By Lemma~\ref{lem:prob_res} with $A = K\nabla f(x) \nabla^\top f(x) K$ and $B=I$, we have
	\begin{align*}
	&\EB_u\left[(u^\top u)^2\cdot(u^\top K\nabla f(x) \nabla^\top f(x) Ku)^2\right] \\
	=&(\tr\; I_d)^2(\tr\; A)^2 + 16[(\tr\; I_d)(\tr\;I_dA^2)+(\tr\;A)(\tr\; I_d^2A)]\\
	&+4[(\tr\; I_d^2)(\tr\; A^2)+2(\tr\; I_dA)^2]+2[(\tr\; I_d)^2(\tr\; A^2)+4(\tr I_d)(\tr\; A)(\tr \; I_dA)\\
	&+(\tr\; A)^2(\tr I_d^2)]+16[\tr(I_dA)^2+2(\tr\; I_d^2A^2)]\\
	=&d^2\norm{\nabla f(x)}_{K^2}^4+16\left[d\norm{\nabla f(x)}_{K^2}^4+\norm{\nabla f(x)}_{K^2}^4\right]+4\left[d\norm{\nabla f(x)}_{K^2}^4+2\norm{\nabla f(x)}_{K^2}^4\right]\\
	&+2\left[d^2\norm{\nabla f(x)}_{K^2}^4+4d\norm{\nabla f(x)}_{K^2}^4+d\norm{\nabla f(x)}_{k^2}^4\right]+16\left[\norm{\nabla f(x)}_{K^2}^4+2\norm{\nabla f(x)}_{K^2}^4\right]\\
	=&(3d^2+30d+72)\cdot\norm{\nabla f(x)}_{K^2}^4.
	\end{align*}
	Thus, we can obtain that
	\begin{align*}
	\EB_u\left[\norm{u}^3\cdot\dotprod{K\nabla f(x), u}^3\right] \leq&\left(\EB_u\left[(u^\top u)^2\cdot(u^\top K\nabla f(x) \nabla^\top f(x) Ku)^2\right]\right)^{3/4}\\
	\leq&\left((3d^2+30d+72)\cdot\norm{\nabla f(x)}_{K^2}^4\right)^{3/4}\\
	\leq&\left(3(d+5)^2\cdot\norm{\nabla f(x)}_{K^2}^4\right)^{3/4}\\
	\leq&3(d+5)^{3/2}\norm{\nabla f(x)}_{K^2}^3
	.
	\end{align*}
	
	Therefore, we have
	\begin{align*}
	\EB_u\norm{K^2 g_\mu(x)}_{K^{-2}}^3 \leq& \frac{1}{\mu^3} \EB_u\left( 4\left[\frac{\mu^2}{2}L\norm{K}^2\norm{u}^2\right]^3\norm{u}^3 + 4\mu^3\langle K\nabla f(x), u\rangle^3 \norm{u}^3\right)\\
	\leq&\frac{\mu^3L^3\norm{K}^6}{2}\EB_u\left[\norm{u}^9\right]+12(d+5)^{3/2}\norm{\nabla f(x)}_{K^2}^3\\
	\leq&2\mu^3L^3\norm{K}^6\cdot(d+9)^{9/2}+12(d+5)^{3/2}\norm{\nabla f(x)}_{K^2}^3
	.
	\end{align*}
	where the last inequality follows from Lemma~\ref{lem:gauss_power}.
\end{proof}

\pb\section{Proof of Convergence Rate of Algorithm~\ref{alg:zero_order}}

In this section, we will prove the convergence rate of Algorithm~\ref{alg:zero_order}. Before that, we first give an important lemma which depicts some properties related to the strong convexity. 

\begin{lemma}\label{lem:str_cvx}
	Let $f$ be continuously differentiable and strongly convex with parameter $\tau$. And $x^\star$ is the minimizer of $f$. Then for any $x\in\RB^d$, we have
	\begin{align*}
	\norm{\nabla f(x)}^2 \geq 2\tau(f(x) - f(x^\star),
	\end{align*}
	and
	\begin{align*}
	\norm{x-x^\star}^2 \leq \frac{2}{\tau}(f(x) - f(x^\star)).
	\end{align*}
\end{lemma}
\begin{proof}
	First, by the strong convexity of $f$, 
	\begin{align*}
	f(x^\star) \geq& f(x) + \dotprod{\nabla f(x), x^\star - x} + \frac{\tau}{2}\norm{x^\star -x}^2\\
	\geq& f(x) + \min_{v}\left(\dotprod{\nabla f(x), v} + \frac{\tau}{2}\norm{v}^2\right)\\
	=&f(x) - \frac{1}{2\tau}\norm{\nabla f(x)}^2.
	\end{align*}
	The last equality holds by plugging in the minimizer $v = -\nabla f(x) / \tau$.
	
	Also by the strong convexity of $f$, we have
	\begin{align*}
	f(x) \geq& f(x^\star) + \dotprod{\nabla f(x^\star), x - x^\star} + \frac{\tau}{2}\norm{x^\star -x}^2\\
	=& f(x^\star) + \frac{\tau}{2}\norm{x^\star -x}^2.
	\end{align*}
\end{proof}

\subsection{Proof of Theorem~\ref{thm:local}}
Now, we give the proof of the local convergence properties depicted in Theorem~\ref{thm:local}. 
\begin{proof} [Proof of Theorem~\ref{thm:local}]
	Let $H$ denote the Hessian $\nabla^2f(x_t)$. 
	Taking a random step from $x_t$, we have
	\begin{align}
	&\EB_u\left[f(x_{t+1})\right]\notag\\
	=&\EB_u\left[f(x_{t} - \eta\TH^{-1}g_\mu(x_t))\right]\notag\\
	\overset{\eqref{eq:gamma_2}}{\leq}&f(x_t)-\eta\dotprod{\nabla f(x_t), \TH^{-1}\EB_u[g_\mu(x_t)]}+\frac{\eta^2}{2}\EB_u\norm{\TH^{-1}g_\mu(x_t)}_H^2+\frac{\eta^3\gamma}{6}\EB_u\norm{\TH^{-1}g_\mu(x_t)}^3\notag\\
	\overset{\eqref{eq:prec_cond}}{\leq}& f(x_t) - \eta\norm{\nabla f(x_t)}_{\TH^{-1}}^2 -\eta\dotprod{\nabla f(x_t), \TH^{-1}\left(\EB_u[g_\mu(x_t)] - \nabla f(x_t)\right)}\notag\\
	& + \eta^2\EB_u\norm{\TH^{-1}g_\mu(x_t)}_{\TH}^2+\frac{\eta^3\gamma}{6}\EB_u\norm{\TH^{-1}g_\mu(x_t)}^3\notag\\
	\leq&f(x_t) - \eta\norm{ \nabla f(x_t)}_{\TH^{-1}}^2 + \frac{\eta}{2}\left(\norm{\nabla f(x_t)}_{\TH^{-1}}^2+\norm{\EB_u[g_\mu(x_t)] - \nabla f(x_t)}_{\TH^{-1}}^2\right)\label{eq:cauchy_i}\\&
	+ \eta^2\EB_u\norm{\TH^{-1}g_\mu(x_t)}_{\TH}^2+\frac{\eta^3\gamma}{6}\EB_u\norm{\TH^{-1}g_\mu(x_t)}^3\notag\\
	=&f(x_t) - \frac{\eta}{2}\norm{ \nabla f(x_t)}_{\TH^{-1}}^2 +\underbrace{\frac{\eta}{2}\norm{\EB_u[g_\mu(x_t)] - \nabla f(x_t)}_{\TH^{-1}}^2}_{T_1}\notag+\underbrace{\eta^2\EB_u\norm{\TH^{-1}g_\mu(x_t)}_{\TH}^2}_{T_2} \\&+\underbrace{\frac{\eta^3\gamma}{6}\EB_u\norm{\TH^{-1}g_\mu(x_t)}^3}_{T_3}
	, \notag
	\end{align}
	where inequality~\eqref{eq:cauchy_i} follows from Cauchy's Inequality. 
	
	Now we begin to bound terms $T_1$, $T_2$, and $T_3$. First, by Lemma~\ref{lem:T_1} with $K = \TH^{-1/2}$, we have 
	\begin{align*}
	\norm{\EB_u[g_\mu(x_t)] - \nabla f(x_t)}_{\TH^{-1}}^2 = \norm{\EB_u[g_\mu(x_t)] - \nabla f(x_t)}_{K^2}^2
	\leq\frac{\mu^2L^2}{4}\norm{\TH^{-1}}^2(d+3)^{3},
	\end{align*}
	that is,
	\begin{align*}
	T_1 \leq \frac{\eta \mu^2L^2}{8}\norm{\TH^{-1}}^2(d+3)^{3}.
	\end{align*}
	
	By Lemma~\ref{lem:T_2} with $K = \TH^{-1/2}$, we have
	\begin{align*}
	\EB_u\norm{\TH^{-1}g_\mu(x_t)}_{\TH}^2 =& \EB_u\norm{g_\mu(x_t)}_{\TH^{-1}}^2\\
	\leq&\frac{\mu^2}{2b}L^2\norm{\TH^{-1}}^2(d+6)^3 + \left(\frac{2(d+2)}{b}+2\right)\norm{\nabla f(x)}_{\TH^{-1}}^2 + \frac{\mu^2L^2(d+3)^3}{2}.
	\end{align*}
	Using the condition that $b\leq d+2$, $T_2$ is upper bounded as 
	\[
	T_2 \leq \eta^2\left(\frac{\mu^2L^2}{2b}\norm{\TH^{-1}}^2(d+6)^3 + \frac{4(d+2)}{b} \norm{\nabla f(x)}_{\TH^{-1}}^2+\frac{\mu^2L^2(d+3)^3}{2}\right).
	\]
	Using Lemma~\ref{lem:T_3}, we have
	\begin{align*}
	T_3 =& \frac{\eta^3}{6}\cdot\EB_u\norm{\TH^{-1}g_\mu(x_t)}^3\\
	\leq&\frac{\eta^3}{6}\cdot\norm{\TH^{-1}}^{3/2}\cdot\EB_u\norm{\TH^{-1}g_\mu(x_t)}_{\TH}^3\\
	\leq&\frac{\gamma\mu^3L^3\eta^3\norm{\TH^{-1}}^{9/2}}{12}\cdot(d+9)^{9/2}+2\gamma\eta^3(d+5)^{3/2}\norm{\TH^{-1}}^{3/2}\cdot\norm{\nabla f(x)}_{\TH^{-1}}^3
	.
	\end{align*}
	
	Thus, we have 
	\begin{align*}
	&T_1+T_2+T_3 \\
	\leq&\frac{\eta \mu^2L^2}{8}\norm{\TH^{-1}}^2(d+3)^{3}\notag+\frac{\mu^2L^2\eta^2}{2b}\norm{\TH^{-1}}^2(d+6)^3 + \frac{4(d+2)\eta^2}{b}\norm{\nabla f(x)}_{\TH^{-1}}^2\\&+\frac{\eta^2\mu^2L^2}{2}(d+3)^3+\frac{\gamma\mu^3L^3\eta^3\norm{\TH^{-1}}^{9/2}}{12}\cdot(d+9)^{9/2}+2\gamma\eta^3(d+5)^{3/2}\norm{\TH^{-1}}^{3/2}\cdot\norm{\nabla f(x)}_{\TH^{-1}}^3.
	\end{align*}
	
	By choosing $\eta = \frac{b}{16(d+2)}$, we obtain that
	\begin{align*}
	&\EB_u\left[f(x_{t+1})\right]\\
	\leq& f(x_t) - \frac{\eta}{2}\norm{ \nabla f(x_t)}_{\TH^{-1}}^2 + T_1+T_2+T_3\\
	\leq&f(x_t) - \frac{b}{64(d+2)}\norm{ \nabla f(x_t)}_{\TH^{-1}}^2+\frac{\mu^2L^2b}{64}\norm{\TH^{-1}}^2(d+5)^2+\frac{b\mu^2L^2\norm{\TH^{-1}}}{128}(d+38)\\
	&+\frac{b^3\gamma\mu^3L^3\norm{\TH^{-1}}^{9/2}}{6144}\cdot(d+110)^{3/2}+\frac{b^2\mu^2L^2}{256}(d+7)+\frac{b^3\gamma }{64}d^{-3/2}\norm{\TH^{-1}}^{3/2}\cdot\norm{\nabla f(x)}_{\TH^{-1}}^3\\
	=&f(x_t) - \frac{b}{32(d+2)}\norm{ \nabla f(x_t)}_{\TH^{-1}}^2+\frac{b^3\gamma }{64}d^{-3/2}\norm{\TH^{-1}}^{3/2}\cdot\norm{\nabla f(x)}_{\TH^{-1}}^3\\
	&+b\cdot \left(\frac{\mu^2L^2}{64}\norm{\TH^{-1}}^2(d+5)^2+\frac{\mu^2L^2\norm{\TH^{-1}}}{128}(d+38)+\frac{\mu^2L^2b}{256}(d+7)+\frac{\gamma b^2\mu^3L^3\norm{\TH^{-1}}^{9/2}}{6144}\cdot(d+110)^{3/2}\right)\\
	=&f(x_t) - \frac{b}{32(d+2)}\norm{ \nabla f(x_t)}_{\TH^{-1}}^2+\frac{b^3\gamma }{64}d^{-3/2}\norm{\TH^{-1}}^{3/2}\cdot\norm{\nabla f(x)}_{\TH^{-1}}^3+\Delta_\mu,
	\end{align*}
	where we denote 
	\[
	\Delta_\mu=b\cdot \left(\frac{\mu^2L^2}{64}\norm{\TH^{-1}}^2(d+5)^2+\frac{\mu^2L^2\norm{\TH^{-1}}}{128}(d+38)+\frac{\mu^2L^2b}{256}(d+7)+\frac{\gamma b^2\mu^3L^3\norm{\TH^{-1}}^{9/2}}{6144}\cdot(d+110)^{3/2}\right).
	\]
	
	Now, we begin to give the connections between $\norm{\nabla f(x_t)}_{\TH^{-1}}^2$ and $f(x_t) - f(x^\star)$. First, by the Taylor's expansion, we have
	\begin{align*}
	\nabla f(x_t) =& \nabla f(x^\star) + \nabla^2f(x^\star)(x_t - x^\star) + \int_{0}^{1}\left(\nabla^2f(x^\star + s (x_t - x^\star)) - \nabla^2f(x^\star)\right)(x_t - x^\star)ds\\
	=& \nabla^2f(x^\star)(x_t - x^\star) + \Delta_1.
	\end{align*}
	where the last equation is because $\nabla f(x^\star) = 0$ and we denote that 
	\begin{align*}
	\Delta_1 \triangleq \int_{0}^{1}\left(\nabla^2f(x^\star + s (x_t - x^\star)) - \nabla^2f(x^\star)\right)(x_t - x^\star)ds
	,
	\end{align*}
	and $\norm{\Delta_1}$ is upper bounded as 
	\begin{align}
	\norm{\Delta_1} \leq \rho \int_{0}^{1} s \norm{x_t - x^\star}^2 \; ds = \frac{\rho}{2}\norm{x_t - x^\star}^2.
	\end{align}
	
	Let us denote $H_\star = \nabla^2 f(x^\star)$. We have
	\begin{align*}
	&-\norm{\nabla f(x_t)}^2_{\TH^{-1}}\\
	=&- \norm{H_\star(x_t - x^\star)+\Delta_1}^2_{\TH^{-1}}\\
	=&-\norm{H(x_t - x^\star)+\Delta_1+(H_\star - H)(x_t - x^\star)}^2_{\TH^{-1}}\\
	\leq&- \left(\norm{H(x_t-x^\star)}_{\TH^{-1}}^2 + \norm{\Delta_1}^2_{\TH^{-1}}+\norm{(H_\star- H) (x_t-x^\star)}^2_{\TH^{-1}}\right)\\
	&+2\bigg(\norm{H(x_t-x^\star)}_{\TH^{-1}} \cdot \norm{\Delta_1}_{\TH^{-1}}+\norm{\Delta_1}_{\TH^{-1}}\cdot\norm{(H_\star- H) (x_t-x^\star)}_{\TH^{-1}}\\
	&+\norm{(H_\star- H) (x_t-x^\star)}_{\TH^{-1}}\cdot\norm{H(x_t-x^\star)}_{\TH^{-1}}\bigg)\\
	\leq& - \rho\norm{x_t - x^\star}_{H}^2 + \Delta_2
	.
	\end{align*}
	where the last inequality is because of the condition that $\rho\TH\preceq H$ and we denote that 
	\begin{align*}
	\Delta_2 =& 2
	\norm{H(x_t-x^\star)}_{\TH^{-1}} \cdot \norm{\Delta_1}_{\TH^{-1}}+2\norm{\Delta_1}_{\TH^{-1}}\cdot\norm{(H_\star- H) (x_t-x^\star)}_{\TH^{-1}}\\&+2\norm{(H_\star- H) (x_t-x^\star)}_{\TH^{-1}}\cdot\norm{H(x_t-x^\star)}_{\TH^{-1}}
	.
	\end{align*}
	Furthermore, we have
	\begin{align*}
	f(x_t) \overset{\eqref{eq:gamma_2}}{\leq}& f(x^\star) + \langle \nabla f(x^\star), x_t - x^\star \rangle + \frac{1}{2}\langle \nabla^2f(x^\star) (x_t-x^\star), x_t - x^\star\rangle + \frac{\gamma}{6}\norm{x_t - x^\star}^3\\
	=&f(x^\star) + \frac{1}{2}\norm{x_t-x^\star}^2_{H_\star} + \frac{\gamma}{6}\norm{x_t - x^\star}^3.
	\end{align*}
	Hence, we have
	\begin{align}
	-\frac{1}{2}\norm{x_t-x^\star}^2_{H_\star} \leq -\left(f(x_t)-f(x^\star)\right) + \frac{\gamma}{6}\norm{x_t-x^\star}^3 \label{eq:H_s}
	\end{align}
	Therefore, we obtain that
	\begin{align*}
	&-\norm{\nabla f(x_t)}^2_{\TH^{-1}}\\
	\leq&- \rho\norm{x_t - x^\star}_{H}^2 + \Delta_2\\
	\leq&- \rho\left(\norm{x_t - x^\star}_{H^\star}^2+\dotprod{x_t - x^\star,(H-H^\star)(x_t - x^\star)}\right) + \Delta_2\\
	\overset{\eqref{eq:H_s}}{\leq}&-2\rho\left(f(x_t)-f(x^\star) - \frac{\gamma}{6}\norm{x_t-x^\star}^3 - \frac{1}{2}\dotprod{x_t - x^\star,(H-H^\star)(x_t - x^\star)}\right) +\Delta_2\\
	\overset{\eqref{eq:gamma_1}}{\leq}&-2\rho(f(x_t)-f(x^\star)) + \frac{4\gamma\rho}{3}\norm{x_t-x^\star}^3 + \Delta_2.
	\end{align*}
	
	Now we begin to bound the value of $\Delta_2$. First, by the condition $\nabla^2f(x) \leq (1+(1-\rho)) \TH$, we have
	\begin{align*}
	\norm{H(x_t-x^\star)}_{\TH^{-1}} \cdot \norm{\Delta_1}_{\TH^{-1}} \leq& \norm{x_t-x^\star}\cdot\norm{H\TH^{-1}H}^{1/2}\cdot\norm{\Delta_1}\cdot\norm{\TH^{-1/2}}\\
	\leq&\norm{x_t-x^\star}\cdot\sqrt{2}\norm{H^{1/2}}\cdot\frac{\gamma}{2}\norm{x_t-x^\star}^2\cdot\norm{\TH^{-1/2}}\\
	\leq&\frac{\sqrt{2}\gamma L^{1/2}}{2}\norm{x_t-x^\star}^3\cdot\norm{\TH^{-1/2}}
	.
	\end{align*}
	We also have
	\begin{align*}
	\norm{\Delta_1}_{\TH^{-1}}\cdot\norm{(H_\star- H) (x_t-x^\star)}_{\TH^{-1}} \leq& \norm{\Delta_1}\cdot\norm{\TH^{-1/2}} \cdot \gamma\norm{x_t-x^\star}\cdot\norm{x_t-x^\star}\norm{\TH^{-1/2}}\\
	\leq&\frac{\gamma^2}{2}\norm{x_t-x^\star}^4\cdot\norm{\TH^{-1}}
	.
	\end{align*}
	where the first inequality is because the condition that $\norm{H_{t} - H_{\star}}\leq \gamma\norm{x_t - x^\star}$.
	Finally, we have
	\begin{align*}
	&\norm{(H_\star- H) (x_t-x^\star)}_{\TH^{-1}}\cdot\norm{H(x_t-x^\star)}_{\TH^{-1}}\\
	\leq&\gamma\norm{x_t-x^\star}^2\cdot\norm{\TH^{-1/2}}\cdot\norm{x_t-x^\star}\cdot\norm{H\TH^{-1}H}^{1/2}\\
	\leq&\sqrt{2}\gamma L^{1/2}\norm{x_t-x^\star}^3\cdot \norm{\TH^{-1/2}}
	.
	\end{align*}
	Combining the conditions that $\norm{x_t - x^\star} \leq \frac{\rho}{\gamma} \cdot \frac{\tau \zeta^{1/2}}{L^{1/2}}$ and $\lambda_{\min}(\TH)\geq \zeta$, we have
	\begin{align}
	\Delta_2 \leq& 3\sqrt{2}\gamma L^{1/2}\norm{x_t-x^\star}^3\cdot \norm{\TH^{-1/2}}
	+\gamma^2\norm{x_t-x^\star}^4\cdot\norm{\TH^{-1}} \notag\\
	\leq& 5\gamma L^{1/2}\zeta^{-1/2}\norm{x_t-x^\star}^3+ \frac{\tau}{L} \cdot \rho\gamma L^{1/2}\zeta^{-1/2}\norm{x_t-x^\star}^3\notag\\
	\leq& 6\gamma L^{1/2}\zeta^{-1/2}\norm{x_t-x^\star}^3, \label{eq:Del_2}
	\end{align}
	where the last inequality is due to $\tau \leq L$ and $\rho \leq 1$.
	Thus, we get that
	\begin{equation}\label{eq:nab_H}
	\begin{split}
	-\norm{\nabla f(x_t)}^2_{\TH^{-1}} 
	\leq& -2\rho(f(x_t)-f(x^\star)) + \frac{4\gamma\rho}{3}\norm{x_t-x^\star}^3 +6\gamma L^{1/2}\zeta^{-1/2}\norm{x_t-x^\star}^3\\
	\leq&-2\rho(f(x_t)-f(x^\star)) + \frac{22}{3}\gamma L^{1/2}\zeta^{-1/2}\norm{x_t-x^\star}^3,
	\end{split}
	\end{equation}
	where the last inequality follows from that $\rho \leq 1$ and $\zeta \leq L$.
	%	Let us omit the higher term $\gamma^2\norm{x_t-x^\star}^4\cdot\norm{\TH^{-1}}$, then we obtain
	%	\begin{align*}
	%	\Delta_2 \leq 3\gamma L^{1/2}\norm{x_t-x^\star}^3\cdot \norm{\TH^{-1/2}}
	%	\end{align*}
	
	Now, we begin to bound the value of term $\norm{\nabla f(x_t)}_{\TH^{-1}}^3$. First, we have
	\begin{align*}
	\norm{\nabla f(x_t)}_{\TH^{-1}}^3=&\norm{\nabla f(x_t)}_{\TH^{-1}}^2\cdot\norm{\nabla f(x_t)}_{\TH^{-1}}\\
	\leq&\norm{\nabla f(x_t)}_{\TH^{-1}}^2\cdot\norm{\TH^{-1/2}} \norm{\nabla f(x_t)}\\
	\overset{\eqref{eq:L_1}}{\leq}&L\zeta^{-1/2}\norm{\nabla f(x_t)}_{\TH^{-1}}^2\cdot\norm{x_t-x^\star}.
	\end{align*}
	
	We can bound the value of $\norm{\nabla f(x)}_{\TH^{-1}}^2$ as follows:
	\begin{align*}
	&\norm{\nabla f(x_t)}_{\TH^{-1}}^2\\
	=&\norm{H(x_t - x^\star)+\Delta_1+(H_\star - H)(x_t - x^\star)}^2_{\TH^{-1}}\\
	\leq&\norm{H(x_t-x^\star)}_{\TH^{-1}}^2 + \norm{\Delta_1}^2_{\TH^{-1}}+\norm{(H_\star - H) (x_t-x^\star)}^2_{\TH^{-1}} + \Delta_2\\
	\overset{\eqref{eq:prec_cond}}{\leq}&\norm{x_t-x^\star}^2_H + 2\gamma^2\norm{\TH^{-1}}\cdot\norm{x_t - x^\star}^4+\Delta_2\\
	=&\norm{x_t-x^\star}_{H^\star}^2+\dotprod{x_t-x^\star, (H-H^\star)(x_t-x^\star)}+ 2\gamma^2\norm{\TH^{-1}}\cdot\norm{x_t - x^\star}^4+\Delta_2\\
	\overset{\eqref{eq:gamma_1}}{\leq}&\norm{x_t-x^\star}_{H^\star}^2+ \gamma \norm{x_t - x^\star}^3+ 2\gamma^2\norm{\TH^{-1}}\cdot\norm{x_t - x^\star}^4+\Delta_2\\
	\leq&\norm{x_t-x^\star}_{H^\star}^2+ 7\gamma L^{1/2}\zeta^{-1/2}\norm{x_t-x^\star}^3.
	\end{align*}
	The last inequality is because of Eqn.~\eqref{eq:Del_2}, $1\leq L^{1/2}\zeta^{-1/2}$ and the condition $\norm{x_t - x^\star} \leq \frac{\rho}{\gamma} \cdot \frac{\tau \zeta^{1/2}}{L^{1/2}}$.
	
	We also have that
	\begin{align*}
	f(x_t) \geq& f(x^\star) + \langle \nabla f(x^\star), x_t - x^\star \rangle + \frac{1}{2}\langle \nabla^2f(x^\star) (x_t-x^\star), x_t - x^\star\rangle - \frac{\gamma}{6}\norm{x_t - x^\star}^3\\
	=&f(x^\star) + \frac{1}{2}\norm{x_t-x^\star}^2_{H_\star} - \frac{\gamma}{6}\norm{x_t - x^\star}^3.
	\end{align*}
	Therefore, we have
	\begin{align}
	\norm{\nabla f(x)}_{\TH^{-1}}^3 \leq &L\zeta^{-1/2}\norm{\nabla f(x_t)}_{\TH^{-1}}^2\cdot\norm{x_t-x^\star}\notag\\
	\leq&L\zeta^{-1/2}\cdot\norm{x_t-x^\star} \left(2(f(x_t) - f(x^\star) + \frac{\gamma}{3}\norm{x_t - x^\star}^3 + 7\gamma L^{1/2}\zeta^{-1/2}\norm{x_t-x^\star}^3\right)\notag\\
	\leq&2L\zeta^{-1/2}\norm{x_t-x^\star}\cdot(f(x_t)-f(x^\star))+15\gamma L^{3/2}\zeta^{-1}\norm{x_t-x^\star}^4, \label{eq:nab3}
	\end{align}
	where the last inequality is due to the fact that $1\leq L^{1/2}\zeta^{-1/2}$.
	
	Therefore, we can obtain that
	\begin{align*}
	&\EB_u[f(x_{t+1}) - f(x^\star)] \\
	\leq& f(x_t) -f(x^\star) - \frac{b}{64(d+2)}\norm{ \nabla f(x_t)}_{\TH^{-1}}^2+\frac{b^3\gamma }{64}d^{-3/2}\norm{\TH^{-1}}^{3/2}\cdot\norm{\nabla f(x)}_{\TH^{-1}}^3+\Delta_\mu\\
	\overset{\eqref{eq:nab_H}}{\leq}& f(x_t) -f(x^\star) - \frac{b\rho}{32(d+2)}(f(x_t) -f(x^\star)) + \frac{2b\gamma L^{1/2}\zeta^{-1/2}}{3(d+2)}\norm{x_t-x^\star}^3\\
	&+\frac{b^3\gamma }{64}d^{-3/2}\norm{\TH^{-1}}^{3/2}\cdot\norm{\nabla f(x)}_{\TH^{-1}}^3+\Delta_\mu\\
	\overset{\eqref{eq:nab3}}{\leq}&\left(1 - \frac{b\rho}{32(d+2)}\right)\left(f(x_t) -f(x^\star) \right)+ \frac{2b\gamma L^{1/2}\zeta^{-1/2}}{3(d+2)}\norm{x_t-x^\star}^3\\
	&+\frac{b^3\gamma L\zeta^{-2}}{64d^{3/2}}\norm{x_t-x^\star}\cdot(f(x_t)-f(x^\star))+\frac{b^3\gamma^2L^{3/2}\zeta^{-5/2}}{4d^{3/2}}\norm{x_t-x^\star}^4+\Delta_\mu.
	\end{align*}
	
	Since the objective function is $\tau$-strongly convex, by Lemma~\ref{lem:str_cvx}, we have
	\begin{align*}
	\norm{x_t-x^\star}^3 \leq \frac{2}{\tau}\cdot \norm{x_t-x^\star}\cdot\left(f(x_t)-f(x^\star)\right)
	,
	\end{align*}
	and by the condition of $\norm{x_t - x^\star} \leq \frac{\rho}{\gamma} \cdot \frac{\tau \zeta^{1/2}}{L^{1/2}}$, we also have
	\begin{align*}
	\frac{b^3\gamma^2L^{3/2}\zeta^{-5/2}}{4d^{3/2}}\norm{x_t-x^\star}^4 \leq& \frac{b^3\gamma^2L^{3/2}\zeta^{-5/2}}{4d^{3/2}} \cdot \frac{2}{\tau} \norm{x_t - x^\star}^2 \cdot\left(f(x_t)-f(x^\star)\right)\\
	\leq& \frac{b^3\gamma L \zeta^{-2}}{2d^{3/2}} \norm{x_t - x^\star} \cdot\left(f(x_t)-f(x^\star)\right)
	.
	\end{align*}
	
	Thus, we have
	\begin{align*}
	&\EB_u\left[f(x_{x+1})-f(x^\star)\right]\\
	\leq& \left(1 - \frac{b\rho}{32(d+2)}\right)\left(f(x_t) -f(x^\star) \right) + \frac{4b\gamma L^{1/2}\zeta^{-1/2}}{3\tau(d+2)}\norm{x_t-x^\star}\cdot\left(f(x_t)-f(x^\star)\right) \\
	&+\left(\frac{b^3\gamma L\zeta^{-2}}{64d^{3/2}}+ \frac{b^3\gamma L \zeta^{-2}}{2d^{3/2}} \right)\norm{x_t-x^\star}\cdot(f(x_t)-f(x^\star))+\Delta_\mu\\
	=& \left(1 - \frac{b\rho}{32(d+2)}\right)\left(f(x_t) -f(x^\star) \right) + b\gamma\cdot \Delta_3\cdot\norm{x_t-x^\star}\cdot\left(f(x_t) -f(x^\star) \right)+\Delta_\mu,
	\end{align*}
	where $\Delta_3$ is defined as 
	\begin{align*}
	\Delta_3 = \frac{ 4L^{1/2}\zeta^{-1/2}}{3\tau(d+2)}+\frac{33b^2 L\zeta^{-2}}{64d^{3/2}}.
	\end{align*}
	
	To keep a fast convergence rate, we need
	\begin{align*}
	\norm{x_t-x^\star} \leq& \frac{1}{2\gamma\Delta_3}\frac{\rho}{32(d+2)}\\
	\leq& \frac{\rho}{\gamma}\cdot\min\left(\frac{3\tau\zeta^{1/2}}{ 256 L^{1/2}}, \frac{d^{3/2}\zeta^{2}}{ 33L(d+2)b^2}\right)
	.
	\end{align*}
	Thus, when $x_t$ satisfies the above condition, we can obtain that
	\begin{align*}
	\EB_u\left[f(x_{x+1})-f(x^\star)\right] \leq \left(1 - \frac{b\rho}{64(d+2)}\right)\left(f(x_t) -f(x^\star) \right) +\Delta_\mu.
	\end{align*}
\end{proof}

\subsection{Proof of Theorem~\ref{thm:glb}}
Now, we give the global convergence property of Algorithm~\ref{alg:zero_order} in Theorem~\ref{thm:glb}.

\begin{proof}[Proof of Theorem~\ref{thm:glb}]
	Taking a random step from $x_t$, we have
	\begin{align}
	&\EB_u\left[f(x_{t+1})\right]\notag\\
	=&\EB_u\left[f(x_{t} - \eta\TH^{-1}g_\mu(x_t))\right]\notag\\
	\overset{\eqref{eq:L_1}}{\leq}&\EB_u\left[f(x_t)-\eta\dotprod{\nabla f(x_t), \TH^{-1}\EB_u[g_\mu(x_t)]}+\frac{\eta^2L}{2}\EB_u\norm{\TH^{-1}g_\mu(x_t)}^2\right]\notag\\
	\leq&f(x_t) - \eta\norm{\nabla f(x_t)}_{\TH^{-1}}^2 -\eta\dotprod{\nabla f(x_t), \TH^{-1}\left(\EB_u[g_\mu(x_t)] - \nabla f(x_t)\right)}+\frac{\eta^2L\norm{\TH^{-1}}}{2}\EB_u\norm{g_\mu(x_t)}_{\TH^{-1}}^2\notag\\
	\leq&f(x_t) - \eta\norm{\nabla f(x_t)}_{\TH^{-1}}^2+\frac{\eta}{2}\left(\norm{\nabla f(x_t)}_{\TH^{-1}}^2+\norm{\EB_u[g_\mu(x_t)] - \nabla f(x_t)}_{\TH^{-1}}^2\right)\label{eq:cauchy_2}\\
	&+\frac{\eta^2L\norm{\TH^{-1}}}{2}\EB_u\norm{g_\mu(x_t)}_{\TH^{-1}}^2\notag\\
	\leq&f(x_t) - \frac{\eta}{2}\norm{\nabla f(x_t)}_{\TH^{-1}}^2+\frac{\eta \mu^2L^2}{8}\norm{\TH^{-1}}^2(d+3)^{3} + \frac{\eta^2L\norm{\TH^{-1}}}{2}\EB_u\norm{g_\mu(x_t)}_{\TH^{-1}}^2\label{eq:grad_err}\\
	\leq&f(x_t) - \frac{\eta}{2}\norm{\nabla f(x_t)}_{\TH^{-1}}^2+\frac{\eta \mu^2L^2}{8}\norm{\TH^{-1}}^2(d+3)^{3} \notag\\
	&+ \frac{\eta^2L\norm{\TH^{-1}}}{2b}\left(\frac{\mu^2}{2}L^2\norm{\TH^{-1}}(d+6)^3 + 2(d+2)\norm{\nabla f(x)}_{\TH^{-1}}^2\right)\label{eq:T_2}\\
	=&f(x_t) - \frac{\eta}{2}\norm{\nabla f(x_t)}_{\TH^{-1}}^2+ \eta^2L(d+2)b^{-1}\norm{\TH^{-1}}\norm{\nabla f(x_t)}_{\TH^{-1}}^2\notag\\
	&+\frac{\eta \mu^2L^2}{8}\norm{\TH^{-1}}^2(d+3)^{3} + \frac{\eta^2L^3\mu^2\norm{\TH^{-2}}}{4b}(d+6)^3\notag
	\end{align}
	Inequality~\eqref{eq:cauchy_2} is because of Cauchy's inequality and $2ab \leq a^2+b^2$. Inequality~\eqref{eq:grad_err} follows from Lemma~\ref{lem:T_1}. and inequality~\eqref{eq:T_2} is due to Lemma~\ref{lem:T_2} with $B=\TH^{-1}$.
	
	Since $\zeta I \preceq \TH$ holds for all iterations, let us set the step size $\eta$ as
	\begin{align*}
	\eta = \frac{b\zeta}{4(d+2)L}.
	\end{align*}
	Thus we obtain that
	\begin{align*}
	\EB_u\left[f(x_{t+1})\right] \leq f(x_t) - \frac{b\zeta}{16(d+2)L}\norm{\nabla f(x_t)}_{\TH^{-1}}^2 +\Delta_\mu,
	\end{align*}
	where $\Delta_\mu$ denotes that
	\begin{align*}
	\Delta_\mu = \frac{b\mu^2L}{32(d+2)\zeta}(d+3)^2+\frac{b\mu^2L}{64\zeta(d+2)^2}(d+6)^3=\frac{b\mu^2L}{32\zeta(d+2)}\left((d+3)^3+\frac{(d+6)^3}{2(d+2)}\right).
	\end{align*}
	
	By Lemma~\ref{lem:str_cvx}, we have
	\begin{align*}
	\EB_u\left[f(x_{t+1})-f(x^\star)\right] \leq& f(x_t) -f(x^\star)- \frac{\zeta}{16(d+2)L^2}\norm{\nabla f(x_t)}^2 +\Delta_\mu\\
	\leq&f(x_t) -f(x^\star) - \frac{\zeta\tau}{16(d+2)L^2}(f(x_t) -f(x^\star))+\Delta_\mu\\
	=&\left(1-\frac{\zeta}{16(d+2)\kappa L}\right)\cdot(f(x_t) -f(x^\star))+\Delta_\mu
	\end{align*}
\end{proof}

\pb\section{Proof of Section~\ref{subsec:qca}}
\subsection{Proof of Theorem~\ref{thm:lw_app}}

Our proof of Theorem~\ref{thm:lw_app} relies on the existing results about noisy power method depicted in Algorithm~\ref{alg:noisy}. 
Because of the approximation, we can cast our zeroth-order power method as the noisy power method and the product $H_\mu V_t$ can be regarded as $H_\mu V_t = \nabla^2f(x) V_t + G_t$ with $G_t$ being the noise matrix. 
\citet{balcan2016improved} showed that the noisy power method converges to the principal eigenvalues and eigenvectors of the matrix and have the follow properties.

\begin{lemma} [\cite{balcan2016improved}]\label{lem:npw}
	Given positive definite matrix $H\in \RB^{d\times d}$ and $0<\epsilon<1$, suppose that noise matrix satisfies 
	\begin{align*}
	\|G_t\| \leq C_1\epsilon^2\lambda_{k+1} \quad\mbox{and}\quad \|V_k^TG_t\|_2 = C_2\epsilon^2\lambda_{k+1}\cos\left(V_k, X_t \right).
	\end{align*}
	Then, after $T$ iterations, $V_T$ returned by Algorithm~\ref{alg:noisy} satisfies
	\[
	\norm{H - V_TV_T^TH}\leq(1+\epsilon)\norm{H - H_k}, \;\;\mbox{if}\quad T = \frac{C_3}{\epsilon}\log\left(\frac{\tan(U_k, V_0)}{\epsilon}\right).
	\]
	where $H_k = U_k \Lambda_k U_k^\top$ is the best rank-$k$ approximation of $H$. $C_1$, $C_2$, and $C_3$ are absolute constants.
\end{lemma}

\begin{proof}[Proof of Theorem~\ref{thm:lw_app}] First, we us denote $H = \nabla^2 f(x)$. By the definition of $V_T$ and $\hat{V}$ in Algorithm~\ref{alg:lw_app}, we have
	\begin{align*}
	\norm{H-V\Lambda V^\top} \leq& \norm{H-V_TV_T^\top H V_TV_T^\top}+\norm{V_TV_T^\top H V_TV_T^\top - V\Lambda V^\top}\\
	\leq&\norm{H-V_TV_T^\top H}+\norm{V_TV_T^\top H - V_TV_T^\top H V_TV_T^\top} + \norm{V_TV_T^\top HV_TV_T^\top - V_T \hat{V} \Lambda\hat{V}^\top V_T^\top} \\
	\leq&2\norm{H - V_TV_T^\top H} + \norm{V_TV_T^\top HV_TV_T^\top - V_T \hat{V} \Lambda\hat{V}^\top V_T^\top}\\
	\leq&3\lambda_{k+1} + \norm{V_TV_T^\top HV_TV_T^\top - V_T \hat{V} \Lambda\hat{V}^\top V_T^\top},
	\end{align*}
	where the last inequality is because of Lemma~\ref{lem:npw} with $\epsilon = 1/2$.
	
	Furthermore, by the step~\ref{step:6} of Algorithm~\ref{alg:lw_app}, we have
	\begin{align*}
	\norm{V_TV_T^\top HV_TV_T^\top - V_T \hat{V} \Lambda\hat{V}^\top V_T^\top}
	=&\norm{V_T^\top HV_T - \hat{V} \Lambda\hat{V}^\top}\\
	=&\norm{V_T^\top HV_T-\hat{V} U^\top H_\mu V_T}\\
	\leq&\norm{V_T^\top - \hat{V} U^\top}\cdot\norm{H V_T - H_\mu V_T}\\
	\leq&2\norm{H V_T - H_\mu V_T},
	\end{align*} 
	where the last inequality is because of 
	$\norm{V_T^\top - \hat{V} U^\top} \leq \norm{V_T} + \norm{\hat{V} U^\top} \leq 2$.
	
	Now, we begin to bound the value of $\norm{H V_T - H_\mu V_T}$. Let $v$ be a unit vector, we have 
	\begin{align*}
	\norm{\frac{\nabla f(x+\mu_1 \cdot v) - \nabla f(x)}{\mu_1} - \nabla^2f(x)v}
	=&\norm{\frac{\nabla f(x+\mu_1 \cdot v) - \nabla f(x) - \mu_1 \nabla^2f(x)v}{\mu_1}}\\
	=&\norm{\frac{\mu_1 \cdot v \left(\nabla^2 f(\tilde{x}) - \nabla^2 f(x)\right)}{\mu_1}}\\
	\leq&\mu_1\cdot \gamma\norm{v}^2=\gamma\mu_1, 
	\end{align*} 
	where the last inequality is because of Eqn.~\eqref{eq:gamma_1}
	and $\tilde{x} = x + \theta\cdot(\mu_1 v) \; \mbox{with}\; 0\leq\theta\leq 1$.
	
	Let $\ti{\nabla} f(x)$ denote the approximate gradient computed as follows:
	\begin{align*}
		\ti{\nabla}_j f(x)= \frac{f(x+\mu_1\cdot e_j)-f(x-\mu_1\cdot e_j)}{2\mu_1}.
	\end{align*}
	We also use $[]$
	Then, we have 
	\begin{align*}
	&\norm{\nabla f(x) - \ti{\nabla} f(x)} \\
	=&\norm{\left[\tilde{\nabla}_1 f(x) - \nabla_1 f(x),\dots,\tilde{\nabla}_d f(x) - \nabla_d f(x)\right]^\top} \\
	=&\frac{1}{2\mu_1}\cdot\bigg\|\big[\left(f(x+\mu_1 \cdot e_1) - f(x-\mu_1\cdot e_1) - 2\mu_1\nabla_j f(x)\right),\dots, \\
	&\qquad\left(f(x+\mu_1 \cdot e_d) - f(x-\mu_1\cdot e_d) - 2\mu_1\nabla_d f(x)\right)\big]^\top\bigg\|\\
	=&\norm{\frac{\mu_1^2\cdot\diag\left(\nabla^2 f(\ti{x}_1) - \nabla^2(\ti{x}_2)\right)}{4\mu_1}}\\
	\overset{\eqref{eq:gamma_1}}{\leq}&\frac{\mu_1}{4}\gamma\norm{\tilde{x}_1 - \tilde{x}_2}\\
	\leq&\frac{\gamma\mu_1^2}{2},
	\end{align*}
	where the last inequality is because of $\tilde{x}_1 = x + \theta_1\cdot(\mu_1 v)$ with $0\leq\theta_1\leq 1$ and $\tilde{x}_2 = x - \theta_2\cdot(\mu_1 v)$ with $0\leq\theta_2\leq 1$.
	
	Combining the above results, we have
	\begin{align*}
	\norm{H V_T - H_\mu V_T}=&\norm{H[v_1, v_2,\dots,v_k]-H_\mu[v_1, v_2,\dots,v_k]}\\
	\leq&k\norm{Hv_1 - H_\mu v_1}\\
	=&k\norm{\frac{\ti{\nabla}f(x+\mu_1 \cdot v) - \ti{\nabla}f(x)}{\mu_1}- \nabla^2f(x)v}\\
	\leq&k\norm{\frac{\nabla f(x+\mu_1 \cdot v) - \nabla f(x)}{\mu_1}- \nabla^2f(x)v}\\
	&+k\norm{\frac{\ti{\nabla}f(x+\mu_1 \cdot v) -\nabla f(x+\mu_1 \cdot v) + \nabla f(x)- \ti{\nabla}f(x)}{\mu_1}}\\
	\leq&k\left(\mu_1\gamma\norm{v_1}^2 + 2\frac{\gamma\mu_1^2}{2\mu_1}\right)\\
	\leq&2k\mu_1\gamma.
	\end{align*}
	
	Therefore, we have
	\begin{align*}
	\norm{H-V\Lambda V^\top} \leq 3\lambda_{k+1} + 2\norm{H V_T - H_\mu V_T} \leq 5 \lambda_{k+1}
	\end{align*}
	where the last inequality is because that we set $\mu$ and $\mu_1$ as follows
	\begin{align*}
	k\gamma\mu_1+2\sqrt{d}L\frac{\mu}{\mu_1} \leq 2\lambda_{k+1}
	\end{align*}
	
	Furthermore, we can obtain that
	\begin{align*}
	&\norm{H-VV^\top H_\mu VV^\top} \leq 5\lambda_{k+1}\\
	\Rightarrow& -5\lambda_{k+1}\norm{y}^2\leq y^\top\left(H-VV^\top H_\mu VV^\top\right)y \leq 5\lambda_{k+1}\norm{y}^2\\
	\Rightarrow&-10\lambda_{k+1}\norm{y}^2 \leq y^\top\left(H - \TH\right)y \leq 0\\
	\Rightarrow&y^\top \TH y -6\lambda_{k+1}\norm{y}^2 \overset{(a)}{\leq} y^\top H y \overset{(b)}{\leq} y^\top \TH y 
	\end{align*}
	Let us consider the $\overset{(a)}{\leq}$ case and have
	\begin{align*}
	&y^\top \TH y -10\lambda_{k+1}\norm{y}^2 \leq y^\top H y\\
	\Rightarrow&y^\top \TH y \leq y^\top H y+ 10\lambda_{k+1}\norm{y}^2 \leq \left(1+\frac{10\lambda_{k+1}}{\lambda_{\min}}\right)y \leq y^\top H y\\
	\Rightarrow&\left(1-\frac{10\lambda_{k+1}}{\lambda_{\min} + 10\lambda_{k+1}}\right)\TH\preceq H
	\end{align*}
	Therefore, we have
	\begin{align*}
	\left(1-\frac{10\lambda_{k+1}}{\lambda_{\min} + 10\lambda_{k+1}}\right)\TH\preceq H \preceq \TH.
	\end{align*}
\end{proof}

\subsection{Proof of Theorem~\ref{thm:query}}

\begin{proof}[Proof of Theorem~\ref{thm:query}]
	By Theorem~\ref{thm:lw_app}, it takes $O\left(dk\cdot\log(\tan(U_k, V_0))\right)$ queries to construct the approximate Hessian. We also need $O(d)$ queries to estimate gradient. By Theorem~\ref{thm:local} and~\ref{thm:lw_app}, we need $O\left(\frac{d\lambda_{k+1}}{b\tau}\log\left(\frac{1}{\epsilon}\right)\right)$ iterations to achieve an $\epsilon$-accuracy. Hence, we have
	\begin{align*}
	Q(\epsilon) =& O\left(\frac{d\lambda_{k+1}}{b\tau} b\log\left(\frac{1}{\epsilon}\right) + \frac{d\lambda_{k+1}}{b\tau }\cdot dk\cdot\tan(U_k, V_0)\log\left(\frac{1}{\epsilon}\right)\right)\\
	=&O\left(\frac{dk\lambda_{k+1}}{\tau }\cdot\log(\tan(U_k, V_0))\cdot\log\left(\frac{1}{\epsilon}\right)\right),\\
	=&\ti{O}\left(\frac{dk\lambda_{k+1}}{\tau }\cdot\log\left(\frac{1}{\epsilon}\right)\right).
	\end{align*}
\end{proof}

\begin{algorithm}[tb]
	\caption{Noisy Power Method.}
	\label{alg:noisy}
	\begin{small}
		\begin{algorithmic}[1]
			\STATE {\bf Input:} A positive definite matrix $H \in \RB^{d\times d}$, orthonormal matrix $X\in\RB^{d\times p}$, target rank $k$;
			\FOR {$t=0,\dots$ until termination}
			\STATE $Y_t = HX_t + G_t$ for some noise matrix $G_t$;
			\STATE QR factorization: $Y_ t= X_{t+1}R_{t+1}$, where $V_{t+1}$ consists of orthonormal columns.
			\ENDFOR
		\end{algorithmic}
	\end{small}
\end{algorithm}	

\section{Proof of Section~\ref{sec:H_app}}

\subsection{Proof of Lemma~\ref{lem:diff_H}}
\begin{proof}[Proof of Lemma~\ref{lem:diff_H}] By the definition of $f_\mu(x)$, we have
	\begin{align*} 
	\norm{\nabla^2f_\mu(x) - \nabla^2f(x)} =& \norm{\nabla^2\EB\left[f(x+\mu u)\right] - \nabla^2f(x)}\\
	=&\norm{\EB\left[\nabla^2f(x+\mu u)\right] - \nabla^2f(x)}\\
	\leq&\EB\left[\norm{\nabla^2f(x+\mu u) - \nabla^2f(x)}\right]\\
	\leq&\gamma\mu\EB\norm{u}\\
	\leq&\gamma\mu(d+1)^{1/2}.
	\end{align*}
	The first inequality is due to Jensen's inequality.  The second inequality is by Eqn.~\eqref{eq:gamma_1}. And the last inequality follows from Lemma~\ref{lem:gauss_power}.
\end{proof}

\subsection{Proof of Lemma~\ref{lem:H_mu}}
\begin{proof}[Proof of Lemma~\ref{lem:H_mu}]
	To get a convenient expression, we rewrite Eqn.~\eqref{eq:f_mu} in another form by introducing a new integration variable $y = x +\mu u$:
	\[
	f_\mu(x) = \frac{1}{\mu^d M}\int_{\RB^d}f(y)\;\exp{\left(-\frac{1}{2\mu^2}\norm{y-x}^2\right)} \,dy.
	\]
	Then, its gradient can be written as, 
	\begin{align*}
	\nabla f_\mu(x) = \frac{1}{\mu^{d+2}M} \int_{E} f(y)\;\exp{\left(-\frac{1}{2\mu^2}\norm{y-x}^2\right)} (y-x) \,dy.
	\end{align*}
	Then, we have
	\begin{align}
	\nabla^2 f_\mu(x) =& - \frac{1}{\mu^{d+2}M} \int_{E} f(y)\;\exp{\left(-\frac{1}{2\mu^2}\norm{y-x}^2\right)} \,dy \notag\\
	&+\frac{1}{\mu^{d+4}M} \int_{E} f(y)\;\exp{\left(-\frac{1}{2\mu^2}\norm{y-x}^2\right)} \cdot (y-x)(y-x)^\top\,dy\notag\\
	=&-\left[\frac{1}{\mu^2 M}\int_{E}f(x+\mu u) \;\exp{\left(-\frac{1}{2}\norm{u}^2\right)} \,du \right]I_d + \frac{1}{\mu^2 M} \int_E f(x+\mu u) u u^\top \;\exp{\left(-\frac{1}{2}\norm{u}^2\right)} \,du\notag\\
	=&\frac{1}{\mu^2 M}\int_E \left[f(x+\mu u) -f(x)\right] uu^\top \;\exp{\left(-\frac{1}{2}\norm{u}^2\right)} \,du + \frac{1}{\mu^2} (f(x) - f_\mu(x)) I_d. \label{eq:H_1}
	\end{align}
	
	Furthermore, $f_\mu(x)$ can also be defined as 
	\[
	f_\mu(x) = \frac{1}{M}\int_{\RB^d} f(x-\mu u)\;\exp{\left(-\frac{1}{2}\norm{u}^2\right)}\,du.
	\]
	Similarly, by introducing $y = x - \mu u$, we have
	\[
	f_\mu(x) = -\frac{1}{\mu^d M}\int_{\RB^d}f(y)\;\exp{\left(-\frac{1}{2\mu^2}\norm{y-x}^2\right)} \,dy.
	\]
	And its gradient is,
	\begin{align*}
	\nabla f_\mu(x) = -\frac{1}{\mu^{d+2}M} \int_{E} f(y)\;\exp{\left(-\frac{1}{2\mu^2}\norm{y-x}^2\right)} (y-x) \,dy.
	\end{align*}
	Then, we have
	\begin{align}
	\nabla^2 f_\mu(x) =& \frac{1}{\mu^{d+2}M} \int_{E} f(y)\;\exp{\left(-\frac{1}{2\mu^2}\norm{y-x}^2\right)} \,dy \notag\\
	&-\frac{1}{\mu^{d+4}M} \int_{E} f(y)\;\exp{\left(-\frac{1}{2\mu^2}\norm{y-x}^2\right)} \cdot (y-x)(y-x)^\top\,dy \notag\\
	=&-\left[\frac{1}{\mu^2 M}\int_{E}f(x-\mu u) \;\exp{\left(-\frac{1}{2}\norm{u}^2\right)} \,du \right]I + \frac{1}{\mu^2 M} \int_E f(x-\mu u) u u^\top \;\exp{\left(-\frac{1}{2}\norm{u}^2\right)} \,du\notag\\
	=&\frac{1}{\mu^2 M}\int_E \left[f(x-\mu u) -f(x)\right] uu^\top \;\exp{\left(-\frac{1}{2}\norm{u}^2\right)} \,du + \frac{1}{\mu^2} (f(x) - f_\mu(x)) I_d. \label{eq:H_2}
	\end{align}
	
	Combining the Eqn.~\eqref{eq:H_1} and~\eqref{eq:H_2}, we have
	\begin{align*}
	\nabla^2f_\mu(x) =& \frac{1}{2}\left(\nabla^2f_\mu(x) + 	\nabla^2f_\mu(x)\right)\\
	=&\frac{1}{M}\int_{\RB^d} \frac{f(x+\mu u) + f(x-\mu u) -2f(x) }{2\mu^2} uu^\top \;\exp{\left(-\frac{1}{2}\norm{u}^2\right)} \,du +\frac{f(x) - f_\mu(x) }{\mu^2} \cdot I_d.
	\end{align*}
	Therefore, we have $$ \EB_u[\TH] = \nabla^2f_\mu(x) + \left(\lambda - \frac{f(x) - f_\mu(x) }{\mu^2}\right) \cdot I_d.$$
	
	The inequality $\nabla^2f_\mu(x) \preceq \EB_u[\TH]$ is because of $f_\mu(x) \geq f(x)$ which is implied by
	\begin{align*}
		f_\mu(x) = \EB [f(x+\mu u)] \geq \EB \left[f(x) +\mu \dotprod{\nabla f(x), u}\right] = f(x)
	\end{align*}
	where the inequality is because of convexity of $f(x)$.
\end{proof}

\pb\section{Parameter Setting}

\begin{table*}
	\centering
	\caption{Parameter settings in MNIST experiments}
	\label{tb:mnist_para}
	\begin{tabular}{c|lcccc}
		\hline
		& Algorithm 		&~~~~ $b$ ~~~~&~~~~ $\mu$ ~~~~&~~~~ lr ~~~~&~~~~ $\nu$~~~~\\ \hline
		\multirow{6}*{targeted} 
		&\texttt{ZOO}  		& 100 & 0.01  &0.5 & -\\
		&\texttt{PGD-NES}  	& 100 & 0.05  &0.02 & -  \\
		&\texttt{ZOHA-Gauss}& 100 & 0.01 & 0.04 & -  \\
		&\texttt{ZOHA-Gauss-DC} &50-200 & 0.01 & 0.04 & -  \\
		&\texttt{ZOHA-Diag}         & 100			& 0.1  & 0.04 &0.85 \\
		&\texttt{ZOHA-Diag-DC}      & 50-200	  	& 0.1  & 0.04 & 0.85\\\hline
		\multirow{6}*{un-targeted} 
		&\texttt{ZOO}  		& 100 & 0.01  &0.5 & - \\
		&\texttt{PGD-NES}  	& 100 & 0.05 &0.02 & -  \\ 
		&\texttt{ZOHA-Gauss}& 100		& 0.01 &0.04 & - \\
		&\texttt{ZOHA-Gauss-DC} & 50-200& 0.01 &0.04 & - \\
		&\texttt{ZOHA-Diag}     & 100      & 0.1 & 0.04  &0.8\\
		&\texttt{ZOHA-Diag-DC}  & 50-200 & 0.1 & 0.04 & 0.8 \\\hline
	\end{tabular}
\end{table*}

\begin{table*}
	\centering
	\caption{Parameter settings in ImageNet experiments}
	\label{tb:imagenet_para}
	\begin{tabular}{c|lcccc}
		\hline
		& Algorithm 		&~~~~ $b$ ~~~~&~~~~ $\mu$ ~~~~&~~~~ lr ~~~~&~~~~ $\nu$~~~~\\ \hline
		\multirow{6}*{targeted} 
		&\texttt{ZOO}  		& 50 & 0.01  &2 & -\\
		&\texttt{PGD-NES}  	& 50 & 0.01  &0.01 & -  \\
		&\texttt{ZOHA-Gauss}& 50 & 0.05 & 0.03 & -  \\
		&\texttt{ZOHA-Gauss-DC} &50-200 & 0.05 & 0.03 & -  \\
		&\texttt{ZOHA-Diag}         & 50			& 0.03  & 0.015 &0.8 \\
		&\texttt{ZOHA-Diag-DC}      & 50-200	  	& 0.03  & 0.015 & 0.8\\\hline
		\multirow{6}*{un-targeted} 
		&\texttt{ZOO}  		& 50 & 0.01  &0.5 & - \\
		&\texttt{PGD-NES}  	& 50 & 0.01 &0.01 & -  \\ 
		&\texttt{ZOHA-Gauss}& 50		& 0.02 &0.02 & - \\
		&\texttt{ZOHA-Gauss-DC} & 50-200& 0.02 &0.02 & - \\
		&\texttt{ZOHA-Diag}     & 50      & 0.03 & 0.03  &0.7\\
		&\texttt{ZOHA-Diag-DC}  & 50-200 & 0.03 & 0.03 & 0.7 \\\hline
	\end{tabular}
\end{table*}

\pb\section{Visualization of Adversarial Attack}\label{app:vis_att}

\begin{figure}[!ht]
	\subfigtopskip = 0pt
	\begin{center}
		\subfigure[\textsf{Targeted}]{\includegraphics[width=85mm]{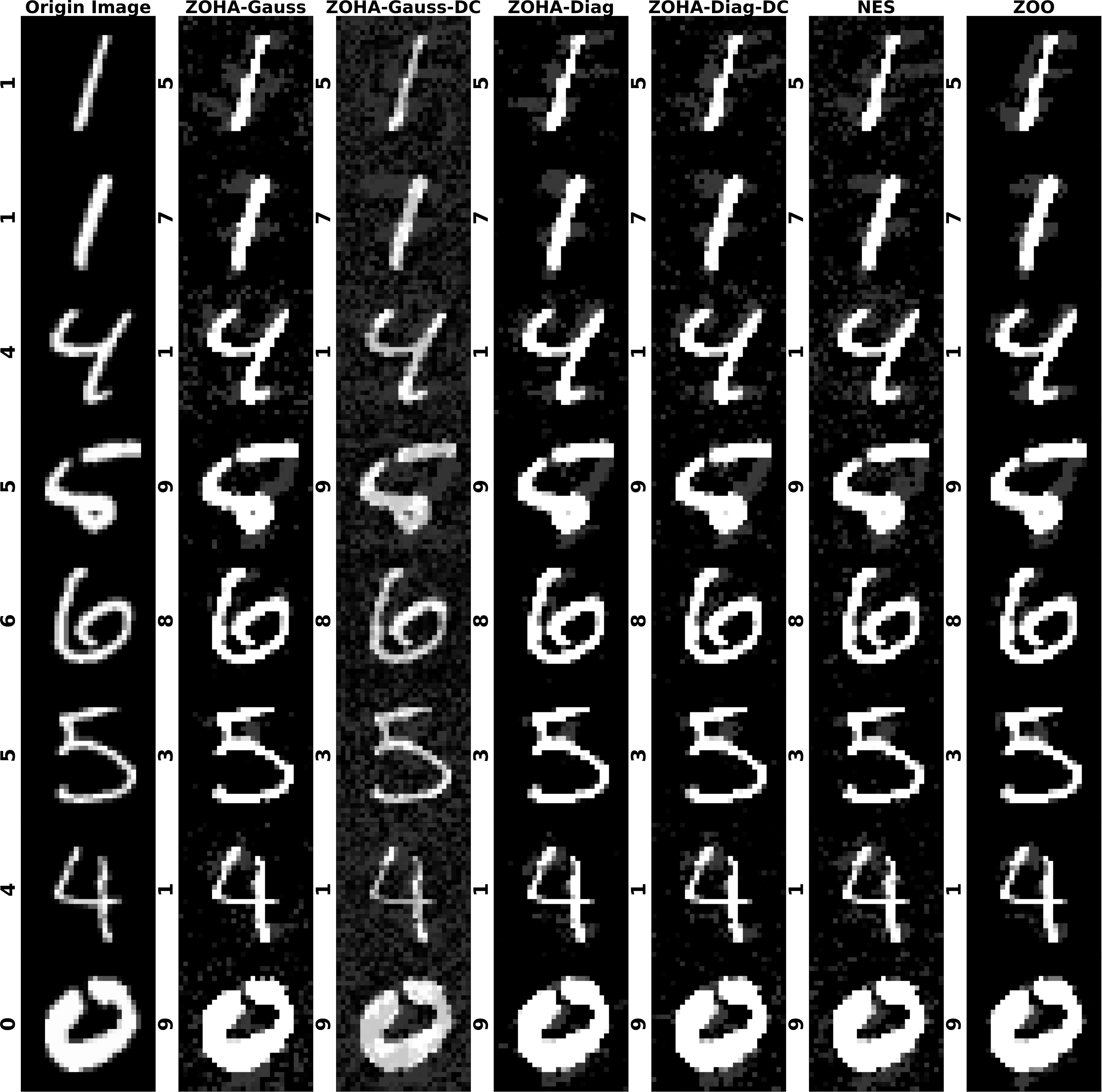}}~
		\subfigure[\textsf{ Un-targeted}]{\includegraphics[width=85mm]{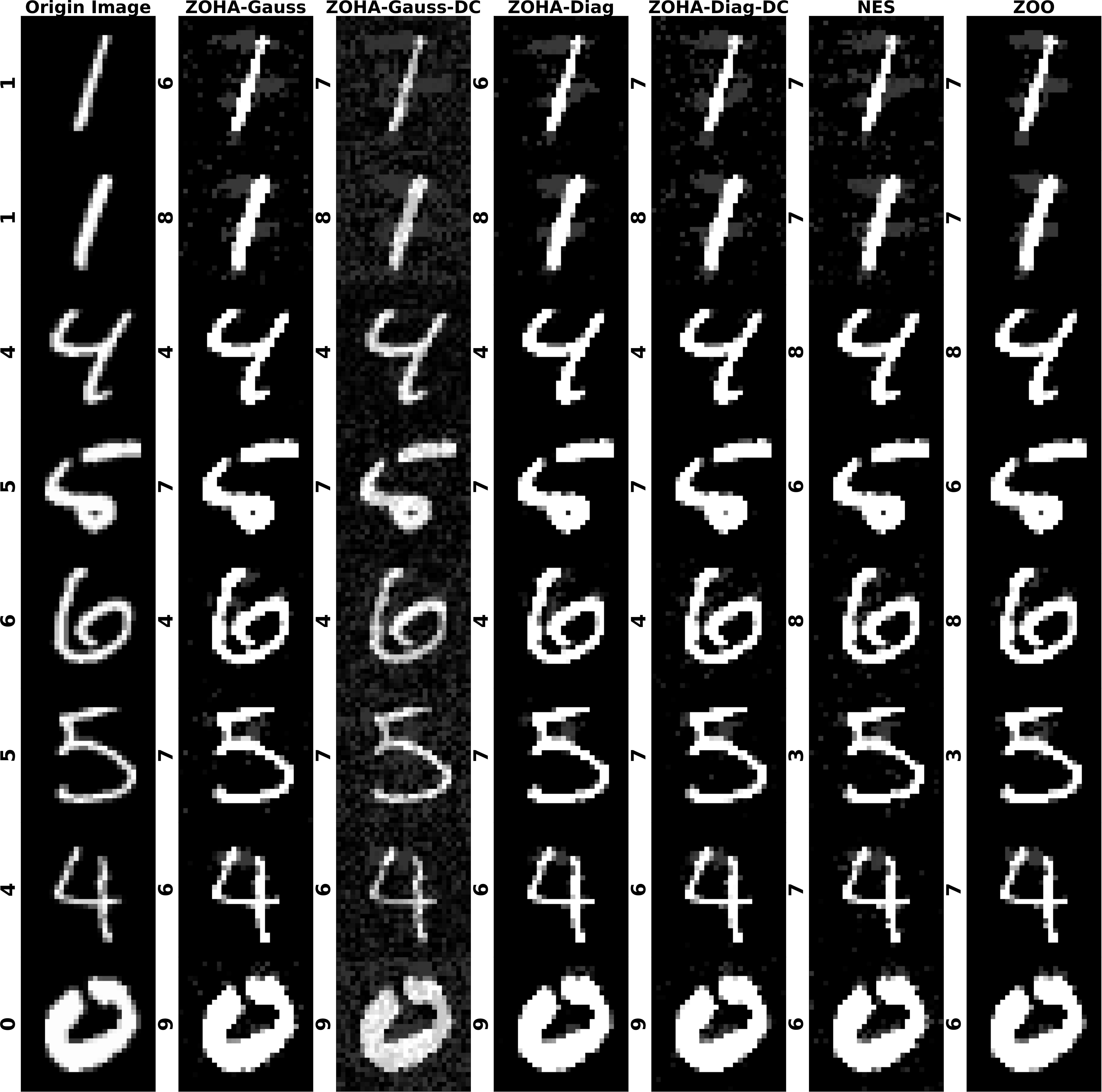}}
	\end{center}
	\caption{Adversarial examples on MNIST. The label is denoted on the left side.}
	\label{fig:mnist_att}
\end{figure}

\begin{figure}[!ht]
	\subfigtopskip = 0pt
	\begin{center}
		\includegraphics[width=170mm]{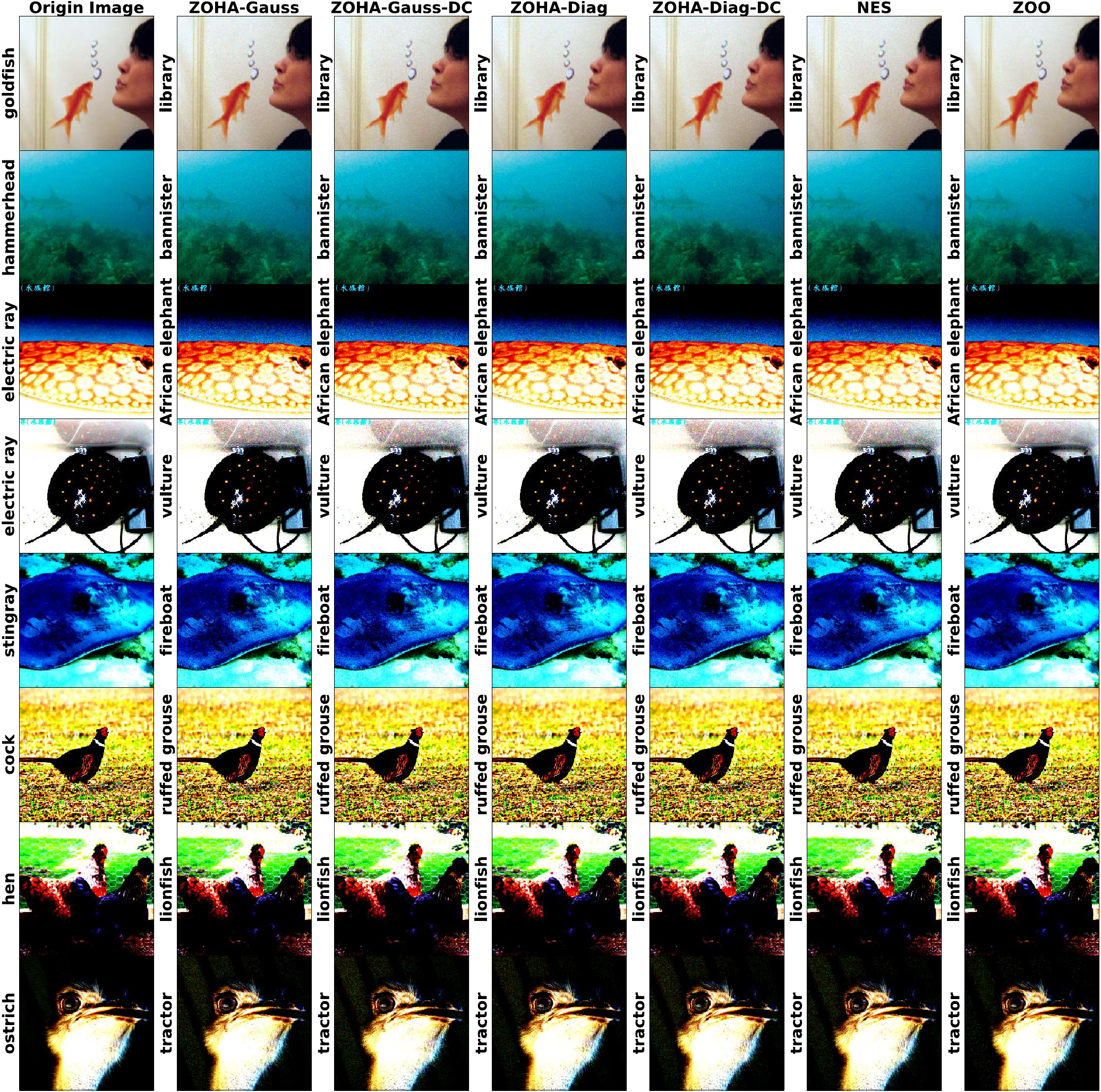}
	\end{center}
	\caption{Targeted adversarial examples on ResNet50 and ImageNet. The label is denoted on the left side.}
	\label{fig:imagenet_tar}
\end{figure}

\begin{figure}[!ht]
	\subfigtopskip = 0pt
	\begin{center}
	\includegraphics[width=155mm]{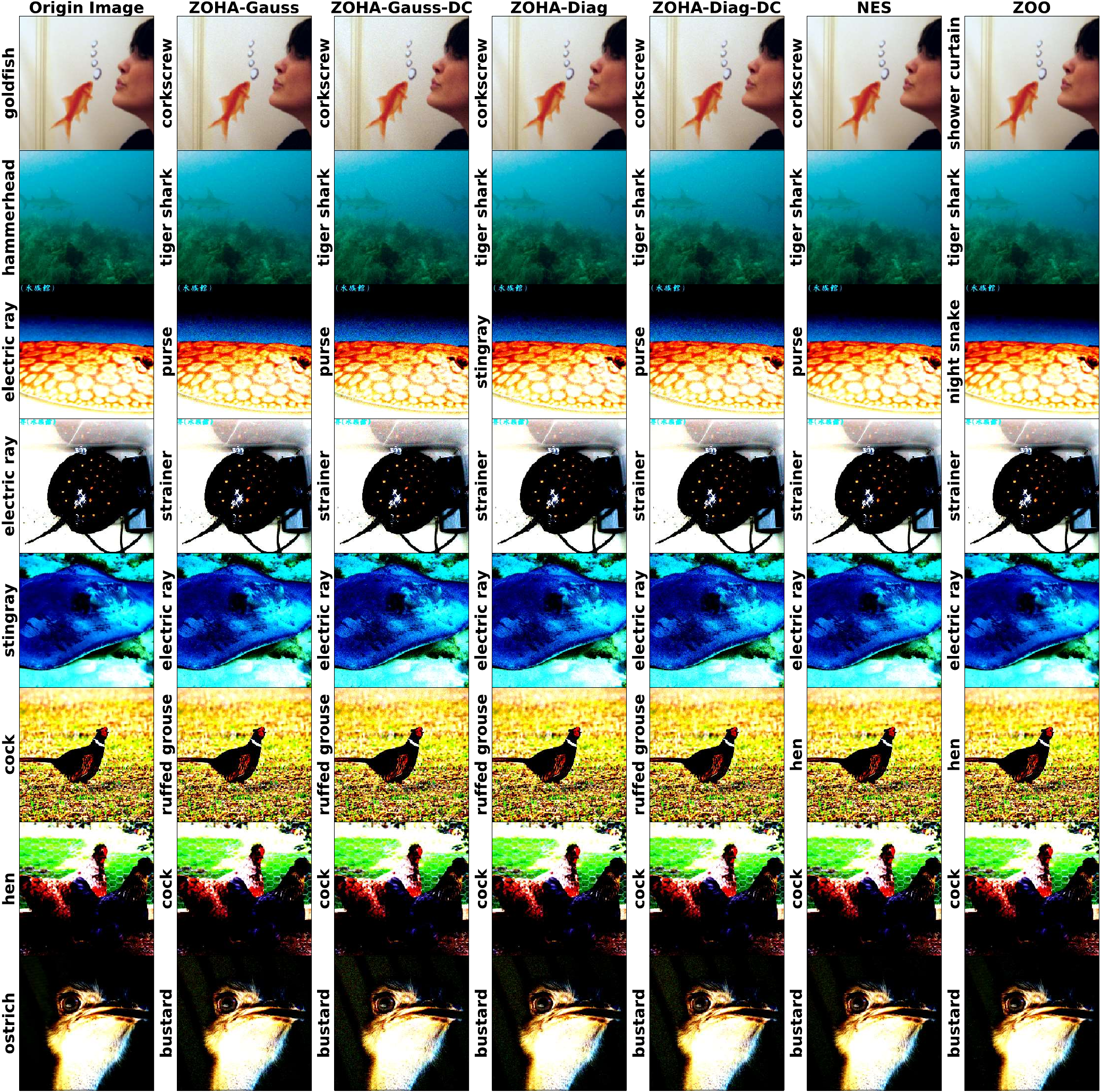}
	\end{center}
	\caption{Un-targeted adversarial examples on ResNet50 and ImageNet. The label is denoted on the left side.}
	\label{fig:imagenet_untar}
\end{figure}

\end{document}